%% file: main.tex
\documentclass[10pt,twocolumn,letterpaper]{article}

\usepackage[pagenumbers]{wacv} 

\usepackage{booktabs}
\usepackage{xcolor}
\usepackage{makecell}
\usepackage{pifont}
\usepackage{adjustbox}
\usepackage{caption}
\usepackage{colortbl}
\usepackage{array}
\usepackage{multirow}
\usepackage[normalem]{ulem}

\usepackage{amssymb}
\newcommand{\cmark}{{\color{green!70!black}\ding{51}}}
\newcommand{\xmark}{{\color{red!70!black}\ding{55}}}
\definecolor{oursrow}{RGB}{232, 245, 255}
\usepackage{fontawesome}
\usepackage{caption}
\usepackage[T1]{fontenc}
\usepackage{amsmath}
\usepackage{minitoc}
\usepackage{titletoc}
\usepackage{placeins}
\usepackage[pagebackref=true,breaklinks=true,colorlinks,bookmarks=false]{hyperref}
\usepackage{xcolor}
\usepackage{graphicx,verbatim}
\input{preamble}
\usepackage{xcolor}

\definecolor{wacvblue}{rgb}{0.21,0.49,0.74}
\PassOptionsToPackage{pagebackref,breaklinks,colorlinks,allcolors=wacvblue}{hyperref}
\def\wacvPaperID{2572} 
\def\confName{WACV}
\def\confYear{2027}

\title{\textsc{Sureon}: A Benchmark and Vision-Language Model for Surgical Reasoning}

\author{%
Alejandra Perez\textsuperscript{1,*}\quad
Anita Rau\textsuperscript{1,*}\quad
Busisiwe Mlambo\textsuperscript{1}\quad
Lee White\textsuperscript{1}\quad
Aina Brunet\textsuperscript{2}\\[0.2em]
Juan Verde\textsuperscript{3,4}\quad
Juan Caicedo\textsuperscript{5}\quad
Jo\"el L. Lavanchy\textsuperscript{6,7}\quad
Adam Schmidt\textsuperscript{1}\\[0.2em]
Chinedu Nwoye\textsuperscript{1}\quad
Omid Mohareri\textsuperscript{1,$\dagger$}\quad
Muhammad Abdullah Jamal\textsuperscript{1,$\dagger$}\\[0.7em]
{\small 
\textsuperscript{1}Intuitive Surgical Inc., USA}\quad
{\small \textsuperscript{2}Hospital Universitari Bellvitge, Spain}\quad
{\small
\textsuperscript{3}IHU Strasbourg, France} \\
{\small\textsuperscript{4}University of Strasbourg, CNRS, France}\quad
{\small
\textsuperscript{5}Fundaci\'on Santa Fe de Bogot\'a, Colombia}\\[0.15em]
{\small
\textsuperscript{6}University Digestive Health Care Center -- Clarunis, Switzerland}\quad
{\small
\textsuperscript{7}University of Basel, Switzerland}\\[0.4em]
{\ttfamily\small
\{alejandra.perezrondon, anita.rau\}@intusurg.com}\\
{\ttfamily\small\url{https://aperezr20.github.io/sureon}}
}

\begin{document}
\maketitle

\renewcommand{\thefootnote}{\fnsymbol{footnote}}
\footnotetext[1]{\textsuperscript{,$\dagger$}Equal contribution.}
\renewcommand{\thefootnote}{\arabic{footnote}}

\input{sec/0_abstract}    
\input{sec/1_intro_new}

\input{sec/related_work}
\input{sec/overview}
\input{sec/2_dataset}
\input{sec/3_model}
\input{sec/4_results}

\input{sec/5_conclusions}

{
    \small
    \bibliographystyle{ieeenat_fullname}
    \bibliography{main}
}
\appendix
\input{supplementary/X_suppl}

\end{document}

%% file: preamble.tex
\usepackage{booktabs}
\usepackage{colortbl}
\usepackage{graphicx}
\newcommand{\grayline}{\arrayrulecolor{gray!40}\hline\arrayrulecolor{black}}
   
\usepackage{enumitem}
\usepackage{listings}
\usepackage{xcolor}

\usepackage{algorithm}
\usepackage{algpseudocode}
\lstdefinelanguage{json}{
 basicstyle=\ttfamily\small,
 showstringspaces=false,
 breaklines=true,
 frame=single,
 backgroundcolor=\color{gray!8},
 rulecolor=\color{gray!40},
 keywordstyle=\color{blue!70},
 stringstyle=\color{green!50!black},
 commentstyle=\color{gray},
 morestring=[b]",
}

\newcommand{\icon}[1]{\raisebox{-0.2em}{\includegraphics[height=1em]{figures/icons/#1}}}

\definecolor{phasecol}{HTML}{EAF3FB}   
\definecolor{toolcol}{HTML}{ECF6EC}    
\definecolor{actioncol}{HTML}{FBEFEF}  
\definecolor{phasecol}{HTML}{EAF3FB}   
\definecolor{toolcol}{HTML}{ECF6EC}    
\definecolor{actioncol}{HTML}{FBEFEF}  
\definecolor{mgrow}{RGB}{240,245,255}

\definecolor{heatlow}{RGB}{235,235,240}
\definecolor{heathigh}{RGB}{50,100,175}

\usepackage{arydshln}
\definecolor{mgpos}{RGB}{0,140,0}
\definecolor{mgneg}{RGB}{190,0,0}
\definecolor{mgrow}{RGB}{248,250,253}
\usepackage{booktabs} 
 \usepackage{supertabular}
\definecolor{ColEntityExistence}{HTML}{EDBCAA}
\definecolor{ColEntityAttribute}{HTML}{FFDFB1}
\definecolor{ColActionDescription}{HTML}{A1CCE8}
\definecolor{ColSafetyPractice}{HTML}{9DB79E}

%% file: sec/0_abstract.tex
\begin{abstract}

A capable surgical AI model should reason from visual evidence; yet, many existing surgical benchmarks can be partially solved without visual input, as models may exploit procedural priors or common anatomy–instrument interactions to infer answers from language alone. To address this, we introduce \textsc{Sureon}, a video question-answering (QA) dataset and benchmark designed to evaluate visually grounded surgical reasoning. Our pipeline generates QA pairs from narrated surgical academic videos, refines incorrect options using multiple models to reduce linguistic biases, and uses trained adversarial guessers to filter out questions answerable without visual evidence. The resulting dataset contains 76K training pairs covering 168 procedures and 10 specialties, alongside a benchmark of 340 questions across 10 categories verified by expert surgeons. Benchmarking leading commercial and open-source models shows that surgical visual reasoning remains largely unsolved—no model exceeds 53\% accuracy and even expert surgeons reach only 64.9\%.  We also introduce \textsc{Sureon}-VLM, a baseline trained on our curated data that avoids language shortcuts, outperforms existing open-source models on \textsc{Sureon}, and matches larger specialist models on multiple standard surgical  benchmarks. Our training ablations show that distractor refinement and blind curation strengthen visual grounding, increasing the multimodal gain from 8.7 to 29.1 percentage points while reducing the training set from 176.6K to 76K filtered samples. \textsc{Sureon} thus offers both a verified testbed for surgical visual reasoning and training data that strengthens visual grounding.

\end{abstract}

%% file: sec/1_intro_new.tex
\section{Introduction}
\label{sec:intro}

\begin{figure*}[t]
    \centering
    \includegraphics[width=0.95\linewidth]{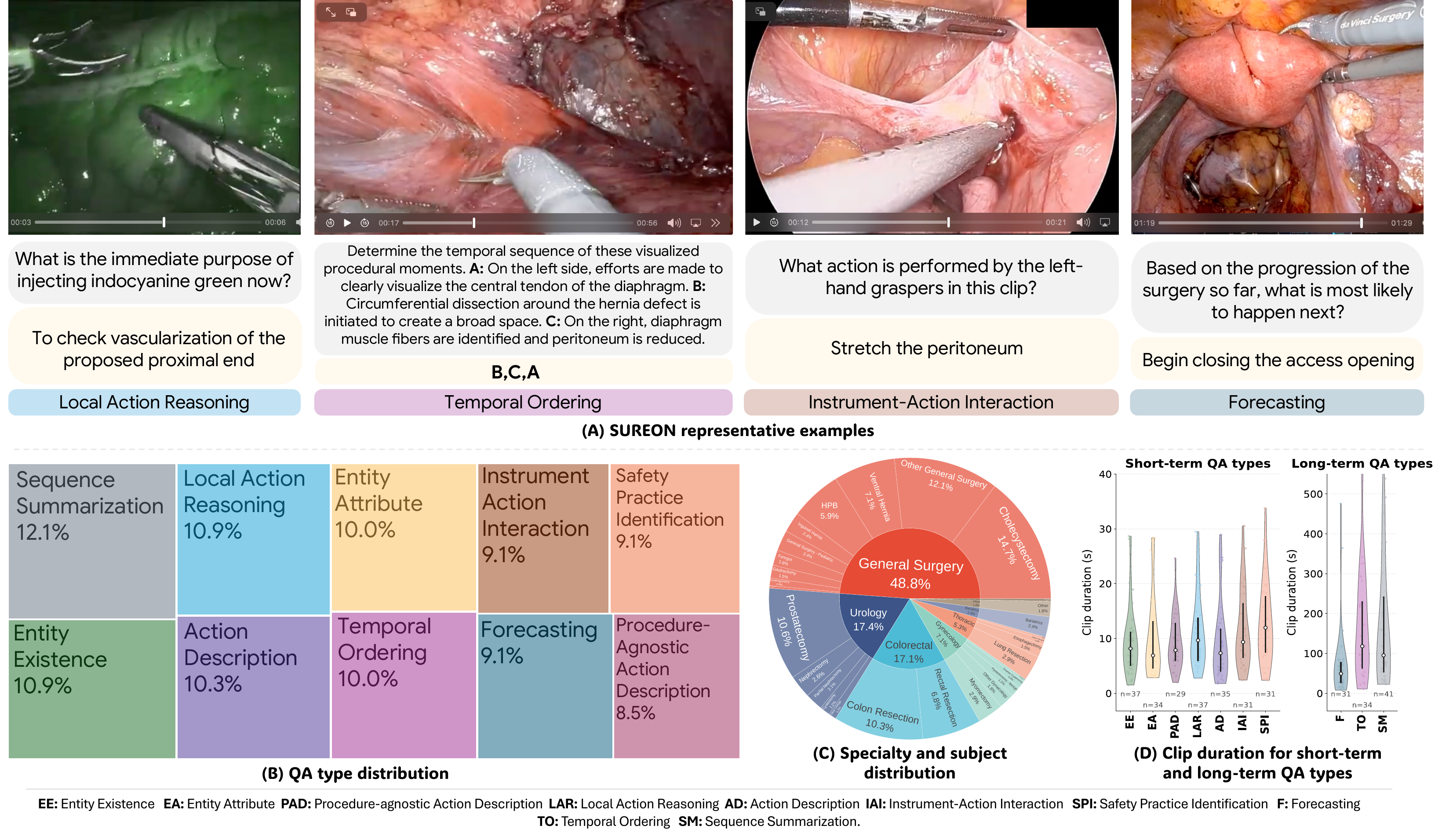}
\caption{\textbf{Overview of \textsc{Sureon} Benchmark.}
\textbf{(A)}~Representative examples for some VQA types
\textbf{(B)}~VQA type distribution.
\textbf{(C)}~Hierarchical distribution of specialties and surgical subjects.
\textbf{(D)}~Clip-duration distributions per QA type, grouped by short-term and 
long-term categories.}
    \label{fig:overview}
\end{figure*}

Surgeons perceive complex visual cues in the operative scene and constantly interpret their clinical meaning, such as whether a dissection plane is safe or what the current tissue state implies for the next step. As AI systems are increasingly explored towards intraoperative decision support and surgical autonomy~\cite{schmidgall2025will, badaloni2026we, shein2023human}, a growing body of work has turned from traditional perception models to Vision Language Models (VLMs) to capture expert-like surgical understanding and reasoning~\cite{zeng2025surgvlm, schmidgall2024gp, wang2025copesd}. Yet existing benchmarks fall short of evaluating the visually grounded reasoning that defines surgical expertise, restricting progress on models with such capabilities.

In the general domain, and increasingly in diagnostic medical fields, evaluation of VLMs has progressed from classification, recognition, and localization tasks to compositional and causal reasoning problems on visual inputs~\cite{yu2025vrbench, yu2026benchmarking, cheng2026v, burgess2025microvqa, zhou2024benchx}. But unlike radiology and pathology, where image–report pairs already exist~\cite{lau2018dataset,chen2024wsi}, surgical video is recorded ad hoc without paired reports. The field has therefore leaned on rigid expert taxonomies, and the benchmarks built on them inherit those closed label spaces~\cite{schmidgall2024gp, seenivasan2023surgicalgpt, liu2026endobench, choi2025surgmllmbench,zeng2025surgvlm, su2026medgrpo}. Moving beyond closed label spaces requires supervision that captures how surgeons describe visual evidence, uncertainty, and clinical implications in natural language, which is hard to obtain at scale.

A particularly rich and scalable source of supervision for surgical VLMs is expert-narrated video lectures that are available for training and communication of best practices. Previous work converts these lectures into open-ended instruction data to train generative surgical assistants~\cite{li2024llava, li2026surgpub, wang2026surgcot}. Such pipelines scale up data generation by leveraging state-of-the-art (SOTA) Large Language Models (LLMs), but they provide limited verification that benchmark questions indeed require visual evidence. It remains unclear whether a high score reflects visual reasoning or just memorized surgical facts—a shortcut recent studies have documented in surgery~\cite{shin2026surgcheck} and beyond~\cite{asadi2026mirage}.  Video lectures are the rare setting in which visual cues and their clinical meaning are made explicit in language; a benchmark built from them should thus evaluate visual reasoning grounded in the surgical scene, not medical knowledge.

The challenge, therefore, is not to generate surgical QA pairs at scale, but to verify that answering them requires vision. Standard multimodal benchmark construction delegates this task to state-of-the-art LLMs: they generate questions, answers, and distractors, judge sample quality, and grade the responses of competing models~\cite{wang2023self,liu2023visual,maaz2024video}. In surgery, every link of this chain is circular. Existing multimodal models perform poorly on visual surgical tasks, and a model cannot grade an exam it would fail itself. The fallback, routing every sample through surgeons, does not scale: expert annotation is the field's long-standing bottleneck. But skipping verification has a cost: in the general domain, large fractions of popular multimodal benchmarks have been shown to be solvable by text-only models, defeating the purpose of multimodal evaluation~\cite{chen2024we}.

Rather than entrusting one general-domain model to generate, judge, and grade its own exam in a domain where it performs poorly, we propose a multi-stage agentic pipeline that yields \textbf{\textsc{Sureon}}, a video QA dataset for surgical visual reasoning. Our data generation pipeline first identifies \emph{Semantic Grounding Moments} (SGMs), time intervals where the surgeon explicitly ties a semantic cue to the visual scene, and converts them into QA pairs spanning ten question types including local action reasoning, forecasting, and safety. Second, retrieval-augmented, multi-model distractor refinement aligns generated distractors with the style of correct answers to minimize statistical biases and mixes LLM generators from different families to remove single-model artifacts. Third, a blind-testing protocol discards samples that an ensemble of finetuned and off-the-shelf LLM \& VLM guessers answers consistently without the video, and an iterative hardening search applies further adversarial pressure to surviving test samples. Lastly, a team of six surgeons verify a subset of the final dataset.

We introduce \textsc{Sureon}, a resource that consists of over 70K rigorously curated video QA pairs for training (\textsc{Sureon}-DB) and 340 expert-reviewed samples for benchmarking. Extensive benchmarking shows that leading commercial and open-source models struggle to reason about visual cues in surgical videos. We further show that a model trained on \textsc{Sureon}-DB yields higher multimodal accuracy gains with fewer training samples than naively generated surgical QA data as it is less reliant on language shortcuts. This model substantially outperforms existing open-source models, and is competitive with leading commercial systems.

%% file: sec/related_work.tex
\section{Related Work}
\label{sec:related_work}
\textbf{Surgical vision--language benchmarks.} 
Surgical benchmarks for multi-modal understanding were first built by translating the annotations of existing datasets into language, reformulating  phase, action, and tool labels into VQA pairs~\cite{seenivasan2022surgical, yuan2024advancing, wang2025copesd, drago2026surgvivqa}. Since these benchmarks were confined to a single procedure, other efforts have aggregated annotations from multiple annotated sources into cross-procedure benchmarks such as SurgVLM-Bench~\cite{zeng2025surgvlm}, EndoBench~\cite{liu2026endobench}, and SurgMLLMBench~\cite{choi2025surgmllmbench} to assess generalization across interventions and tasks. While such annotations give a broad, systematic representation of a scene's observable semantics, these benchmarks stay bound to the rigid label sets of their sources, never reaching perception or reasoning beyond their fixed taxonomies. To mitigate these limitations, recent work has leveraged narrated surgical procedures to source supervision from expert lectures: SurgCoT~\cite{wang2026surgcot} uses narration to probe reasoning such as causal action ordering and cue–action alignment, though its open-ended QA format restricts systematic scoring, while SurgPub-Video~\cite{li2026surgpub} structures narration-derived questions as multiple choice for unambiguous evaluation. However, none of these approaches filter out questions answerable from language priors alone---as is now standard in vision-language benchmark construction in other domains~\cite{burgess2025microvqa,yi2026exact,yang2026captionqa,chen2025document}---nor do they evaluate models when no visual input is provided, despite evidence that Multimodal Large Language Models (MLLMs) score deceptively high on visual perception benchmarks under blind testing~\cite{asadi2026mirage,shin2026surgcheck,rahmanzadehgervi2024vision}. Concurrent work SurgCheck~\cite{shin2026surgcheck} diagnoses this shortcut reliance via paired questions that replace entity names in the question with spatial grounding cues, \textcolor{black}{but is a post-hoc diagnostic tool that requires scene-graph annotations and stays constrained within existing perception ontologies. \textsc{Sureon} instead addresses shortcut reliance at construction time by curating questions under blind-test conditions to encourage visual reliance during both training and evaluation and targeting visual reasoning capabilities beyond fixed label spaces.}

\begin{table}[t]
\centering
\caption{Comparison against existing surgical VL \emph{datasets} (top) and
         \emph{benchmarks} (bottom). \textsc{Sureon} contributes both a large-scale
         training database (\textsc{Sureon}-DB) and an expert-verified evaluation
         benchmark (\textsc{Sureon}), and supports perception
         and reasoning across diverse procedures and video
         modalities. \textbf{Cap.}: Captioning; \textbf{QA}: Question
         Answering; \textbf{P}: Perception \textbf{R}: Reasoning;
         \faEyeSlash: blind-test curation.}
\label{tab:dataset_comparison}
\setlength{\tabcolsep}{4pt}
\resizebox{\columnwidth}{!}{%
\begin{tabular}{lcccccccc}
\toprule
\textbf{Dataset / Benchmark} & \textbf{Format} & \makecell{\textbf{\#Proc.}\\\textbf{Types}} & \textbf{\#Clips} & \makecell{\textbf{\#VL}\\\textbf{Pairs}} & \textbf{P} & \makecell{\textbf{\#VL Pair}\\\textbf{Types}} & \textbf{R} & \textbf{\faEyeSlash} \\
\midrule
\multicolumn{9}{l}{\textit{Training datasets}} \\
\midrule
SVL~\cite{yuan2025learning}                         & Cap. & $>$1   & 25.5K  & 25.5K  & \cmark & 1  & \xmark & \xmark \\
SurgLaVi~\cite{perez2026surglavi}                   & Cap. & 217    & 240K   & 240K   & \cmark & 3  & \xmark & \xmark \\
SurgLaVi-$\mathbf{\beta}$~\cite{perez2026surglavi}  & Cap. & 170    & 112.7K & 112.7K & \cmark & 3  & \xmark & \xmark \\
Surg-396K~\cite{wang2025endochat}                   & QA   & 3      & $-$    & 396K   & \cmark & 7  & \xmark & \xmark \\
Surg-QA~\cite{li2024llava}                      & QA   & $-$   & 44K    & 102K   & \cmark & 5  & \cmark & \xmark \\
SurgVLM-DB~\cite{zeng2025surgvlm}                   & QA   & 16     & $-$    & 7.79M  & \cmark & 10 & \cmark & \xmark \\
SurgPub-Video~\cite{li2026surgpub}                  & QA   & 75     & 10.9K  & 48.5K  & \cmark & 5  & \xmark & \xmark \\
SurgAtlas~\cite{bellos2026surgatlas}                & Cap.+QA & $>$5K & 393K   & 793.2K & \cmark & 10 & \cmark & \xmark \\
\midrule
\textbf{\textsc{Sureon}-DB (Ours)}                           & \textbf{QA} & \textbf{168} & \textbf{63.1K} & \textbf{76.1K} & \cmark & \textbf{10} & \cmark & \cmark \\
\midrule
\multicolumn{9}{l}{\textit{Evaluation benchmarks}} \\
\midrule
Cholec80-VQA~\cite{seenivasan2022surgical}          & QA   & 1      & $-$    & 43K    & \cmark & 3  & \xmark & \xmark \\
EndoVis-18-VQA~\cite{seenivasan2022surgical}        & QA   & 1      & $-$    & 11.8K  & \cmark & 3  & \xmark & \xmark \\
EndoVis-VQLA~\cite{bai2023surgical}                 & QA   & 1      & $-$    & 9.5K   & \cmark & 4  & \xmark & \xmark \\
PSI-AVA-VQA~\cite{seenivasan2023surgicalgpt}        & QA   & 1      & $-$    & 10.3K  & \cmark & 3  & \xmark & \xmark \\
CoPESD~\cite{wang2025copesd}                        & QA   & 1      & $-$    & 17.7K  & \cmark & 2  & \xmark & \xmark \\
SSG-VQA~\cite{yuan2024advancing}                    & QA   & 1      & $-$    & 960K   & \cmark & 5  & \xmark & \xmark \\
SurgVLM-Bench~\cite{zeng2025surgvlm}                & QA   & 16     & $-$    & 6K     & \cmark & 6  & \cmark & \xmark \\
SurgPub-Video~\cite{li2026surgpub}                  & QA   & 75   & $-$    & 3.3K   & \cmark & 5  & \xmark & \xmark \\
SurgCoT~\cite{wang2026surgcot}                      & QA   & 35     & 2.8K   & 78.5K  & \cmark & 5  & \cmark & \xmark \\
SurgAtlas-Bench~\cite{bellos2026surgatlas}          & QA   & $-$    & $-$    & 2.96K  & \cmark & 10 & \cmark & \xmark \\
\midrule
\textbf{\textsc{Sureon} (Ours)}                        & \textbf{QA} & \textbf{69} & \textbf{335} & \textbf{340} & \cmark & \textbf{10} & \cmark & \cmark \\
\bottomrule
\end{tabular}%
}
\end{table}

\noindent \textbf{Multimodal models in surgery.} Data scarcity first made CLIP-style dual-encoder models predominant in surgery, pretrained on lecture narration through contrastive alignment and extended with hierarchical objectives and LLM-based text denoising~\cite{yuan2025learning,yuan2024procedure,perez2026surglavi}. Although these models can match a scene to a label, they cannot describe, explain, or reason. Generative surgical MLLMs followed to cover these needs, some trained on instruction data derived from the classification and spatial annotations of existing datasets~\cite{seenivasan2023surgicalgpt,zeng2025surgvlm} and others on narration-sourced supervision~\cite{wang2025endochat,li2026surgpub}, with recent work targeting reasoning rather than perception alone~\cite{su2026medgrpo}. Concurrently, annotation- and narration-derived supervision have been integrated into a single training corpus with better generalization~\cite{bellos2026surgatlas}.

\noindent \textbf{Automated VQA generation.} Since manual annotation does not scale, recent benchmarks synthesize VQA pairs automatically from annotations or narrations with LLMs, typically in multiple-choice form for objective scoring~\citep{zhang2025automated}. However, naively generated options introduce \textit{language shortcuts}: distractors leave statistical artifacts in length, specificity, or style that models exploit without visual grounding~\citep{wang2023dataset}, and many questions stay answerable from textual priors alone, allowing that blind LLMs achieve high performance~\citep{chen2024we}. Two remedies have emerged to suppress this leakage: refining questions through chain-of-thought rewriting or length-matched distractors~\citep{burgess2025microvqa,yi2026exact}, and removing text-answerable items via blind testing that discards any question a text-only LLM answers consistently across trials~\citep{yue2025mmmu,yang2026captionqa,shabtay2025livexiv}. Such curation is often drastic: ExAct~\citep{yi2026exact} retains only ${\sim}0.9\%$ of its pool, for instance. These pipelines, however, rely almost exclusively on off-the-shelf LLMs prompted zero-shot, which underestimate exploitable leakage; a text-only model \textit{fine-tuned} on the answer distribution can still solve items that such zero-shot testers pass~\citep{asadi2026mirage}.  This fine-tuned-guesser signal has so far been used primarily to audit completed benchmarks rather than to curate samples during generation. In contrast, \textsc{Sureon} integrates distractor refinement, blind filtering, and fine-tuned guesser feedback into a single construction pipeline for surgical video QA.

%% file: sec/overview.tex
\section{SUREON Dataset, Benchmark, and Model}
\label{sec:overview}

\noindent  Moving beyond the closed-space recognition of predefined taxonomies, \textsc{Sureon} tests whether VLMs can reason over actions, states, safety-relevant events, and procedural progression in surgical video. Built from a heterogeneous pool of narrated minimally-invasive laparoscopic and robot-assisted procedures across multiple specialties, it comprises two complementary resources: \textbf{\textsc{Sureon}-DB}, a large and diverse training database of {76,061} automatically pseudo-labeled Video QA pairs for adapting VLMs to surgery, spanning 10 specialties and 168 procedure types, and \textsc{Sureon}, a {340}-sample,  surgeon-verified benchmark for rigorous assessment. This benchmark is highly diverse (8 specialties and 69 procedure types; \textcolor{black}{fewer than DB as the benchmark is drawn from the smaller expert-verified validation split}) and comparable in size with widely used benchmarks in other domains~\cite{dekoninck2026matharena,zan2026multi,yu2023mm}. Unlike previous work (see Table \ref{tab:dataset_comparison}), this data allows not only training surgical VLMs but also systematically assessing their visual reasoning capabilities. For more details, key attributes are summarized in Table~\ref{tab:sureon_stats}.

\begin{figure}[t]
    \centering
    \includegraphics[width=\linewidth]{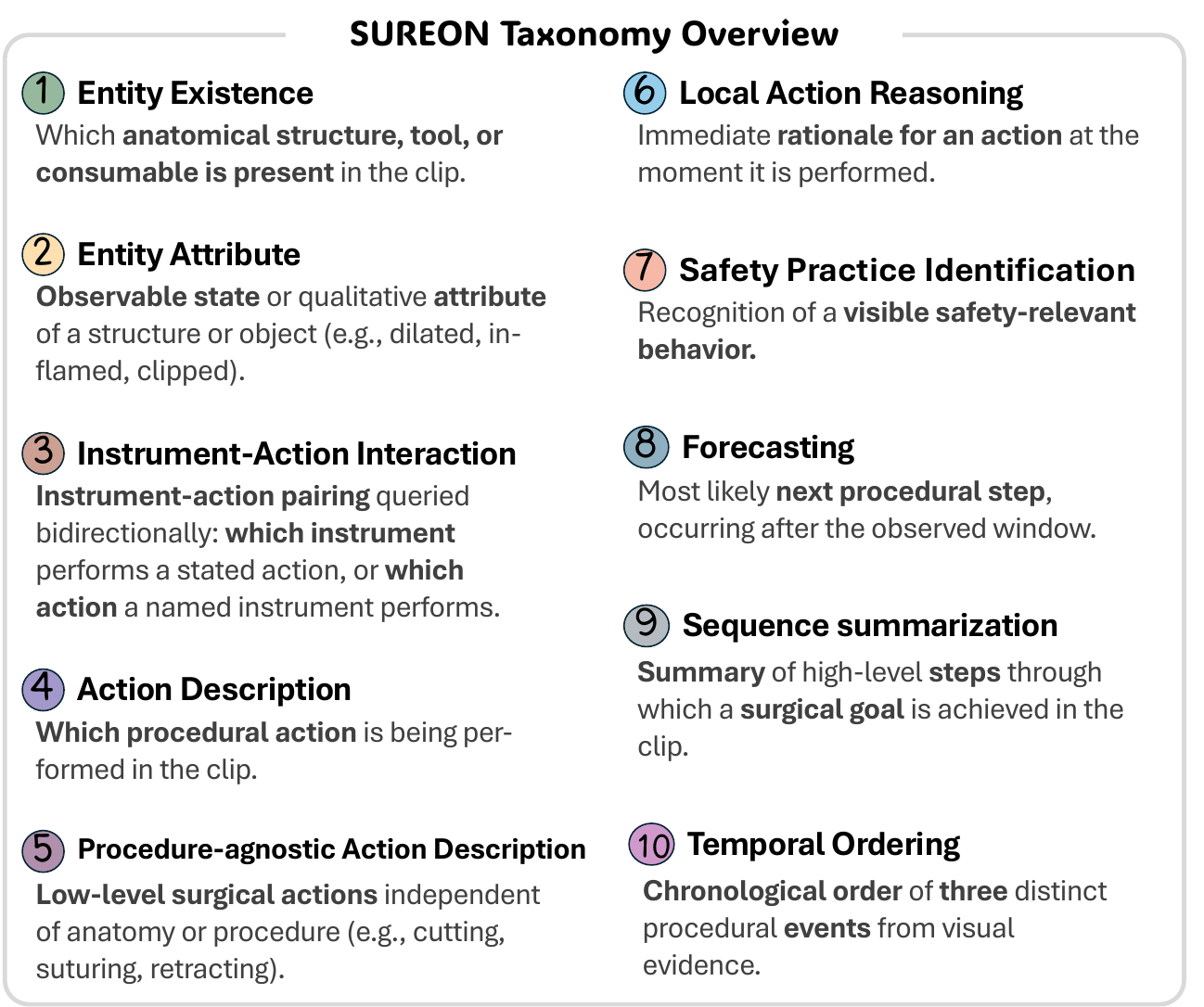}
    \caption{\textbf{\textsc{Sureon} Question--Answer Types.}  Taxonomy definition for all ten QA types in the dataset}
    \label{fig:taxonomy_def}
\end{figure}

\textsc{Sureon} unifies a broad set of capabilities within a single evaluation framework, organized into 10 categories across two temporal granularities (Figure~\ref{fig:taxonomy_def}). In \textit{short temporal windows}, it evaluates recognition of anatomical structures, instruments, instrument-action interactions, entity states, and safety-driven actions, together with locally grounded action reasoning. In \textit{longer contexts}, \textsc{Sureon} evaluates procedural reasoning through sequence summarization, temporal ordering, and prediction of the next procedural step. The first eight categories are posed as single-answer multiple-choice questions, while sequence summarization and temporal ordering are open-ended.

\begin{table}[t]
\centering
\caption{Key attributes of \textsc{Sureon} database  and benchmark.}
\label{tab:sureon_stats}
\small
\setlength{\tabcolsep}{6pt}
\renewcommand{\arraystretch}{1.1}
\resizebox{\columnwidth}{!}{%
\begin{tabular}{lrr}
\toprule
\textbf{Attribute} & \textbf{DB} & \textbf{Bench} \\
\midrule
Samples (QAs) & 76{,}061 & 340 \\
\quad Labeling & auto, pseudo & expert-verified \\
Question categories & 10 & 10 \\
\quad short-window / long-horizon & 7 / 3 & 7 / 3 \\
\quad MCQ / open-ended & 8 / 2 & 8 / 2 \\
Surgical specialties & 10 & 8 \\
Subjects & 37 & 27 \\
Procedure types & 168 & 69 \\
Unique clips (temporal windows) & 63{,}123 & 335 \\
Source narrated videos & 2{,}191 & 165 \\
Expert surgeon reviewers & -- & 6 \\
Median clip duration, short-window (s) & 7.3 & 8.5 \\
Median clip duration, long-horizon (s) & 113.0 & 80.6 \\
\bottomrule
\end{tabular}}
\end{table}

\begin{figure*}[t]
    \centering
    \includegraphics[width=\linewidth]{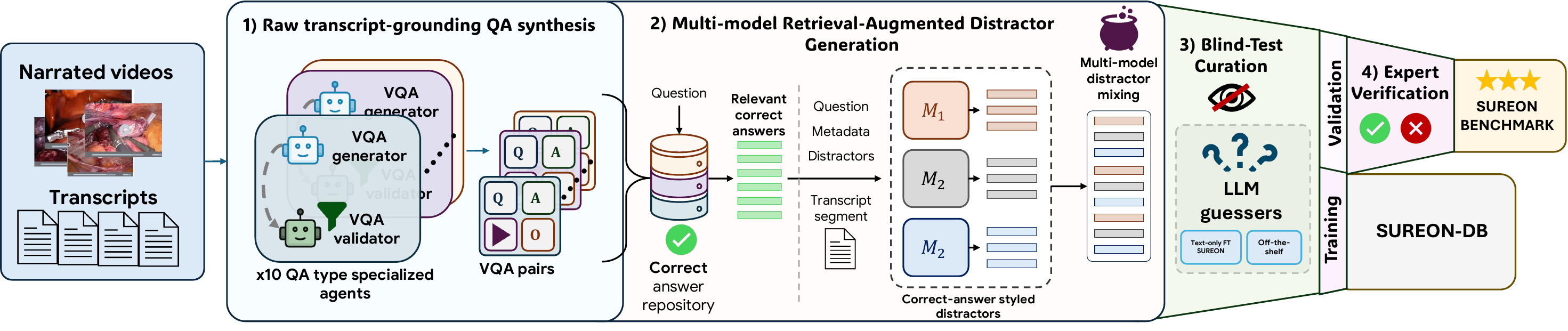}
\caption{\textbf{The \textsc{Sureon} construction pipeline.} A four-stage automated pipeline converting narrated surgical videos into visually grounded QA. \textbf{(1)}~Ten category-specialized generator–validator agent pairs locate
\emph{Semantic Grounding Moments} in the transcript and synthesize validated QA pairs. \textbf{(2)}~Retrieval-augmented distractor generation (RADG) conditions three
LLMs ($M_1$–$M_3$) on retrieved exemplars of correct answers, and mixing across models yields distractors free of single-model shortcuts. \textbf{(3)}~Blind-test
curation discards samples that LLM \emph{guessers} solve without the video, keeping only questions that require visual evidence (guessers include both
an off-the-shelf and fine-tuned models on \textsc{Sureon}'s text distribution). \textbf{(4)}~Surgeons verify the validation split for correctness and grounding.
The curated training split forms the large-scale \textbf{\textsc{Sureon}-DB}; the
expert-verified validation split forms the compact \textbf{\textsc{Sureon}} Benchmark.}
    \label{fig:pipeline}
\end{figure*}

%% file: sec/2_dataset.tex
\subsection{Dataset Construction}
\label{sec:dataset_construction}

As illustrated in Figure~\ref{fig:pipeline}, we introduce a four-stage automated pipeline to build \textsc{Sureon}.
The pipeline addresses three challenges in surgical video-language dataset construction: \textit{(i)} narrations are \textit{selective} and do not densely describe all visible content, \textit{(ii)} naively LLM generated multiple-choice distractors introduce \textit{linguistic shortcuts} that  can be exploited by other LLMs to successfully guess the correct answer, and \textit{(iii)} some generated questions may be answerable from \textit{prior procedural knowledge} without requiring visual evidence. To mitigate these, we first identify \textit{Semantic Grounding Moments} (SGMs) from narration transcripts and use them as anchors for QA generation, then we systematically refine  multiple-choice distractors to mitigate linguistic biases, and finally curate the dataset by removing questions that are solvable without accessing the video, followed by expert verification to  consolidate the benchmark.

\subsubsection*{Stage 1: Raw transcript-grounded QA synthesis}
\label{subsec:sgm_synthesis}

\noindent In lectures, surgeons describe predominantly
clinically salient events, transitions, and safety-relevant observations, so transcripts alone do not offer dense visual supervision for VLM learning. We therefore  anchor QA synthesis to \textit{Semantic Grounding Moments} (SGMs) in the narration  (examples in Figure~\ref{fig:sgm_example} in the  Appendix~\ref{sec:appendix-sgms}).

Given a video $v_i$ and its aligned transcript of timestamped narrated
sentences (formalized in Appendix~\ref{sec:appendix-stage1}), each SGM, $z_{ij}^{k}$ is
a temporal window in which the narration provides task-specific supervision
for a visually observable surgical concept:
\begin{equation}
z_{ij}^{k} = \left(v_i[t_s,t_e],\; s_{ij},\; c_k\right),
\end{equation}
where $v_i[t_s,t_e]$ is the video clip, $s_{ij}$ is the supporting
transcript context,  $c_k$ is the target semantic category, and $i,j,k$ index video, clip, and category. \\

To extract SGMs at scale and convert them into QA samples, we design a pipeline of $K$ generator--validator agent pairs, one per semantic category in Figure~\ref{fig:taxonomy_def}. This decomposition follows the ``divide and conquer'' principle of agentic data synthesis~\cite{mitra2024agentinstruct,deng2025scivideobench}, in which role and category specialization yields higher data quality than prompting a single LLM. Formally, the generation--validation pipeline is described as follows:

\begin{itemize}
    \item \noindent \textit{Generation:} Given a transcript $T_i$, generator $\mathcal{A}_k$ identifies the intervals in which the surgeon explicitly describes content relevant to its assigned category and produces a set of instructional samples $\{x_{ij}^k\}_j$, each defined as:
\begin{equation}
x_{ij}^{k} = \left(v_i[t_s,t_e],\; Q_{ij}^{k},\; A_{ij}^{k},\;
\mathcal{O}_{ij}^{k}\right),
\end{equation}
where $Q_{ij}^{k}$ is the question, $A_{ij}^{k}$ the answer, and $\mathcal{O}_{ij}^{k}$ the set of multiple-choice options, when applicable.  
    \item \noindent \textit{Validation:} The paired validator filters the
generator's output by enforcing category-specific constraints alongside
four universal criteria: hallucination filtering, visual
grounding, temporal-window correctness, and semantic
deduplication (detailed in Appendix~\ref{sec:appendix-multiagent}). The
output of Stage~1 is a transcript-grounded set of surgical video QA pairs
$\mathcal{D}_{\mathrm{SGM}} = \{x_{ij}^{k}\}_{i,j,k}$. \\
\end{itemize}

\noindent Additional implementation details, agent system prompts, per-category agent specifications (Table~\ref{tab:agent_properties}), and the full per-category sample breakdown (Table~\ref{tab:stage1-distribution}) are provided in Appendix~\ref{sec:appendix-stage1}.

\subsubsection*{Stage 2: Multi-Model Retrieval-Augmented Distractor Generation for Bias Mitigation}
\label{subsec:refinement}

\noindent The naively generated distractors from Stage~1 introduce two
compounding biases that let models guess correctly without video. A \textit{distributional gap}, where correct answers and distractors differ
in length, specificity, clinical detail, and phrasing, reveals which option
\textit{looks} correct without any visual grounding; and
\textit{model-specific artifacts}---distractors from a single LLM share
syntactic and lexical patterns---provide a further shortcut that becomes
especially exploitable during fine-tuning. We mitigate both with a
two-component strategy: \textit{(i)}
Retrieval-Augmented Distractor Generation (RADG) to align the distractor
distribution with that of correct answers, and \textit{(ii)} multi-model
mixing to suppress model-specific artifacts:

 \textbf{(i) RADG.}
To close the distributional gap between correct answers and distractors,
we aggregate all correct answers from the training set into a typed
retrieval pool for each question category, $c$:
\begin{equation}
    \mathcal{P}_c = \left\{ A_{ij}^k \;\middle|\; c_k = c \right\},
\end{equation}
and encode each answer with a dense embedding model for retrieval. For each question $Q_{ij}^k$, we retrieve answer-style exemplars from
$\mathcal{P}_{c_k}$, using video metadata to favor answers from procedures
similar to the current one while preserving category-wide diversity, and
merge them into a unified exemplar set $E_{ij}^k$
(Appendix~\ref{sec:appendix-radg}). Each generator $M_\ell$, instantiated
as a state-of-the-art LLM, is then conditioned on the question, its correct answer, the
local transcript context, and the exemplar set $E_{ij}^k$ to produce
distractors that are factually incorrect for the transcript context,
clinically plausible, and stylistically aligned with the retrieved
exemplars.

\textbf{(ii) Multi-model mixing.} Each of the three
distractor-generator LLMs ($M_1$, $M_2$, $M_3$) produces $N$ distractors,
and the final set is drawn at random from the combined $3N$-candidate pool
$\mathcal{D}_{ij}^{k,\mathrm{pool}}$, which empirically reduces short-cut learning compared to employing distractors from a single model (ablation in Sec.~\ref{sec:appendix-stage2}). The output of Stage~2 is the
refined dataset $\mathcal{D}_{\mathrm{ref}}$, in which each Stage-1 sample
is paired with its refined distractor set $\mathcal{O}_{ij}^{k,\mathrm{ref}}$.

\subsubsection*{Stage 3: Visual-Dependence Curation via Blind-Testing Filtering}
\label{subsec:visual_curation}
Distractor refinement reduces linguistic artifacts in the answer
options, yet some samples remain answerable without the video---either
from residual language shortcuts, or because LLMs can resolve certain
questions from procedural knowledge acquired during pretraining. The goal of Step 3 is to remove these samples. To this end, we employ a \textit{blind-testing curation protocol}: an ensemble of LLM
\textit{guessers} $\mathcal{G}$ is queried without video access, receiving
only the question $Q_{ij}^k$ and the refined options
$\mathcal{O}_{ij}^{k,\mathrm{ref}}$. Any sample the \textit{guessers} can solve is
discarded. To separate genuine shortcut exploitation from lucky guesses,
we query each guesser multiple times at non-zero temperature and score it
by its \textit{consistency}, the fraction of trials in which it recovers
the correct answer, rather than by a single correct response, inspired by prior work~\cite{yang2026captionqa,shabtay2025livexiv} (formalized
in Appendix~\ref{sec:appendix-stage3-guessers}). 

To detect candidates answerable from prior knowledge alone, and to additionally account for language shortcuts that only become exploitable once a model is fine-tuned on the answer distribution, our guesser ensemble combines off-the-shelf and blind-fine-tuned models (full specification in Appendix~\ref{sec:appendix-stage3-guessers}). We curate the training set for these fine-tuned guessers via stratified 4-fold cross-validation---guessers trained on the complementary folds and scored on the held-out
fold---yielding the curated training set
\textsc{Sureon}-DB. The validation split is evaluated by the guessers trained on the full training split and evaluated blindly,  yielding $\mathcal{D}_{\mathrm{val}}$.  
\begin{figure}[t]
    \centering
    \includegraphics[width=0.85\linewidth]{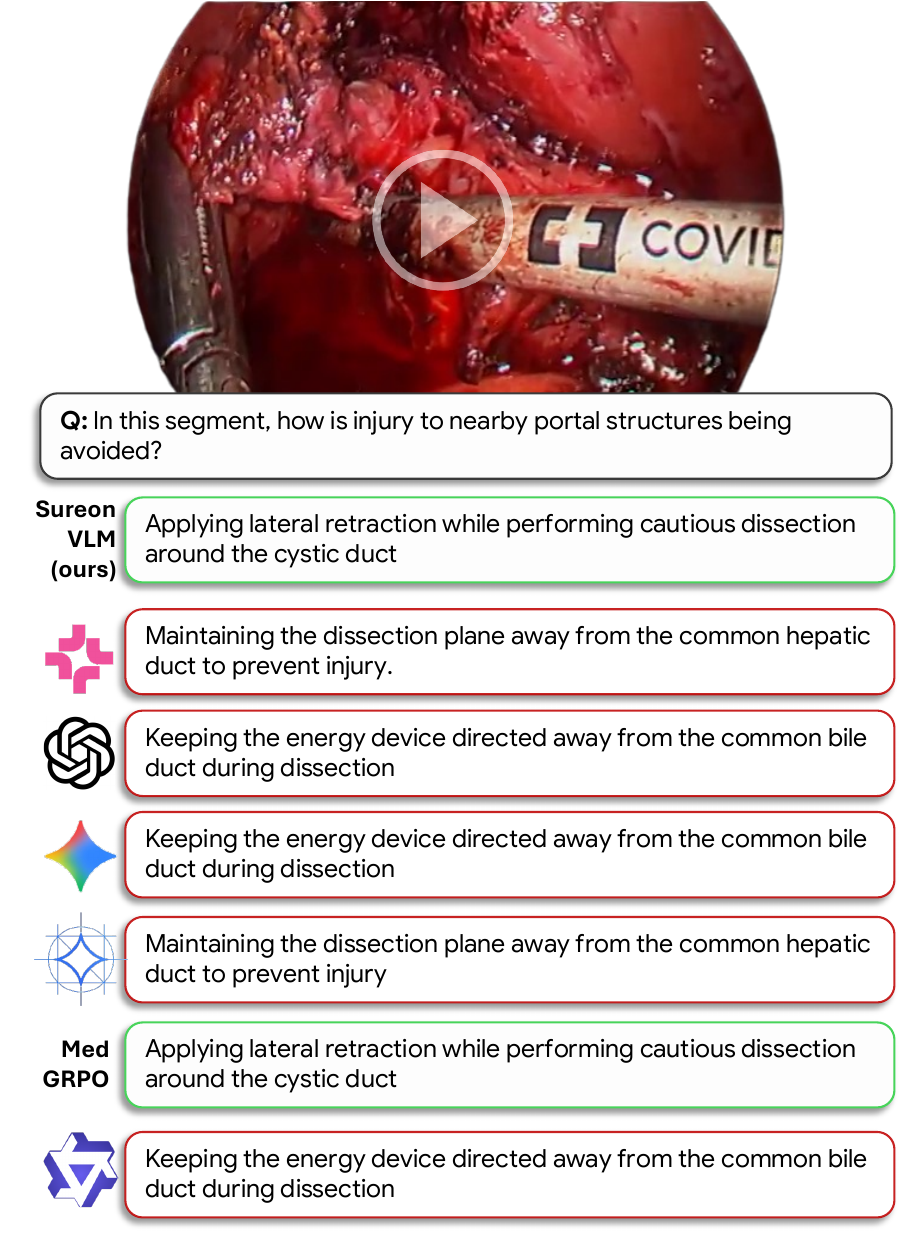}
\caption{\textbf{Qualitative comparison} on a Safety Practice Identification question of \textsc{Sureon}-VLM, (\icon{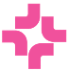})~Molmo2,
(\icon{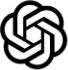})~GPT-5.1, (\icon{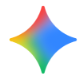})~Gemini~3.1~Pro, 
(\icon{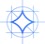})~MedGemma~1.5,
\textcolor{black}{MedGRPO},
and (\icon{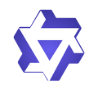})~Qwen3-VL-32B-Thinking. Green boxes denote correct responses and red boxes denote incorrect responses. \textit{Options are presented along with the question, but are omitted for simplicity.}}
\label{fig:qualitative}
\end{figure}

As a final curation step for $\mathcal{D}_{\mathrm{val}}$, we employ \textit{distractor hardening via iterative blind testing}. We fine-tune a
stronger blind guesser on \textsc{Sureon}-DB
and, for each retained sample, iteratively resample distractor
configurations from the pool $\mathcal{D}_{ij}^{k,\mathrm{pool}}$---subject to the cross-model diversity constraint---until the guesser fails
to recover the correct answer under blind testing. The retained
configuration yields the hardened validation set
$\mathcal{D}_{\mathrm{val}}^{\star}$; samples for which no such
configuration exists are discarded
(Appendix~\ref{sec:appendix-stage3-hardening}, Alg.~\ref{alg:hardening}).

\subsubsection*{Stage 4: Expert Verification}
\label{subsec:expert_verification}
Samples from the hardened validation set $\mathcal{D}_{\mathrm{val}}^{\star}$
undergo \textit{manual verification by 6 expert surgeons}. Each reviewer is presented with the
video clip, the aligned transcript segment, the question, the correct
answer, and all distractors, and accepts a sample only when its temporal
alignment, answer correctness, visual grounding, and distractor validity
all hold simultaneously (Appendix~\ref{sec:appendix-reviewers}). Any
sample that is rejected or flagged is discarded under a conservative
retention policy, ensuring the final \textsc{Sureon} benchmark
$\mathcal{B}_{\text{\textsc{Sureon}}} \subset \mathcal{D}_{\mathrm{val}}^{\star}$
satisfies both visual-dependence criteria and expert clinical
judgment. \textcolor{black}{A dual-reviewed subset is additionally used to assess inter-rater reliability (Appendix~\ref{sec:appendix-reviewers}).}

%% file: sec/3_model.tex
\subsection{Sureon-VLM}
\label{sec:model}

Beyond evaluation, \textsc{Sureon}-DB provides a large training dataset. To assess its value, we train a surgical VLM and use it to probe two questions. First, does the standard surgical-VLM recipe generalize to reasoning? Existing surgical assistants are trained by transforming classification-annotated surgery datasets into templated instructions augmented with LLMs~\cite{zeng2025surgvlm,wang2025endochat}; this supervision is mostly perceptual (presence, localization, phase), leaving open whether it extends to reasoning. And second, does consolidating public templated data with \textsc{Sureon} yield a single model strong at both perception and reasoning?

To answer these questions, we assemble a unified instruction-tuning corpus from two complementary sources:
\textit{(i)} image-level annotations from public datasets (tool and anatomy
localization, localized action recognition, and presence/state labels derived from them); and \textit{(ii)} video clips from workflow annotations (phase,
step, and action recognition). Following common practice,
each annotation is mapped to a schema, rendered into natural-language QA
via task-specific templates, and augmented with an LLM for linguistic diversity
(Appendix~\ref{sec:appendix-model-data}). This yields
$\mathcal{D}_{\mathrm{pub}}$ (5.5M QA pairs over 22 public datasets), which we
combine with the \textsc{Sureon} training split into
$\mathcal{D}_{\mathrm{all}} = \mathcal{D}_{\text{\textsc{Sureon}-DB}} \cup \mathcal{D}_{\mathrm{pub}}$.
 
 We then fine-tune Qwen3-VL-8B-Instruct~\cite{bai2025qwen3} on $\mathcal{D}_{\mathrm{all}}$ using 8 frames per clip to obtain \textsc{Sureon}-VLM. We study data composition and staging order (single-stage vs. two-stage schedules) and select the best configuration; the full ablation is reported in Appendix Table~\ref{tab:training-ablation}.

%% file: sec/4_results.tex
\section{Experiments}
\label{sec:experiments}
Experiments on the \textsc{Sureon} benchmark show that (i) no existing model can reliably answer complex questions about surgery, (ii) \textsc{Sureon}  measures visual reasoning, not short-cut learning, and (iii) training on \textsc{Sureon}-DB does not lead to overfitting to a specific task, but we show that \textsc{Sureon}-VLM outperforms existing open-source VLMs on \textsc{Sureon} and has strong performance on traditional classification tasks in surgery.

\input{sec/tables/main_results_round2}

\label{sec:exp-benchmark}
We benchmark commercial general-purpose VLMs (GPT-5.1~\cite{singh2025openai} (high reasoning), Gemini~3.1~Pro~\cite{gemini3pro2025} (high reasoning)); open generalist models (Qwen3-VL-8B-Instruct and Qwen3-VL-32B-Thinking~\cite{bai2025qwen3}, Gemma~4~\cite{gemma42026}, Molmo2~\cite{clark2026molmo2}); and open medical VLMs (MedGemma~1.5~\cite{sellergren2025medgemma}, MedGRPO), and include a human expert baseline. We also evaluate \textbf{\textsc{Sureon}-VLM}, our model fine-tuned on $\mathcal{D}_{\mathrm{all}}$, along with the same model trained on \textsc{Sureon}-DB alone or the public surgical corpus alone.


Of the ten \textsc{Sureon} question types, eight are multiple-choice and scored by accuracy. Temporal ordering (TO) is scored by exact-match accuracy, while sequence summarization (SM) is open-ended and scored by a validated LLM judge (Appendix~\ref{sec:appendix-evaluation}). We report the mean over the accuracy-based types and report SM separately. Additionally, we report multimodal gain~\cite{chen2024we} as a complementary metric assessing how much a model's answers reflect the surgical video rather than language priors. It is defined as the accuracy difference between a model given the visual input ($S_v$) and the same model answering blind ($S_{wv}$): $\mathrm{MG}=S_v-S_{wv}$. We provide additional evaluation and inference details in Appendix (\ref{sec:appendix-eval-sureon}). \\

\noindent \textbf{Visual reasoning in surgery is an unsolved task.}
\label{sec:unsolved}

Table~\ref{tab:generalization} reports per-category accuracy with 95\% confidence intervals on the average. Gemini 3.1 Pro obtains the highest numerical mean ($\mu=52.2$), followed by \textsc{Sureon}-VLM ($\mu=48.2$) and GPT-5.1 ($\mu=46.8$). However, the differences among the strongest models are not statistically significant (Appendix~\ref{sec:appendix-eval-sureon}). In contrast, the human expert baseline reaches ($\mu=64.9$) and significantly outperforms every evaluated model, indicating remaining headroom for surgical visual understanding. Expert performance also varies by category: surgeons approach ceiling on Procedure-Agnostic Action Description (96.5\%), but perform substantially lower on Entity Existence (51.3\%) and Local Action Reasoning (35.1\%), likely reflecting specialty-specific anatomical knowledge and the added difficulty of inferring the purpose of an observed action. Models show a different pattern, with Forecasting among the hardest categories, suggesting that anticipating procedural progression remains harder than describing the current scene. Given the wider confidence intervals at the category level, we interpret these differences as broad trends rather than fine-grained rankings. \\

\noindent \textbf{\textsc{Sureon} measures visual reasoning.}
Across the strongest models, multimodal gain remains substantial: +29.1 for \textsc{Sureon}-VLM, +20.4 for Gemini 3.1 Pro, and +21.1 for GPT-5.1. Notably, \textsc{Sureon}-VLM achieves the largest MG, indicating that its benchmark performance depends strongly on the surgical video rather than on language priors. This visual dependence is progressively strengthened by the \textsc{Sureon} construction pipeline. Table~\ref{tab:benchmark-ablation} shows that MG increases from 8.7 with the raw Stage-1 data to 16.8 after retrieval-augmented distractor generation and to 29.1 after blind-test curation. The final improvement occurs while reducing the training set from 176.6K to 76K samples. Thus, stronger visual grounding results from removing supervision that rewards language shortcuts rather than simply increasing training-set size. \\

\noindent \textbf{\textsc{Sureon}-VLM has broad capabilities.}
The surgical field’s data scarcity has pushed existing surgical VLMs toward classification-derived datasets. The data-composition ablations in Table~\ref{tab:generalization} show that training on public surgical datasets alone performs poorly on \textsc{Sureon} (\(\mu=20.7\), MG \(=8.7\)), whereas training on \textsc{Sureon}-DB alone reaches \(\mu=44.1\) with MG \(=22.1\). Both \textsc{Sureon}-only and combined training significantly outperform public-only training after multiple-comparison correction. When combining the two sources, performance raises \textsc{Sureon} from 44.1 to 48.2.

Public data nevertheless remains complementary. The full \textsc{Sureon}-VLM is trained on both sources so that reasoning-oriented supervision from \textsc{Sureon}-DB is combined with conventional surgical perception supervision.  Table~\ref{tab:public-benchmarks}  shows that this produces a model that remains competitive on established surgical perception benchmarks while also performing strongly on \textsc{Sureon}. Despite its 8B backbone, \textsc{Sureon}-VLM matches or exceeds several larger specialist models on Cholec80 and Endoscapes-CVS, while clip-level inference further improves temporal recognition. The lower performance on HeiChole action recognition suggests that breadth still comes with trade-offs relative to models optimized for narrower tasks.  Consolidating reasoning and perception supervision thus yields a single model that keeps the perception abilities of specialist systems while gaining the visual reasoning they lack (Table~\ref{tab:generalization}).

\begin{table}[t]
\centering
\caption{Benchmark construction ablation. \textsc{Sureon}-VLM trained on the distractors of Stage 1 (Sec. ~\ref{subsec:sgm_synthesis}),  Stage 2 (Sec.~\ref{subsec:refinement}), and Stage 3 (Sec.~\ref{subsec:visual_curation}) and evaluated on \textsc{Sureon}, with multimodal gain $\mathrm{MG} = S_v - S_{wv}$ between visual ($S_v$) and blind ($S_{wv}$) accuracy. MG increases across stages, notably in Stage 3 with less than half the data.}
\label{tab:benchmark-ablation}
\setlength{\tabcolsep}{4pt}
\renewcommand{\arraystretch}{1.05}
\resizebox{\linewidth}{!}{%
\begin{tabular}{@{}lcccc@{}}
\toprule
Distractors & \makecell{\# training \\ samples} & $S_v$ & $S_{wv}$ & MG \\
\midrule
Stage 1 (Baseline)       & 176.6K & 22.4 & 13.7 & 8.7 \\
Stage 2 (RADG)   & 176.6K & 39.5 & \textbf{22.7} & 16.8 \\
Stage 3 (Blind-Test Curation)  & 76K & \textbf{48.2} & 19.1 & \textbf{29.1} \\
\bottomrule
\end{tabular}}
\end{table}

\begin{table}[t]
\centering
\caption{Comparison on public surgical benchmarks. (a) Macro-F1 (\%) on phase and action recognition. (b) Cholec80 detailed metrics (\%). (c) Endoscapes-CVS criterion accuracy (\%). \textcolor{black}{Inference is done at frame-level unless \textbf{(clip)} is specified.}}
\label{tab:public-benchmarks}
\setlength{\tabcolsep}{4pt}
\renewcommand{\arraystretch}{1.05}
\begin{subtable}{\columnwidth}
\centering
\caption{Macro F1 (\%) on phase and action benchmarks.}
\label{tab:public-f1}
\resizebox{\columnwidth}{!}{%
\begin{tabular}{@{}lccc@{}}
\toprule
Method
  & \makecell{Cholec80~\cite{twinanda2016endonet}}
  & \makecell{HeiChole Phase~\cite{wagner2023comparative}}
  & \makecell{HeiChole Action~\cite{wagner2023comparative}} \\
\midrule
GPT-5.1            & 36.0 & 29.0 & 18.0 \\
Gemini~3.1~Pro     & 47.0 & 35.0 & 21.0 \\
Qwen3-VL-8B        & 17.0 & 12.0 & 17.0 \\
SurgAtlas-VLM      & 68.2 & 54.5 & \textbf{55.8} \\
\midrule
\textbf{\textsc{Sureon}-VLM} & 55.33          & 47.1          & 14.5   \\
\textbf{\textsc{Sureon}-VLM (clip)}  & \textbf{78.85} & \textbf{71.4} &\tiny{clip eval not applicable}   \\
\bottomrule
\end{tabular}}
\end{subtable}
\vspace{0.5em}
\begin{subtable}{0.48\columnwidth}
\centering
\caption{Cholec80 detailed.}
\label{tab:cholec80-detailed}
\resizebox{\linewidth}{!}{%
\begin{tabular}{@{}lccc@{}}
\toprule
Method & Acc & Rec. & Prec. \\
\midrule
InternVL3-8B     & 23.9 & 15.2 & 15.2 \\
Qwen2.5-VL-7B    & 30.5 & 16.7 & 26.2 \\
GPT-4o           & 36.4 & 31.0 & 33.0 \\
Gemini~2.0~Fl.   & 38.9 & 36.8 & 40.0 \\
SurgVLM-7B       & 70.3 & 61.9 & 59.8 \\
SurgVLM-72B      & 76.4 & 70.8 & 66.0 \\
SurgAtlas-VLM    & 66.9 & 66.8 & 73.5 \\
\midrule
\textbf{\textsc{Sureon}-VLM} & 71.9           & 60.8           & 57.4           \\
\textbf{\textsc{Sureon}-VLM (clip)}  & \textbf{87.45} & \textbf{82.21} & \textbf{79.78} \\
\bottomrule
\end{tabular}}
\end{subtable}%
\hfill
\begin{subtable}{0.48\columnwidth}
\centering
\caption{Endoscapes-CVS~\cite{murali2023latent}}
\label{tab:endoscapes-cvs}
\resizebox{\linewidth}{!}{%
\begin{tabular}{@{}lcccc@{}}
\toprule
Method & Avg & C1 & C2 & C3 \\
\midrule
InternVL3-8B     & 48.2 & 40.0 & 53.3 & 51.4 \\
Qwen2.5-VL-7B    & 65.9 & 56.1 & 82.4 & 59.2 \\
GPT-4o           &  \textcolor{black}{74.5} & \textcolor{black}{72.6}  & \textcolor{black}{79.5}  & \textcolor{black}{71.2}  \\
Gemini~2.0~Fl.   & 59.6 & 47.8 & 63.9 & 67.1 \\
SurgVLM-7B       & 76.9 & 75.3 & 82.4 & 72.9 \\
SurgVLM-72B      & 76.7 & \textbf{76.5} & 82.8 & 71.0 \\
SurgAtlas-VLM    & 77.7 & 77.1 & 82.9 & 73.0 \\
\midrule
\textbf{\textsc{Sureon}-VLM} & \textbf{78.5}   & 75.6  & \textbf{84.3}   & \textbf{75.6}   \\
\bottomrule
\end{tabular}}
\end{subtable}
\end{table}

%% file: sec/tables/main_results_round2.tex
\begin{table}[t]
\centering

\caption{\textbf{\textsc{Sureon} benchmarking results.}
         Each model is evaluated with image inputs (\faEye) and
         \emph{blind} (\faEyeSlash, no visual input). \textsc{Sureon} is split into
         perception (AD, EE, EA, PAD, IAI) and reasoning (F, TO, SM, LAR, SPI),
         all reported as accuracy. $\mu$ is the pooled accuracy over all
         accuracy-scored items (matching the bootstrap CIs in
         Appendix~D.1, Table~14); SM is scored with an LLM judge
         (\textit{LLMJ}).
         MG (multi-modal gain) $= S_v - S_{wv}$ is reported per model.}
\label{tab:generalization}
\setlength{\tabcolsep}{1.5pt}
\resizebox{\columnwidth}{!}{%
\begin{tabular}{l l |c| ccccc cccc |c}
\toprule
 & & & \multicolumn{10}{c}{\textbf{\textsc{Sureon}}} \\
\cmidrule(lr){4-13}
 & & & \multicolumn{5}{c}{\textbf{Perception}}
 & \multicolumn{4}{c}{\textbf{Reasoning}}
 & \\
\cmidrule(lr){4-8}\cmidrule(lr){9-12}
\textbf{Model} & \textbf{Eval} & $\boldsymbol{\mu}$
    & \textbf{AD} & \textbf{EE} & \textbf{EA} & \textbf{PAD} & \textbf{IAI}    & \textbf{LAR} & \textbf{SPI} & \textbf{F} & \textbf{TO}    & \textbf{SM} \\
 & & \footnotesize\textit{acc.} & \multicolumn{9}{c}{\footnotesize\textit{acc.}} & \footnotesize\textit{llmj} \\
\midrule
    Random & & 14.6 & 14.3 & 14.3 & 14.3 & 14.3 & 14.3 & 14.3 & 14.3 & 14.3 & 16.7 & --- \\
    Human & \faEye & 64.9\,\scriptsize{$\pm$5.3} & 74.3 & 51.3 & 73.5 & 96.5 & 74.2 & 35.1 & 71.0 & 54.8 & 61.8 & --- \\
\midrule
\textit{Commercial} \\
    GPT-5.1 (high)~\cite{singh2025openai} & \faEye & \cellcolor[RGB]{253,124,55}46.8\,\scriptsize{$\pm$5.7} & \cellcolor[RGB]{250,74,41}\underline{57.1} & \cellcolor[RGB]{252,89,45}\textbf{54.0} & \cellcolor[RGB]{253,156,66}38.2 & \cellcolor[RGB]{253,157,67}37.9 & \cellcolor[RGB]{242,57,36}\textbf{61.3} & \cellcolor[RGB]{253,128,56}\underline{46.0} & \cellcolor[RGB]{253,132,57}45.2 & \cellcolor[RGB]{254,192,91}25.8 & \cellcolor[RGB]{252,95,47}\underline{52.9} & \cellcolor[RGB]{254,205,105}29.3 \\
     & \faEyeSlash & \cellcolor[RGB]{254,192,91}25.8\,\scriptsize{$\pm$4.9} & \cellcolor[RGB]{254,218,120}17.1 & \cellcolor[RGB]{254,165,70}35.1 & \cellcolor[RGB]{253,156,66}38.2 & \cellcolor[RGB]{254,208,108}20.7 & \cellcolor[RGB]{254,173,74}32.3 & \cellcolor[RGB]{254,213,114}18.9 & \cellcolor[RGB]{254,202,102}22.6 & \cellcolor[RGB]{254,219,123}16.1 & \cellcolor[RGB]{254,181,79}29.4 & \cellcolor[RGB]{254,233,151}12.2 \\
     & MG & \cellcolor{mgrow}\scriptsize{\textcolor{mgpos}{+21.1}} & \cellcolor{mgrow}\scriptsize{\textcolor{mgpos}{+40.0}} & \cellcolor{mgrow}\scriptsize{\textcolor{mgpos}{+18.9}} & \cellcolor{mgrow}\scriptsize{\textcolor{mgpos}{+0.0}} & \cellcolor{mgrow}\scriptsize{\textcolor{mgpos}{+17.2}} & \cellcolor{mgrow}\scriptsize{\textcolor{mgpos}{+29.0}} & \cellcolor{mgrow}\scriptsize{\textcolor{mgpos}{+27.0}} & \cellcolor{mgrow}\scriptsize{\textcolor{mgpos}{+22.6}} & \cellcolor{mgrow}\scriptsize{\textcolor{mgpos}{+9.7}} & \cellcolor{mgrow}\scriptsize{\textcolor{mgpos}{+23.5}} & \cellcolor{mgrow}\scriptsize{\textcolor{mgpos}{+17.1}} \\
     \addlinespace[3pt]
    Gemini 3.1 Pro~\cite{gemini3pro2025} & \faEye & \cellcolor[RGB]{252,97,47}\textbf{52.2}\,\scriptsize{$\pm$5.5} & \cellcolor[RGB]{252,87,44}54.3 & \cellcolor[RGB]{253,114,52}\underline{48.6} & \cellcolor[RGB]{253,147,63}41.2 & \cellcolor[RGB]{226,25,28}\textbf{69.0} & \cellcolor[RGB]{252,85,44}\underline{54.8} & \cellcolor[RGB]{251,75,41}\textbf{56.8} & \cellcolor[RGB]{242,57,36}\textbf{61.3} & \cellcolor[RGB]{254,212,112}19.4 & \cellcolor[RGB]{235,43,33}\textbf{64.7} & \cellcolor[RGB]{255,248,187}0.0 \\
     & \faEyeSlash & \cellcolor[RGB]{254,174,74}31.8\,\scriptsize{$\pm$5.2} & \cellcolor[RGB]{254,222,128}14.3 & \cellcolor[RGB]{253,157,67}37.8 & \cellcolor[RGB]{253,156,66}38.2 & \cellcolor[RGB]{254,197,97}24.1 & \cellcolor[RGB]{254,202,102}22.6 & \cellcolor[RGB]{245,64,38}59.5 & \cellcolor[RGB]{254,173,74}32.3 & \cellcolor[RGB]{255,229,144}9.7 & \cellcolor[RGB]{253,147,63}41.2 & \cellcolor[RGB]{254,236,158}9.8 \\
     & MG & \cellcolor{mgrow}\scriptsize{\textcolor{mgpos}{+20.4}} & \cellcolor{mgrow}\scriptsize{\textcolor{mgpos}{+40.0}} & \cellcolor{mgrow}\scriptsize{\textcolor{mgpos}{+10.8}} & \cellcolor{mgrow}\scriptsize{\textcolor{mgpos}{+2.9}} & \cellcolor{mgrow}\scriptsize{\textcolor{mgpos}{+44.8}} & \cellcolor{mgrow}\scriptsize{\textcolor{mgpos}{+32.3}} & \cellcolor{mgrow}\scriptsize{\textcolor{mgneg}{-2.7}} & \cellcolor{mgrow}\scriptsize{\textcolor{mgpos}{+29.0}} & \cellcolor{mgrow}\scriptsize{\textcolor{mgpos}{+9.7}} & \cellcolor{mgrow}\scriptsize{\textcolor{mgpos}{+23.5}} & \cellcolor{mgrow}\scriptsize{\textcolor{mgneg}{-9.8}} \\
\midrule
\textit{Open-source generalist} \\
    Qwen3VL-32B-T~\cite{bai2025qwen3} & \faEye & \cellcolor[RGB]{254,167,71}34.1\,\scriptsize{$\pm$5.3} & \cellcolor[RGB]{253,151,64}40.0 & \cellcolor[RGB]{254,189,87}27.0 & \cellcolor[RGB]{253,147,63}41.2 & \cellcolor[RGB]{241,54,36}\underline{62.1} & \cellcolor[RGB]{253,132,57}45.2 & \cellcolor[RGB]{254,189,87}27.0 & \cellcolor[RGB]{254,212,112}19.4 & \cellcolor[RGB]{254,224,133}12.9 & \cellcolor[RGB]{254,164,70}35.3 & \cellcolor[RGB]{254,236,158}9.8 \\
     & \faEyeSlash & \cellcolor[RGB]{254,186,85}27.8\,\scriptsize{$\pm$4.9} & \cellcolor[RGB]{254,184,82}28.6 & \cellcolor[RGB]{254,165,70}35.1 & \cellcolor[RGB]{253,156,66}38.2 & \cellcolor[RGB]{254,223,131}13.8 & \cellcolor[RGB]{253,132,57}45.2 & \cellcolor[RGB]{254,189,87}27.0 & \cellcolor[RGB]{254,192,91}25.8 & \cellcolor[RGB]{254,219,123}16.1 & \cellcolor[RGB]{254,217,118}17.6 & \cellcolor[RGB]{255,239,166}7.3 \\
     & MG & \cellcolor{mgrow}\scriptsize{\textcolor{mgpos}{+6.3}} & \cellcolor{mgrow}\scriptsize{\textcolor{mgpos}{+11.4}} & \cellcolor{mgrow}\scriptsize{\textcolor{mgneg}{-8.1}} & \cellcolor{mgrow}\scriptsize{\textcolor{mgpos}{+2.9}} & \cellcolor{mgrow}\scriptsize{\textcolor{mgpos}{+48.3}} & \cellcolor{mgrow}\scriptsize{\textcolor{mgpos}{+0.0}} & \cellcolor{mgrow}\scriptsize{\textcolor{mgpos}{+0.0}} & \cellcolor{mgrow}\scriptsize{\textcolor{mgneg}{-6.5}} & \cellcolor{mgrow}\scriptsize{\textcolor{mgneg}{-3.2}} & \cellcolor{mgrow}\scriptsize{\textcolor{mgpos}{+17.6}} & \cellcolor{mgrow}\scriptsize{\textcolor{mgpos}{+2.4}} \\
     \addlinespace[3pt]
    Qwen3VL-8B-I~\cite{bai2025qwen3} & \faEye & \cellcolor[RGB]{254,201,101}23.1\,\scriptsize{$\pm$4.7} & \cellcolor[RGB]{254,184,82}28.6 & \cellcolor[RGB]{254,219,123}16.2 & \cellcolor[RGB]{254,217,118}17.6 & \cellcolor[RGB]{254,166,71}34.5 & \cellcolor[RGB]{253,154,66}38.7 & \cellcolor[RGB]{254,213,114}18.9 & \cellcolor[RGB]{254,219,123}16.1 & \cellcolor[RGB]{254,224,133}12.9 & \cellcolor[RGB]{254,190,89}26.5 & \cellcolor[RGB]{255,248,187}0.0 \\
     & \faEyeSlash & \cellcolor[RGB]{254,224,132}13.4\,\scriptsize{$\pm$3.9} & \cellcolor[RGB]{255,231,148}8.6 & \cellcolor[RGB]{255,236,159}5.4 & \cellcolor[RGB]{254,226,137}11.8 & \cellcolor[RGB]{254,197,97}24.1 & \cellcolor[RGB]{254,182,81}29.0 & \cellcolor[RGB]{254,223,131}13.5 & \cellcolor[RGB]{254,212,112}19.4 & \cellcolor[RGB]{255,234,155}6.5 & \cellcolor[RGB]{255,235,156}5.9 & \cellcolor[RGB]{255,248,187}0.0 \\
     & MG & \cellcolor{mgrow}\scriptsize{\textcolor{mgpos}{+9.7}} & \cellcolor{mgrow}\scriptsize{\textcolor{mgpos}{+20.0}} & \cellcolor{mgrow}\scriptsize{\textcolor{mgpos}{+10.8}} & \cellcolor{mgrow}\scriptsize{\textcolor{mgpos}{+5.9}} & \cellcolor{mgrow}\scriptsize{\textcolor{mgpos}{+10.3}} & \cellcolor{mgrow}\scriptsize{\textcolor{mgpos}{+9.7}} & \cellcolor{mgrow}\scriptsize{\textcolor{mgpos}{+5.4}} & \cellcolor{mgrow}\scriptsize{\textcolor{mgneg}{-3.2}} & \cellcolor{mgrow}\scriptsize{\textcolor{mgpos}{+6.5}} & \cellcolor{mgrow}\scriptsize{\textcolor{mgpos}{+20.6}} & \cellcolor{mgrow}\scriptsize{\textcolor{mgpos}{+0.0}} \\
     \addlinespace[3pt]
    Gemma 4~\cite{gemma42026} & \faEye & \cellcolor[RGB]{254,162,69}35.8\,\scriptsize{$\pm$5.3} & \cellcolor[RGB]{253,114,52}48.6 & \cellcolor[RGB]{254,189,87}27.0 & \cellcolor[RGB]{254,172,73}32.4 & \cellcolor[RGB]{252,100,48}51.7 & \cellcolor[RGB]{253,145,62}41.9 & \cellcolor[RGB]{254,180,78}29.7 & \cellcolor[RGB]{254,164,70}35.5 & \cellcolor[RGB]{254,219,123}16.1 & \cellcolor[RGB]{253,147,63}41.2 & \cellcolor[RGB]{255,242,173}4.9 \\
     & \faEyeSlash & \cellcolor[RGB]{254,206,106}21.4\,\scriptsize{$\pm$4.5} & \cellcolor[RGB]{254,227,139}11.4 & \cellcolor[RGB]{254,204,104}21.6 & \cellcolor[RGB]{254,181,79}29.4 & \cellcolor[RGB]{254,166,71}34.5 & \cellcolor[RGB]{254,192,91}25.8 & \cellcolor[RGB]{254,204,104}21.6 & \cellcolor[RGB]{254,219,123}16.1 & \cellcolor[RGB]{255,229,144}9.7 & \cellcolor[RGB]{254,200,99}23.5 & \cellcolor[RGB]{255,242,173}4.9 \\
     & MG & \cellcolor{mgrow}\scriptsize{\textcolor{mgpos}{+14.4}} & \cellcolor{mgrow}\scriptsize{\textcolor{mgpos}{+37.1}} & \cellcolor{mgrow}\scriptsize{\textcolor{mgpos}{+5.4}} & \cellcolor{mgrow}\scriptsize{\textcolor{mgpos}{+2.9}} & \cellcolor{mgrow}\scriptsize{\textcolor{mgpos}{+17.2}} & \cellcolor{mgrow}\scriptsize{\textcolor{mgpos}{+16.1}} & \cellcolor{mgrow}\scriptsize{\textcolor{mgpos}{+8.1}} & \cellcolor{mgrow}\scriptsize{\textcolor{mgpos}{+19.4}} & \cellcolor{mgrow}\scriptsize{\textcolor{mgpos}{+6.5}} & \cellcolor{mgrow}\scriptsize{\textcolor{mgpos}{+17.6}} & \cellcolor{mgrow}\scriptsize{\textcolor{mgpos}{+0.0}} \\
     \addlinespace[3pt]
    Molmo2~\cite{clark2026molmo2} & \faEye & \cellcolor[RGB]{254,221,127}15.0\,\scriptsize{$\pm$4.0} & \cellcolor[RGB]{254,175,75}31.4 & \cellcolor[RGB]{255,228,140}10.8 & \cellcolor[RGB]{254,208,108}20.6 & \cellcolor[RGB]{254,197,97}24.1 & \cellcolor[RGB]{254,219,123}16.1 & \cellcolor[RGB]{255,232,149}8.1 & \cellcolor[RGB]{254,224,133}12.9 & \cellcolor[RGB]{255,239,165}3.2 & \cellcolor[RGB]{255,231,147}8.8 & \cellcolor[RGB]{255,248,187}0.0 \\
     & \faEyeSlash & \cellcolor[RGB]{254,218,120}17.1\,\scriptsize{$\pm$4.2} & \cellcolor[RGB]{254,218,120}17.1 & \cellcolor[RGB]{255,232,149}8.1 & \cellcolor[RGB]{254,181,79}29.4 & \cellcolor[RGB]{254,197,97}24.1 & \cellcolor[RGB]{254,192,91}25.8 & \cellcolor[RGB]{255,236,159}5.4 & \cellcolor[RGB]{254,219,123}16.1 & \cellcolor[RGB]{255,234,155}6.5 & \cellcolor[RGB]{254,200,99}23.5 & \cellcolor[RGB]{255,248,187}0.0 \\
     & MG & \cellcolor{mgrow}\scriptsize{\textcolor{mgneg}{-2.0}} & \cellcolor{mgrow}\scriptsize{\textcolor{mgpos}{+14.3}} & \cellcolor{mgrow}\scriptsize{\textcolor{mgpos}{+2.7}} & \cellcolor{mgrow}\scriptsize{\textcolor{mgneg}{-8.8}} & \cellcolor{mgrow}\scriptsize{\textcolor{mgpos}{+0.0}} & \cellcolor{mgrow}\scriptsize{\textcolor{mgneg}{-9.7}} & \cellcolor{mgrow}\scriptsize{\textcolor{mgpos}{+2.7}} & \cellcolor{mgrow}\scriptsize{\textcolor{mgneg}{-3.2}} & \cellcolor{mgrow}\scriptsize{\textcolor{mgneg}{-3.2}} & \cellcolor{mgrow}\scriptsize{\textcolor{mgneg}{-14.7}} & \cellcolor{mgrow}\scriptsize{\textcolor{mgpos}{+0.0}} \\
\midrule
\textit{Open-source medical} \\
    MedGemma 1.5~\cite{sellergren2025medgemma} & \faEye & \cellcolor[RGB]{255,228,140}10.7\,\scriptsize{$\pm$3.5} & \cellcolor[RGB]{254,227,139}11.4 & \cellcolor[RGB]{255,228,140}10.8 & \cellcolor[RGB]{255,231,147}8.8 & \cellcolor[RGB]{254,197,97}24.1 & \cellcolor[RGB]{255,234,155}6.5 & \cellcolor[RGB]{255,228,140}10.8 & \cellcolor[RGB]{255,234,155}6.5 & \cellcolor[RGB]{255,234,155}6.5 & \cellcolor[RGB]{254,226,137}11.8 & \cellcolor[RGB]{255,248,187}0.0 \\
     & \faEyeSlash & \cellcolor[RGB]{255,228,140}11.0\,\scriptsize{$\pm$3.5} & \cellcolor[RGB]{254,222,128}14.3 & \cellcolor[RGB]{255,232,149}8.1 & \cellcolor[RGB]{255,231,147}8.8 & \cellcolor[RGB]{255,244,176}0.0 & \cellcolor[RGB]{255,234,155}6.5 & \cellcolor[RGB]{254,213,114}18.9 & \cellcolor[RGB]{255,234,155}6.5 & \cellcolor[RGB]{255,234,155}6.5 & \cellcolor[RGB]{254,190,89}26.5 & \cellcolor[RGB]{255,248,187}0.0 \\
     & MG & \cellcolor{mgrow}\scriptsize{\textcolor{mgneg}{-0.3}} & \cellcolor{mgrow}\scriptsize{\textcolor{mgneg}{-2.9}} & \cellcolor{mgrow}\scriptsize{\textcolor{mgpos}{+2.7}} & \cellcolor{mgrow}\scriptsize{\textcolor{mgpos}{+0.0}} & \cellcolor{mgrow}\scriptsize{\textcolor{mgpos}{+24.1}} & \cellcolor{mgrow}\scriptsize{\textcolor{mgpos}{+0.0}} & \cellcolor{mgrow}\scriptsize{\textcolor{mgneg}{-8.1}} & \cellcolor{mgrow}\scriptsize{\textcolor{mgpos}{+0.0}} & \cellcolor{mgrow}\scriptsize{\textcolor{mgpos}{+0.0}} & \cellcolor{mgrow}\scriptsize{\textcolor{mgneg}{-14.7}} & \cellcolor{mgrow}\scriptsize{\textcolor{mgpos}{+0.0}} \\
     \addlinespace[3pt]
    MedGRPO~\cite{su2026medgrpo} & \faEye & \cellcolor[RGB]{254,219,122}16.7\,\scriptsize{$\pm$4.2} & \cellcolor[RGB]{254,192,91}25.7 & \cellcolor[RGB]{254,219,123}16.2 & \cellcolor[RGB]{254,217,118}17.6 & \cellcolor[RGB]{255,228,141}10.3 & \cellcolor[RGB]{254,192,91}25.8 & \cellcolor[RGB]{254,223,131}13.5 & \cellcolor[RGB]{254,224,133}12.9 & \cellcolor[RGB]{255,234,155}6.5 & \cellcolor[RGB]{254,208,108}20.6 & \cellcolor[RGB]{255,248,187}0.0 \\
     & \faEyeSlash & \cellcolor[RGB]{255,229,144}9.7\,\scriptsize{$\pm$3.3} & \cellcolor[RGB]{255,240,167}2.9 & \cellcolor[RGB]{254,204,104}21.6 & \cellcolor[RGB]{255,240,167}2.9 & \cellcolor[RGB]{255,234,153}6.9 & \cellcolor[RGB]{254,224,133}12.9 & \cellcolor[RGB]{255,236,159}5.4 & \cellcolor[RGB]{255,239,165}3.2 & \cellcolor[RGB]{255,239,165}3.2 & \cellcolor[RGB]{254,190,89}26.5 & \cellcolor[RGB]{255,248,187}0.0 \\
     & MG & \cellcolor{mgrow}\scriptsize{\textcolor{mgpos}{+7.0}} & \cellcolor{mgrow}\scriptsize{\textcolor{mgpos}{+22.8}} & \cellcolor{mgrow}\scriptsize{\textcolor{mgneg}{-5.4}} & \cellcolor{mgrow}\scriptsize{\textcolor{mgpos}{+14.7}} & \cellcolor{mgrow}\scriptsize{\textcolor{mgpos}{+3.4}} & \cellcolor{mgrow}\scriptsize{\textcolor{mgpos}{+12.9}} & \cellcolor{mgrow}\scriptsize{\textcolor{mgpos}{+8.1}} & \cellcolor{mgrow}\scriptsize{\textcolor{mgpos}{+9.7}} & \cellcolor{mgrow}\scriptsize{\textcolor{mgpos}{+3.2}} & \cellcolor{mgrow}\scriptsize{\textcolor{mgneg}{-5.9}} & \cellcolor{mgrow}\scriptsize{\textcolor{mgpos}{+0.0}} \\
\midrule
\textit{Ours} \\
    \textsc{Sureon}-VLM (\textsc{Sureon}-only) & \faEye & \cellcolor[RGB]{253,136,59}44.1\,\scriptsize{$\pm$5.7} & \cellcolor[RGB]{252,100,48}51.4 & \cellcolor[RGB]{253,114,52}\underline{48.6} & \cellcolor[RGB]{253,136,59}\underline{44.1} & \cellcolor[RGB]{253,146,62}41.4 & \cellcolor[RGB]{254,192,91}25.8 & \cellcolor[RGB]{253,149,63}40.5 & \cellcolor[RGB]{242,57,36}\textbf{61.3} & \cellcolor[RGB]{253,154,66}\textbf{38.7} & \cellcolor[RGB]{253,136,59}44.1 & \cellcolor[RGB]{255,239,166}7.3 \\
     & \faEyeSlash & \cellcolor[RGB]{254,203,103}22.1\,\scriptsize{$\pm$4.7} & \cellcolor[RGB]{254,175,75}31.4 & \cellcolor[RGB]{254,219,123}16.2 & \cellcolor[RGB]{254,200,99}23.5 & \cellcolor[RGB]{255,234,153}6.9 & \cellcolor[RGB]{254,212,112}19.4 & \cellcolor[RGB]{254,219,123}16.2 & \cellcolor[RGB]{254,164,70}35.5 & \cellcolor[RGB]{254,173,74}32.3 & \cellcolor[RGB]{254,217,118}17.6 & \cellcolor[RGB]{255,245,180}2.4 \\
     & MG & \cellcolor{mgrow}\scriptsize{\textcolor{mgpos}{+22.1}} & \cellcolor{mgrow}\scriptsize{\textcolor{mgpos}{+20.0}} & \cellcolor{mgrow}\scriptsize{\textcolor{mgpos}{+32.4}} & \cellcolor{mgrow}\scriptsize{\textcolor{mgpos}{+20.6}} & \cellcolor{mgrow}\scriptsize{\textcolor{mgpos}{+34.5}} & \cellcolor{mgrow}\scriptsize{\textcolor{mgpos}{+6.5}} & \cellcolor{mgrow}\scriptsize{\textcolor{mgpos}{+24.3}} & \cellcolor{mgrow}\scriptsize{\textcolor{mgpos}{+25.8}} & \cellcolor{mgrow}\scriptsize{\textcolor{mgpos}{+6.5}} & \cellcolor{mgrow}\scriptsize{\textcolor{mgpos}{+26.5}} & \cellcolor{mgrow}\scriptsize{\textcolor{mgpos}{+4.9}} \\
     \addlinespace[3pt]
    \textsc{Sureon}-VLM (Public-only) & \faEye & \cellcolor[RGB]{254,208,108}20.7\,\scriptsize{$\pm$4.5} & \cellcolor[RGB]{253,142,60}42.9 & \cellcolor[RGB]{254,219,123}16.2 & \cellcolor[RGB]{254,221,127}14.7 & \cellcolor[RGB]{254,166,71}34.5 & \cellcolor[RGB]{254,182,81}29.0 & \cellcolor[RGB]{255,228,140}10.8 & \cellcolor[RGB]{254,212,112}19.4 & \cellcolor[RGB]{255,234,155}6.5 & \cellcolor[RGB]{254,221,127}14.7 & \cellcolor[RGB]{255,245,180}2.4 \\
     & \faEyeSlash & \cellcolor[RGB]{254,226,136}12.0\,\scriptsize{$\pm$3.7} & \cellcolor[RGB]{255,231,148}8.6 & \cellcolor[RGB]{255,232,149}8.1 & \cellcolor[RGB]{255,231,147}8.8 & \cellcolor[RGB]{254,208,108}20.7 & \cellcolor[RGB]{254,202,102}22.6 & \cellcolor[RGB]{255,232,149}8.1 & \cellcolor[RGB]{254,212,112}19.4 & \cellcolor[RGB]{255,234,155}6.5 & \cellcolor[RGB]{255,231,147}8.8 & \cellcolor[RGB]{255,245,180}2.4 \\
     & MG & \cellcolor{mgrow}\scriptsize{\textcolor{mgpos}{+8.7}} & \cellcolor{mgrow}\scriptsize{\textcolor{mgpos}{+34.3}} & \cellcolor{mgrow}\scriptsize{\textcolor{mgpos}{+8.1}} & \cellcolor{mgrow}\scriptsize{\textcolor{mgpos}{+5.9}} & \cellcolor{mgrow}\scriptsize{\textcolor{mgpos}{+13.8}} & \cellcolor{mgrow}\scriptsize{\textcolor{mgpos}{+6.5}} & \cellcolor{mgrow}\scriptsize{\textcolor{mgpos}{+2.7}} & \cellcolor{mgrow}\scriptsize{\textcolor{mgpos}{+0.0}} & \cellcolor{mgrow}\scriptsize{\textcolor{mgpos}{+0.0}} & \cellcolor{mgrow}\scriptsize{\textcolor{mgpos}{+5.9}} & \cellcolor{mgrow}\scriptsize{\textcolor{mgpos}{+0.0}} \\
     \addlinespace[3pt]
    \textsc{Sureon}-VLM & \faEye & \cellcolor[RGB]{253,116,53}\underline{48.2}\,\scriptsize{$\pm$5.7} & \cellcolor[RGB]{233,39,32}\textbf{65.7} & \cellcolor[RGB]{253,114,52}\underline{48.6} & \cellcolor[RGB]{253,122,55}\textbf{47.1} & \cellcolor[RGB]{253,146,62}41.4 & \cellcolor[RGB]{253,154,66}38.7 & \cellcolor[RGB]{253,142,60}43.2 & \cellcolor[RGB]{248,70,40}\underline{58.1} & \cellcolor[RGB]{254,164,70}\underline{35.5} & \cellcolor[RGB]{252,95,47}\underline{52.9} & \cellcolor[RGB]{254,230,145}14.6 \\
     & \faEyeSlash & \cellcolor[RGB]{254,213,114}19.1\,\scriptsize{$\pm$4.3} & \cellcolor[RGB]{254,201,101}22.9 & \cellcolor[RGB]{255,232,149}8.1 & \cellcolor[RGB]{254,208,108}20.6 & \cellcolor[RGB]{255,234,153}6.9 & \cellcolor[RGB]{254,192,91}25.8 & \cellcolor[RGB]{254,223,131}13.5 & \cellcolor[RGB]{254,173,74}32.3 & \cellcolor[RGB]{254,182,81}29.0 & \cellcolor[RGB]{254,221,127}14.7 & \cellcolor[RGB]{255,245,180}2.4 \\
     & MG & \cellcolor{mgrow}\scriptsize{\textcolor{mgpos}{+29.1}} & \cellcolor{mgrow}\scriptsize{\textcolor{mgpos}{+42.9}} & \cellcolor{mgrow}\scriptsize{\textcolor{mgpos}{+40.5}} & \cellcolor{mgrow}\scriptsize{\textcolor{mgpos}{+26.5}} & \cellcolor{mgrow}\scriptsize{\textcolor{mgpos}{+34.5}} & \cellcolor{mgrow}\scriptsize{\textcolor{mgpos}{+12.9}} & \cellcolor{mgrow}\scriptsize{\textcolor{mgpos}{+29.7}} & \cellcolor{mgrow}\scriptsize{\textcolor{mgpos}{+25.8}} & \cellcolor{mgrow}\scriptsize{\textcolor{mgpos}{+6.5}} & \cellcolor{mgrow}\scriptsize{\textcolor{mgpos}{+38.2}} & \cellcolor{mgrow}\scriptsize{\textcolor{mgpos}{+12.2}} \\
\bottomrule
\end{tabular}}
\end{table}

%% file: sec/5_conclusions.tex
\section{Conclusions}
\label{sec:conclusion}
In this work, we introduce \textsc{Sureon}, a high-quality surgical visual reasoning benchmark that leverages expert narration in educational videos. Rather than naively generating VQA samples from lecture videos, which can introduce language-only shortcuts, \textsc{Sureon} combines distractor refinement, blind-test curation, and expert review to preserve questions that require visual evidence from the surgical video. As a result, \textsc{Sureon} supports the training and evaluation of models across multiple semantic categories spanning perception and reasoning tasks in surgery. Our experiments show two things. First, surgical visual reasoning remains far from solved: no model exceeds $53\%$ accuracy, and multimodal gain does not exceed $29.1\%$. Second, verification matters beyond data volume: removing text-solvable samples raises multimodal gain from $8.7$ to $29.1$ while reducing the training set size by $57.0\%$. 
A model trained on this data matches larger specialist models on perception benchmarks while also demonstrating strong visual reasoning on \textsc{Sureon}, outperforming existing open-source models. Together, these findings show that progress in surgical vision--language models requires evaluating and training for dependence on visual evidence. \textsc{Sureon} is a step towards more reliable and visually grounded surgical reasoning.

%% file: supplementary/X_suppl.tex
\clearpage
\setcounter{page}{1}
\onecolumn
\maketitlesupplementary

This supplementary material provides additional details, results, and discussion supporting the main paper. \cref{sec:limitations} covers limitations and future work; the remaining sections expand on the main paper: \cref{sec:appendix-dataset-construction} details the four-stage construction pipeline, including the final dataset configuration (main \cref{sec:dataset_construction}); \cref{sec:appendix-model-data} details the public-data training corpus $\mathcal{D}_{\mathrm{pub}}$ (main \cref{sec:model}); \cref{sec:appendix-evaluation} specifies inference and scoring protocols (main \cref{sec:experiments}); \cref{sec:appendix-training} reports the training data-composition and staging ablations (main \cref{sec:experiments}); and \cref{sec:ethics} discusses ethical considerations.

\vspace{1em}

\startcontents[supplementary]
{
  \hypersetup{linkcolor=black}
  \printcontents[supplementary]{}{1}{\setcounter{tocdepth}{2}}
}

\section{Limitations and future work}   
\label{sec:limitations}

While \textsc{Sureon} advances visually grounded surgical reasoning evaluation, several limitations and avenues for future work remain:

\begin{itemize}
    \item Our refinement and blind-test curation (Sec.~\ref{sec:appendix-stage2}--\ref{sec:appendix-stage3}) reduce text-solvability by operating on the answer options. However, text-solvability is partly a property of how a question is framed: a question that names the operative context or telegraphs the procedural phase can leak its answer regardless of the distractors. Jointly reframing questions alongside distractors is a natural extension that could reduce the number of removed samples in Stage 3.
    
    \item Our guesser ensemble is built from an open-source model (\texttt{Qwen3-VL-8B}), used off-the-shelf and in fine-tuned variants, rather than from commercial models. 
    However, unlike previous work that only use models off-the-shelf, the fine-tuned guessers reveal language shortcuts that only become learnable once a model is exposed to the answer distribution -- leakage that zero-shot off-the-shelf testers, including commercial ones, systematically underestimate~\cite{asadi2026mirage}. \textcolor{black}{Additionally, the precise set of samples removed during curation may vary under a different guesser family or as foundation models become more capable. We therefore complement overall benchmark accuracy with evaluation-time multimodal gain, $MG=S_v-S_{wv}$, which evaluates each model both with and without visual input and measures how much of its performance is attributable to the surgical video rather than language priors or shortcuts. This provides a model-specific check for visual dependence beyond the fixed curation ensemble and allows the benchmark to remain auditable as model capabilities improve.}
    
    \item Every QA pair is anchored to a Semantic Grounding Moment, excluding visually salient content that lecturers do not verbalize. Thus, our benchmark under-samples reasoning that is observable but un-narrated, and its coverage is shaped by what surgeons choose to teach rather than by everything a model could be asked to see.
    
    \textcolor{black}{\item Because \textsc{Sureon} is sourced from educational lecture videos, the case mix likely over-represents textbook presentations, well-rehearsed techniques, and controlled operative conditions relative to the full diversity of real-world practice. Atypical anatomies, intraoperative complications, and less common surgical approaches may therefore be underrepresented, and strong benchmark performance should not be assumed to transfer to the long tail of clinical variability.}
    
    \item \textsc{Sureon} is built from edited, narrated, minimally-invasive lecture recordings. Intraoperative video differs (uncut, without narration, with degraded visibility from smoke, blood, and motion) so strong \textsc{Sureon} performance does not by itself establish readiness for live surgical settings. \textsc{Sureon}-DB is additionally pseudo-labeled and not expert-verified; both resources are intended for research and evaluation, and not for clinical use.
\end{itemize}

\section{\textsc{Sureon} Construction Details}
\label{sec:appendix-dataset-construction}

\subsection{Data Source}
\textsc{Sureon} is built on the SurgLaVi-$\mathbf{\beta}$ dataset~\cite{perez2026surglavi}, which provides timestamped transcriptions of surgical academic lectures for 3,148 
YouTube videos, along with metadata on the surgical specialty, subject, and procedure type of each video.

\begin{figure}
    \centering
    \includegraphics[width=\linewidth]{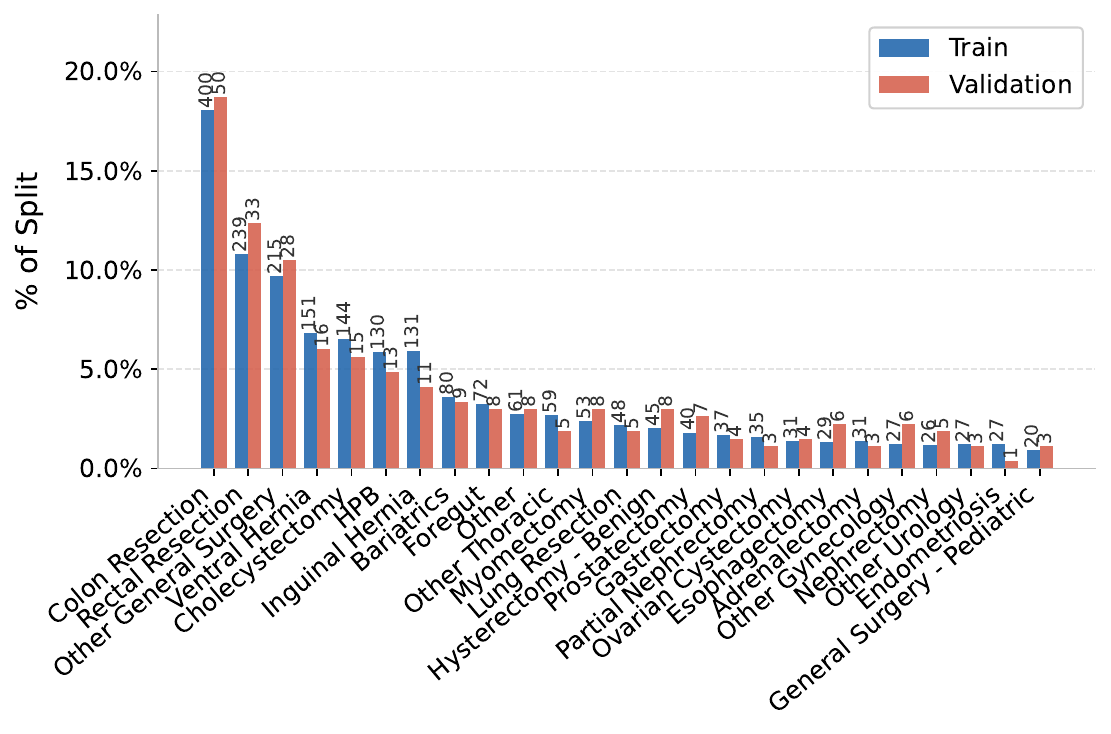}
    \caption{Video subject distribution across training and validation splits (\textit{Included subjects with at least 1\% of representation}).}
    \label{fig:train_test_subject_distr}
\end{figure}

\noindent \textbf{Dataset splits.} SurgLaVi-$\mathbf{\beta}$ was originally released as a pre-training corpus and therefore does not include predefined train/validation/test splits. We performed stratified video-level splitting based on procedure subject, ensuring that the training and validation distributions are balanced across subjects. Approximately 10\% of videos were held out for validation, yielding 2,809 videos for training and 339 for validation. The resulting subject distributions are shown in Figure~\ref{fig:train_test_subject_distr}.

\subsection{Stage 1: Raw transcript-grounded QA synthesis}
\label{sec:appendix-stage1}

\subsubsection{Notation}
\label{sec:appendix-stage1-notation}
Each video $v_i$ is paired with an aligned transcript
$T_i = \{(s_{ij}, [t_{ij}^{\mathrm{start}}, t_{ij}^{\mathrm{end}}])\}_{j=1}^{m_i}$
of $m_i$ timestamped narrated sentences, where $s_{ij}$ is the $j$-th
narrated sentence and $[t_{ij}^{\mathrm{start}}, t_{ij}^{\mathrm{end}}]$
its spoken interval. From this transcript, a Semantic Grounding Moment is
the tuple
$z_{ij}^{k} = (v_i[t_s,t_e],\, s_{ij:j+r},\, c_k)$, pairing a video clip
$v_i[t_s,t_e]$ with its supporting transcript context $s_{ij:j+r}$ and a
target semantic category $c_k$. Each SGM is converted into one or more QA
samples
$x_{ij}^{k} = (v_i[t_s,t_e],\, Q_{ij}^{k},\, A_{ij}^{k},\, \mathcal{O}_{ij}^{k})$,
where $Q_{ij}^{k}$ is the question, $A_{ij}^{k}$ the ground-truth answer,
and $\mathcal{O}_{ij}^{k}$ the multiple-choice option set when applicable.
This notation is used throughout the remaining construction stages.

\subsubsection{Semantic Grounding Moments}
\label{sec:appendix-sgms}
\begin{figure*}
    \centering
    \includegraphics[width=\linewidth]{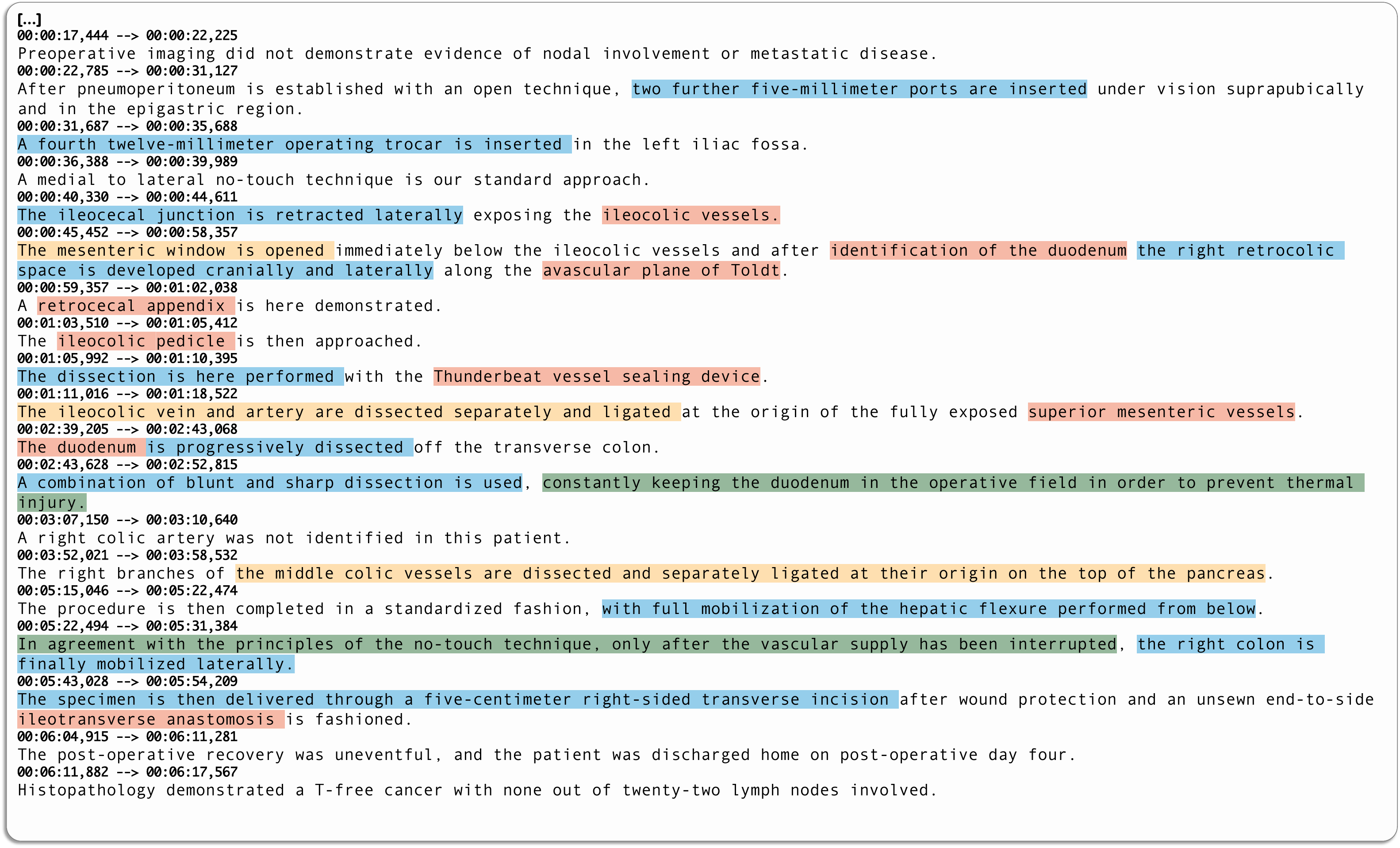}
    \caption{Semantic Grounding Moments (SGMs) identified in a transcript  from a laparoscopic colon resection procedure, color-coded by semantic category: 
    \colorbox{ColEntityExistence}{Entity Existence}, 
    \colorbox{ColEntityAttribute}{Entity Attribute}, 
    \colorbox{ColActionDescription}{Action Description}, and 
    \colorbox{ColSafetyPractice}{Safety Practice Identification}. Four categories are shown for visual clarity, as displaying all categories 
    simultaneously would result in overlapping highlights.}
    \label{fig:sgm_example}
\end{figure*} 
Surgical narrations are sparse: even when instruments, anatomical structures, or actions are visible in the operative field, surgeons do not continuously annotate them in speech. If raw narration segments are used directly as scene descriptions, the resulting labels are incomplete and potentially misleading-- for instance, a narrator may mention a vessel sealer while a bipolar forceps is simultaneously visible but left unaddressed. Treating such a segment as a full scene description would introduce label noise rather than grounded supervision. To address this, we identify \textit{Semantic Grounding Moments} (SGMs): focused intervals in the transcript where the narrator explicitly anchors a semantic cue to the current surgical scene. SGMs serve as reliable anchors for structured QA generation, since they represent moments of intentional visual--linguistic correspondence. Figure~\ref{fig:sgm_example} illustrates representative SGMs across multiple semantic categories in the transcript of a surgical video.

\subsubsection{Multi-Agent Data synthesis details}
\label{sec:appendix-multiagent}
\noindent \textbf{Model and inference configuration.}
All generator and validator agents are built on \textcolor{black}{\texttt{gpt-5-2025-08-28}} (OpenAI), with the \texttt{reasoning} parameter set to \texttt{high} to maximize generation quality. Generator agents are queried at temperature $T{=}0.1$ to allow mild variation across repeated runs while preserving factual consistency. Validator agents are queried at temperature $T{=}0.0$ to enforce deterministic, rule-based judgments. \textcolor{black}{Because GPT-5 is a proprietary API model, exact regeneration remains dependent on continued access to this model version and may be affected by provider-side changes.}
 
\noindent \textbf{Specialized agent pairs.}
We instantiate $K{=}10$ generator--validator agent pairs (20 agents in total), one per semantic category, each equipped with detailed system prompts specifying its role, generation instructions, and validation constraints. Table~\ref{tab:agent_properties} summarizes the key instructions and constraints for each agent pair (Full prompts available in the \href{https://github.com/sureonanon-svg/sureonanon-svg.github.io/tree/main/agent_prompts}{anonymized repository}).
 
\noindent \textbf{Multi-call generation.} Each generator agent is invoked $R{=}3$ times per transcript. Given that a single call can miss opportunities for QA generation within a given transcript, we run three independent calls to increase overall recall on finding SGMs and generating QAs from them. The combined outputs of all $R$ calls are then passed jointly to the paired validator, which validates each candidate sample and removes semantic near-duplicates.
 
\noindent \textbf{Validation criteria.} Additional to applying the category-specific constraints listed in Table~\ref{tab:agent_properties}, the validator enforces three universal filtering rules on every generated sample:
\begin{itemize}[leftmargin=1.5em]
    \item \textit{Hallucination filtering.} A sample is discarded if the question or correct answer introduces an entity, action, or clinical claim absent from the paired transcript segment. 
 
    \item \textit{Visual grounding.} A sample is discarded if the question and its ground-truth answer correspond to a narration comment that is not referring to the present visual scene (e.g. comments referring to events that took place in the past, generic procedural information, etc).

    \item \textit{Temporal window correctness.} A sample is discarded if its assigned temporal window does not cover the transcript segment from which the correct answer is derived.

    \item \textit{Semantic deduplication.} Near-duplicate samples arising from multiple generator calls are identified and collapsed. Only the most informative candidate is retained.
\end{itemize}
 
\noindent \textbf{Structured output schemas.} To enable downstream parsing, both agents are prompted to return structured JSON with reasoning traces justifying its decisions. Listing~\ref{lst:gen-schema} and Listing~\ref{lst:val-schema} show the generator and validator output schemas, respectively.
 
\begin{lstlisting}[language=json, caption={Generator agent output schema. Each entry encodes the question, options, answer, temporal window, video scope, and a brief reasoning trace.}, label=lst:gen-schema]
{
  "question":    "...",
  "options":     ["...", "...", "..."],
  "answer":      "...",
  "time_window": {
    "start": "HH:MM:SS.xxx",
    "end":   "HH:MM:SS.xxx"
  },
  "scope":     "clip" | "full_video",
  "reasoning": "1-3 sentences explaining grounding, uniqueness, and category fit."
}
\end{lstlisting}
 
\begin{lstlisting}[language=json, caption={Validator agent output schema. The validator returns a summary of kept and removed samples, with per-removed-sample issue annotations.}, label=lst:val-schema]
{
  "summary": {
    "total_candidates": N,
    "kept":             K,
    "removed":          R
  },
  "kept": [
    {
      "question":    "...",
      "options":     ["...", "...", "..."],
      "answer":      "...",
      "time_window": {"start": "...", "end": "..."},
      "scope":       "clip",
      "reasoning":   "..."
    }
  ],
  "removed": [
    {
      "index":  0,
      "issues": ["...", "..."]
    }
  ]
}
\end{lstlisting}
 
\noindent \textbf{Quantitative summary.}
Stage~1 processes the transcripts of all 3,148 videos in the corpus, of which 2,478~(78.7\%) produced at least one validated QA sample (train: 2,211 | val: 267), yielding 198,248 agent-validated samples in total -- 176,593 in the training split and 21,655 in the validation split, \textcolor{black}{leading to a total cost of approximately 20,000 USD}. Table~\ref{tab:stage1-distribution} reports the sample count and relative frequency per semantic category for each split and the full corpus. The per-category proportions are broadly consistent across splits.

\begin{table}[t]
\centering
\caption{Per-category sample counts after Stage~1 validation ($\mathcal{D}_{\mathrm{SGM}}$), reported for the training split, validation split, and full corpus.}
\label{tab:stage1-distribution}
\setlength{\tabcolsep}{6pt}
\resizebox{\linewidth}{!}{%
\begin{tabular}{lrrrrrr}
\toprule
 & \multicolumn{2}{c}{\textbf{Train}} & \multicolumn{2}{c}{\textbf{Validation}} & \multicolumn{2}{c}{\textbf{Total}} \\
\cmidrule(lr){2-3}\cmidrule(lr){4-5}\cmidrule(lr){6-7}
\textbf{Category} & \textbf{Count} & \textbf{\%} & \textbf{Count} & \textbf{\%} & \textbf{Count} & \textbf{\%} \\
\midrule
AD  & 27,776 & 15.7 &  3,430 & 15.8 & 31,206 & 15.7 \\
EA  & 26,685 & 15.1 &  3,216 & 14.9 & 29,901 & 15.1 \\
EE  & 23,131 & 13.1 &  2,919 & 13.5 & 26,050 & 13.1 \\
LAR & 18,951 & 10.7 &  2,238 & 10.3 & 21,189 & 10.7 \\
TO  & 17,600 & 10.0 &  2,055 &  9.5 & 19,655 &  9.9 \\
PAD & 16,191 &  9.2 &  2,029 &  9.4 & 18,220 &  9.2 \\
SM  & 14,926 &  8.5 &  1,784 &  8.4 & 16,735 &  8.4 \\
IAI & 13,133 &  7.4 &  1,675 &  7.7 & 14,808 &  7.5 \\
SPI &  9,609 &  5.4 &  1,192 &  5.5 & 10,801 &  5.4 \\
F   &  8,591 &  4.9 &  1,092 &  5.0 &  9,683 &  4.9 \\
\midrule
\textbf{Total}                           & \textbf{176,593} & \textbf{100.0} & \textbf{21,655} & \textbf{100.0} & \textbf{198,248} & \textbf{100.0} \\
\bottomrule
\end{tabular}}
\end{table}

\begin{table*}[t]
\centering
\caption{Per-category specialized agent properties. For each of the 10 semantic categories, we report the answer format and the primary instruction and constraints enforced at generation and validation.}
\label{tab:agent_properties}
\setlength{\arrayrulewidth}{0.4pt}
\small
\resizebox{\linewidth}{!}{%
\begin{tabular}{p{0.24\linewidth}p{0.09\linewidth}p{0.34\linewidth}p{0.34\linewidth}}
\toprule
\textbf{Category} & \textbf{Format} & \textbf{Generator — Key Instructions} & \textbf{Validator — Key Rules} \\
\midrule
\multicolumn{4}{l}{\textit{Short-Term }} \\
\midrule
Entity Existence (EE)
    & MCQ
    & Anchored existence-only questions; answer entity must not appear in question phrasing
    & Reject if action, location, or attribute implied; unique answer enforced per window \\
\grayline
Entity Attribute (EA)
    & MCQ
    & Observable qualitative states only; numeric or measured attributes prohibited
    & Reject if action, causality, or non-visual condition; no quantitative measures \\
\grayline
Instrument--Action Interaction (IAI)
    & MCQ
    & Instrument must be explicitly named in transcript; bidirectional query formulation
    & Bidirectional consistency enforced; exactly one instrument--action pair allowed \\
\grayline
Action Description (AD)
    & MCQ
    & Standard procedural verbs only; no intent, outcome, or instrument information
    & Action-only constraint; answer must be visually observable, not inferred from narration \\
\grayline
Procedure-agnostic \ Action Description\ (PAD)
    & MCQ
    & Labels $\leq$3 words; anatomy-, instrument-, and procedure-free phrasing required
    & Reject if label $>$3 words or references anatomy or instruments; one action per window \\
\grayline
Local Action Reasoning (LAR)
    & MCQ
    & Reason must be explicitly narrated; immediate and local purpose only
    & Narration-backed correct answer required; global strategy reasoning rejected \\
\grayline
Safety Practice Identification (SPI)
    & MCQ
    & Explicit safety cue required in transcript; behavior must be directly video-observable
    & Transcript justifies window selection only; answer identifiable from video alone \\
\midrule
\multicolumn{4}{l}{\textit{Mid-, Long-Term}} \\
\midrule
Forecasting (F)
    & MCQ
    & Mid-procedure context only; multiple completed steps required; generic question phrasing
    & Multiple completed steps must support the predicted action; window ends before predicted action \\
\grayline
Sequence Summarization (SM)
    & Open-ended
    & Goal-directed multi-step workflow; ordered procedural steps across clips
    & Temporal sequence of steps required; single-action or unordered answers rejected \\
\grayline
Temporal Ordering (TO)
    & Open-ended
    & Three distinct events; randomized label presentation
    & Order-neutral question phrasing; unambiguous temporal separability required \\
\bottomrule
\end{tabular}%
}
\end{table*}

\subsection{Stage 2: Multi-Model Retrieval-Augmented Distractor Generation}
\label{sec:appendix-stage2}

\begin{figure}
    \centering
    \includegraphics[width=\linewidth]{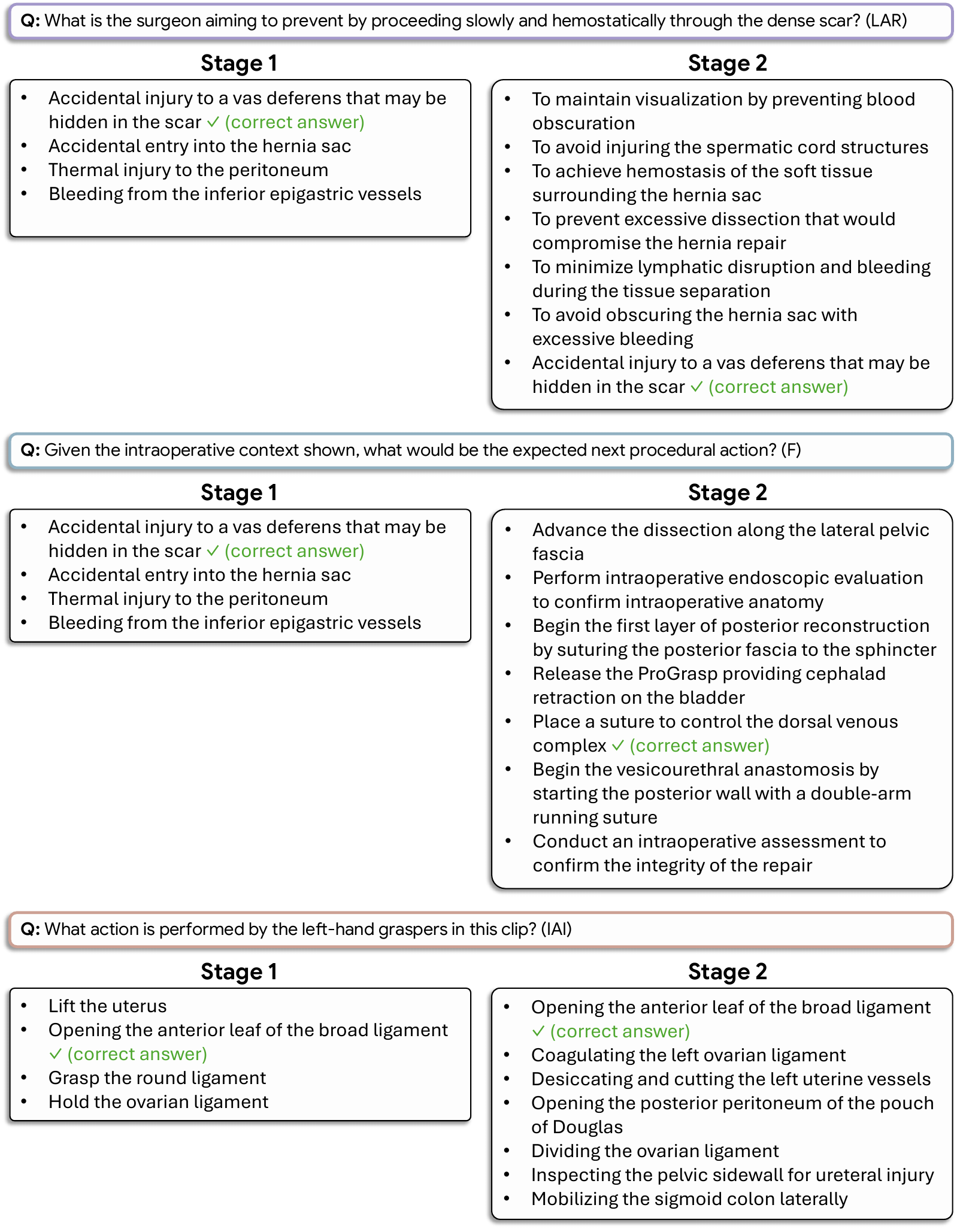}
\caption{\textbf{Qualitative comparison of automatically generated distractors at Stage~1 and Stage~2 on representative examples.} Baseline (Stage~1) distractors are \textbf{vaguer} and shorter than the correct answers, exposing a length and specificity shortcut. In contrast, distractors produced by Retrieval-Augmented Distractor Generation (RADG) and multi-model mixing (Stage~2) are harder, \textbf{matching} the correct answer in length, phrasing, and clinical specificity to reduce blind guessing.}
    \label{fig:dis_comparison}
\end{figure}

\noindent
Stage~1 produces transcript-grounded QA pairs with naively generated multiple-choice distractors. Inspection of these distractors reveals two compounding failure modes. First, correct answers and distractors differ in word length, vocabulary, and level of clinical specificity because distractors are generated from the LLM's base knowledge distribution while the correct answers are derived from the transcript. Overall, this creates surface shortcuts that allow LLMs to identify the correct answer without any visual grounding (see Stage 1 distractors in Figure~\ref{fig:dis_comparison} for representative examples). Second, distractors generated by a LLMs share model-specific syntactic patterns and vocabulary, introducing an additional model-artifact shortcut that is even more harmful during fine-tuning, when the model can learn from these patterns.

Stage~2 mitigates both failure modes through a two-component strategy: Retrieval-Augmented Distractor Generation (RADG) to align the distractor distribution and style with that of correct answers, and \textit{multi-model mixing} to suppress model-specific artifacts. The following sections detail the implementation of each component.

\subsubsection{Retrieval-Augmented Distractor Generation (RADG)}
\label{sec:appendix-radg}

\noindent\textbf{Embedding model and pool construction.} All correct answers in the training set are encoded with \texttt{BAAI/bge-m3}~\cite{chen2024m3}, a dense text embedding model optimized for retrieval-augmented generation and semantic similarity search. Each embedding is stored in a vector database alongside four metadata fields: procedure type, specialty, subject, and question type. \\

\noindent\textbf{Per-question retrieval.} For each question $Q_{ij}^k$ we query four metadata-filtered pools, retrieving the top 100 candidates per pool ranked by cosine similarity to the query embedding: answers matching the same procedure type, specialty, and subject as the source video, plus a broad pool spanning all answers of the same question type with no further metadata restriction. This yields a total of 400 exemplar candidates of correct answers in the dataset across pools. \\
 
\noindent\textbf{Merging and deduplication.} Candidates from the four pools are merged into the exemplar set $E_{ij}^k$ by iterating through them in priority order (procedure $\to$ specialty $\to$ subject $\to$ broad). Candidates already present in the merged list are discarded, yielding a deduplicated exemplar set of at most $K{=}100$ answers that is then passed to distractor generators as answer-style references for the present question.

\noindent \textbf{Distractor Generators.} We instantiate three distractor-generator LLMs available on Hugging Face under the following identifiers:
\begin{itemize}[leftmargin=1.5em]
\item $M_1$: \texttt{google/medgemma-27b-text-it}~\cite{sellergren2025medgemma}
    \item $M_2$: \texttt{meta-llama/Llama-3.1-70B}~\cite{grattafiori2024llama}
    \item $M_3$: \texttt{Qwen/Qwen3.5-35B-A3B}~\cite{team2026qwen3}
\end{itemize}
Each generator produces $N{=}6$ distractors per question, yielding a combined pool of $3N{=}18$ candidates.

\noindent\textbf{Generation conditioning.} Each generator $M_\ell$ is
conditioned on the question $Q_{ij}^k$, its correct answer $A_{ij}^k$, the
original option set $\mathcal{O}_{ij}^k$, the local transcript context
window $H_{ij}^k$, the retrieved exemplar set $E_{ij}^k$, and video-level
metadata $m_i$:
\begin{equation}
    \mathcal{O}_{ij}^{k,\ell} = M_\ell\!\left(
        Q_{ij}^k,\;A_{ij}^k,\;\mathcal{O}_{ij}^k,\;
        H_{ij}^k,\;E_{ij}^k,\;m_i\right).
\end{equation}

\textcolor{black}{\noindent \textbf{Implementation details.} All distractor generators are served via vLLM~\cite{kwon2023vllm} on 4$\times$ NVIDIA H200 GPUs with $T{=}0.1$. Requests are dispatched asynchronously with a concurrency limit of 128 (35B and 27B models) or 64 (122B models) in-flight calls. End-to-end distractor generation for the entire dataset takes approximately 6.4 days: Qwen3.5-35B (4.5\,d), LLaMA-70B (1.2\,d), and MedGemma-27B (0.7\,d).}

\begin{figure}[t]
    \centering
    \includegraphics[width=\linewidth]{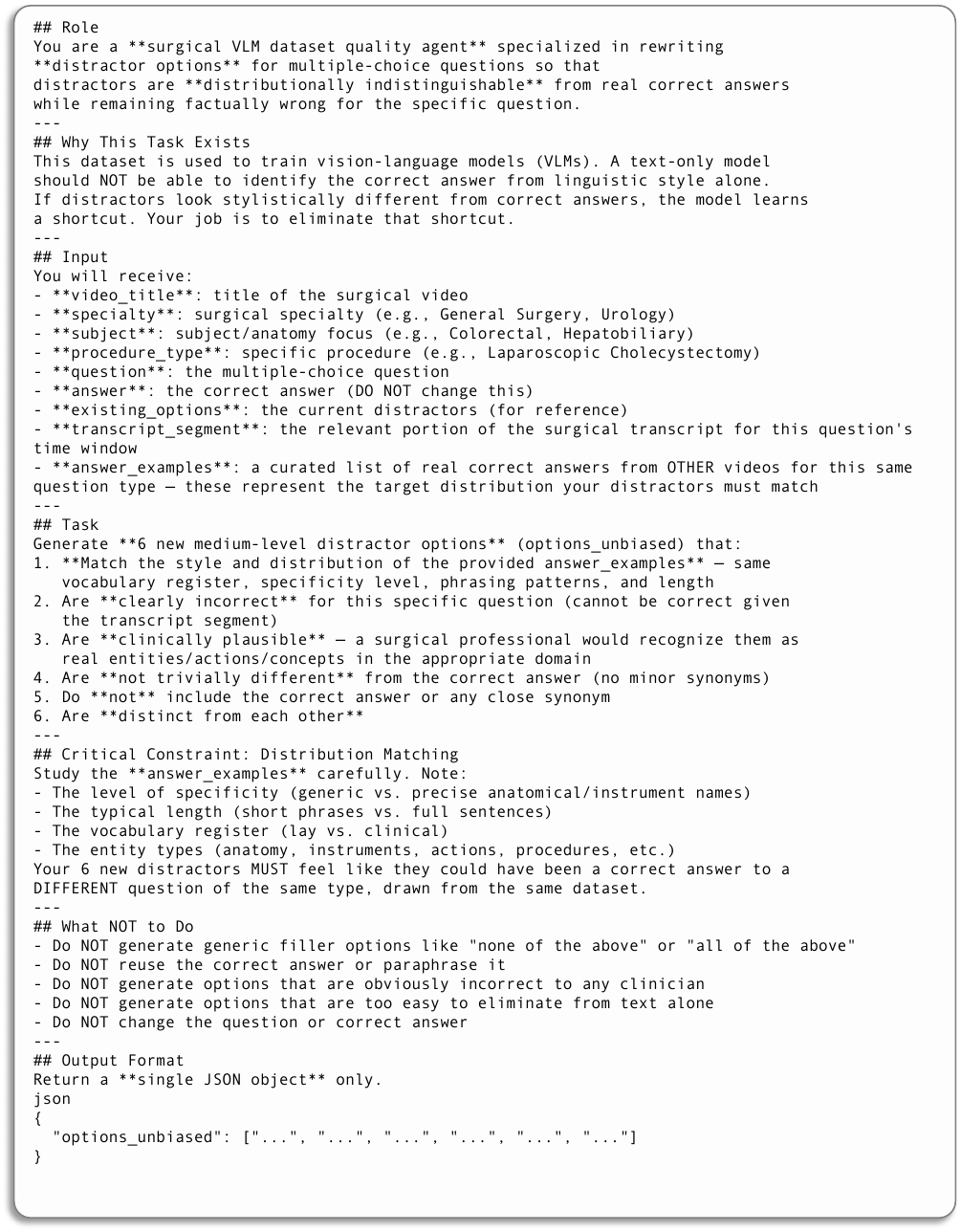}
    \caption{Distractor generation prompt to create distractors that are closer in style to correct answers. }
    \label{fig:distractor-prompt}
\end{figure}

\noindent The same system prompt is applied identically to all three generators and is shown in Figure~\ref{fig:distractor-prompt}. We instruct generators to produce distractors whose length, vocabulary, and
specificity are indistinguishable from the correct-answer style of the provided exemplar, while still being grounded on the transcript context.

\noindent\textbf{Multi-model mixing.}
Each of the three generators contributes $N{=}6$ distractors to the combined pool
$\mathcal{D}_{ij}^{k,\mathrm{pool}} =
\mathcal{O}_{ij}^{k,M_1} \cup
\mathcal{O}_{ij}^{k,M_2} \cup
\mathcal{O}_{ij}^{k,M_3}$. The final distractor set is drawn by random sampling from this pool subject to a cross-model diversity constraint, ensuring that the selected distractors are not all sourced from a single generator.

\noindent\textbf{Blind validation.}
We validate each distractor setup by measuring whether models can exploit linguistic artifacts under \textit{blind testing}: the model receives only the question and answer options, with no video input, and must select the correct answer. Effective distractor refinement should drive blind-tested accuracy toward the random-choice baseline ($p_{\mathrm{rand}} = 0.250$ for 3 distractors and $0.143$ for 6 distractors).

\begin{table}[t]
\centering
\caption{Blind-tested accuracy under different distractor refinement
strategies ($\downarrow$ is better).
\textit{Baseline}: original, unrefined distractors.
\textit{Resampled}: correct answers from other questions of the same
type used as distractors.
$M_1$: MedGemma, $M_2$: Llama, $M_3$: Qwen.
OS: \texttt{Qwen3-VL-8B-Instruct} under blind testing.
FT: same model after blind fine-tuning.
Avg: mean of OS and FT.}
\label{tab:distractor}
\resizebox{\columnwidth}{!}{%
\begin{tabular}{lcc ccc ccc}
\toprule
& & & \multicolumn{3}{c}{3 distractors}
    & \multicolumn{3}{c}{6 distractors} \\
\cmidrule(lr){4-6}\cmidrule(lr){7-9}
Method & RADG & Mix
    & OS $(\downarrow)$ & FT $(\downarrow)$ & Avg $(\downarrow)$
    & OS $(\downarrow)$ & FT $(\downarrow)$ & Avg $(\downarrow)$ \\
\midrule
Random              & $-$        & $-$        & 0.250 & 0.250 & 0.250 & 0.143 & 0.143 & 0.143 \\
\midrule
Baseline            & $-$        & $-$        & 0.577 & 0.818 & 0.698 & $-$   & $-$   & $-$   \\
Resampled           & $-$        & $-$        & 0.556 & \textbf{0.699} & 0.628 & 0.463 & \textbf{0.623} & 0.543 \\
\midrule\midrule
\rowcolor[gray]{0.92}
$M_1$               & $-$        & $-$        & 0.558 & 0.876 & 0.717 & 0.452 & 0.843 & 0.648 \\
$M_2$               & $-$        & $-$        & 0.481 & 0.882 & 0.682 & 0.389 & 0.867 & 0.628 \\
\rowcolor[gray]{0.92}
$M_3$               & $-$        & $-$        & 0.449 & 0.812 & 0.631 & 0.339 & 0.764 & 0.552 \\
\midrule
$M_1$               & \checkmark & $-$        & 0.446 & 0.836 & 0.641 & 0.326 & 0.795 & 0.561 \\
\rowcolor[gray]{0.92}
$M_2$               & \checkmark & $-$        & 0.408 & 0.904 & 0.656 & 0.295 & 0.888 & 0.592 \\
$M_3$               & \checkmark & $-$        & 0.415 & 0.783 & 0.599 & 0.302 & 0.732 & 0.517 \\
\midrule\midrule
\textit{LLM-Mix}    & \checkmark & \checkmark & \textbf{0.401} & 0.789 & \textbf{0.595} & \textbf{0.277} & 0.725 & \textbf{0.501} \\
\bottomrule
\end{tabular}}
\end{table}

\noindent We evaluate two conditions. The \textit{off-the-shelf} (OS) condition applies the model directly under blind testing without any task-specific training, measuring whether the distractor configuration is exploitable at inference time. To assess for shortcut patterns that a zero-shot model cannot exploit but may become learnable once the model is exposed to the distractor distribution during training, the \textit{blind fine-tuning} (FT) condition trains the same model exclusively on (question, options) pairs -- no video -- for the distractor variant being evaluated, then evaluates it under blind testing.

\noindent\textbf{Model and training configuration.}
Both OS and FT conditions use the \texttt{Qwen/Qwen3-VL-8B-Instruct}~\cite{bai2025qwen3} multimodal model. Blind fine-tuning is performed for one epoch with full LLM fine-tuning (no layers frozen), using a learning rate of $1\times10^{-5}$, a per-device batch size of~12, and 4 NVIDIA H200 GPUs. 

\noindent\textbf{Quantitative Results.} Table~\ref{tab:distractor} reports blind-tested accuracy for all configurations under both conditions and both distractor counts (3 and 6). RADG alone reduces OS blind accuracy across all three generators relative to their unaugmented counterparts, confirming that exemplar-conditioned generation narrows the distributional gap. Multi-model mixing further reduces OS accuracy in both settings, with LLM-Mix achieving the lowest blind accuracy overall ($0.401$ for 3 distractors; $0.277$ for 6 distractors). The FT results reveal that the resampled baseline achieves competitive FT scores despite higher OS scores: resampled distractors are harder for fine-tuned models to exploit but remain vulnerable at zero-shot inference time. LLM-Mix reduces OS blind accuracy most consistently across both distractor counts and both evaluation conditions, making it the preferred configuration for the final benchmark. \textcolor{black}{Importantly, RADG reduces blind accuracy across all three model families, while their mixture performs best overall, suggesting that the refinement strategy is not dependent on a single distractor generator.}

\begin{figure}[t]
    \centering
    \includegraphics[width=\linewidth]{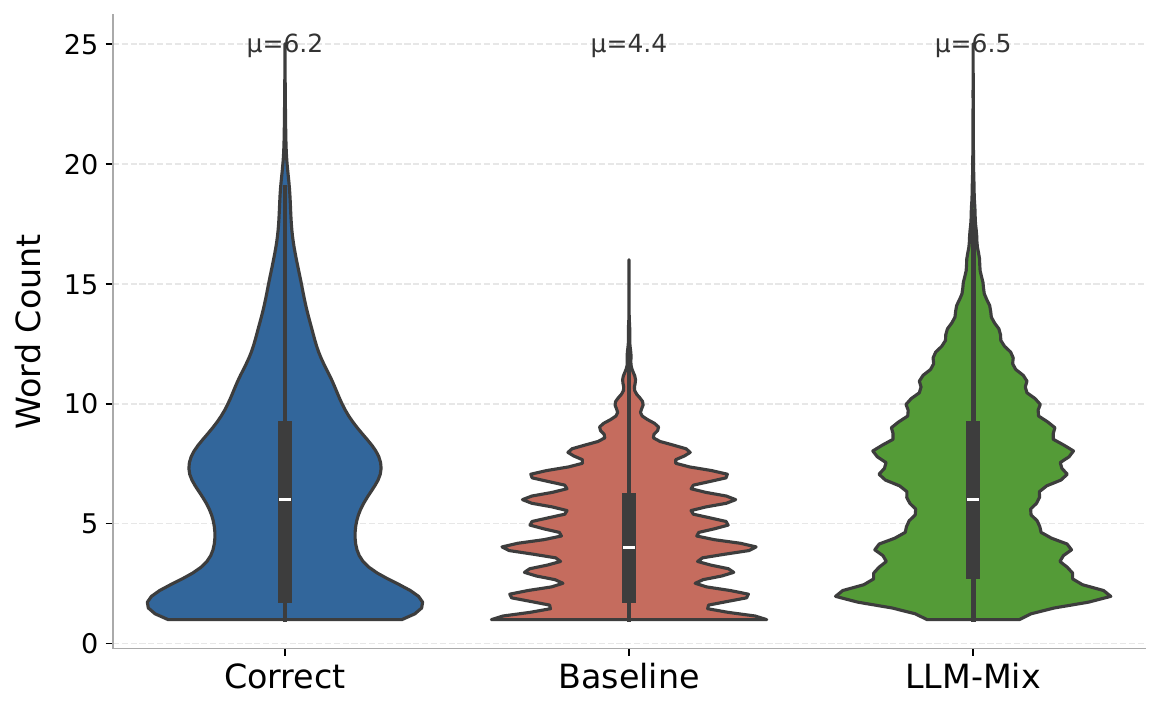}
    \caption{Word-length distributions of correct answers, baseline
             distractors, and LLM-Mix distractors across
             multiple-choice training samples.}
    \label{fig:token_length_dist}
\end{figure}

\noindent \textbf{Qualitative results.} While semantic differences between naive distractors and correct answers--such as clinical specificity or depth of medical knowledge--are difficult to quantify directly, one surface-level characteristic stands out clearly: response length. Correct answers tend to be more explicit and precise, making them longer than naive distractors. This length asymmetry is a spurious cue that off-the-shelf models
can exploit to identify the correct answer without any task-specific fine-tuning. Figure~\ref{fig:token_length_dist} quantifies this gap by plotting the word-count distributions of correct answers, baseline distractors, and LLM-Mix distractors across all multiple-choice samples in the training set. Baseline distractors exhibit shorter sentences compared to correct answers. In contrast, LLM-Mix distractors--conditioned on retrieved exemplars via RADG--closely match the correct-answer distribution in terms of length.

\subsection{Stage 3: Visual-Dependence Curation Details}
\label{sec:appendix-stage3}

As reported in Table~\ref{tab:distractor}, blind fine-tuning on the
refined distractor set drives accuracy above $70\%$ without any video
input, confirming that residual language shortcuts persist even after
Stage~2 refinement. Stage~3 removes samples solvable without visual
evidence by applying a blind-testing consistency filter to both the
training set (via cross-validation) and the validation set.

\subsubsection{Training Set Cross-Validation Setup}
\label{sec:appendix-stage3-folds}

\begin{figure}
    \centering
    \includegraphics[width=\linewidth]{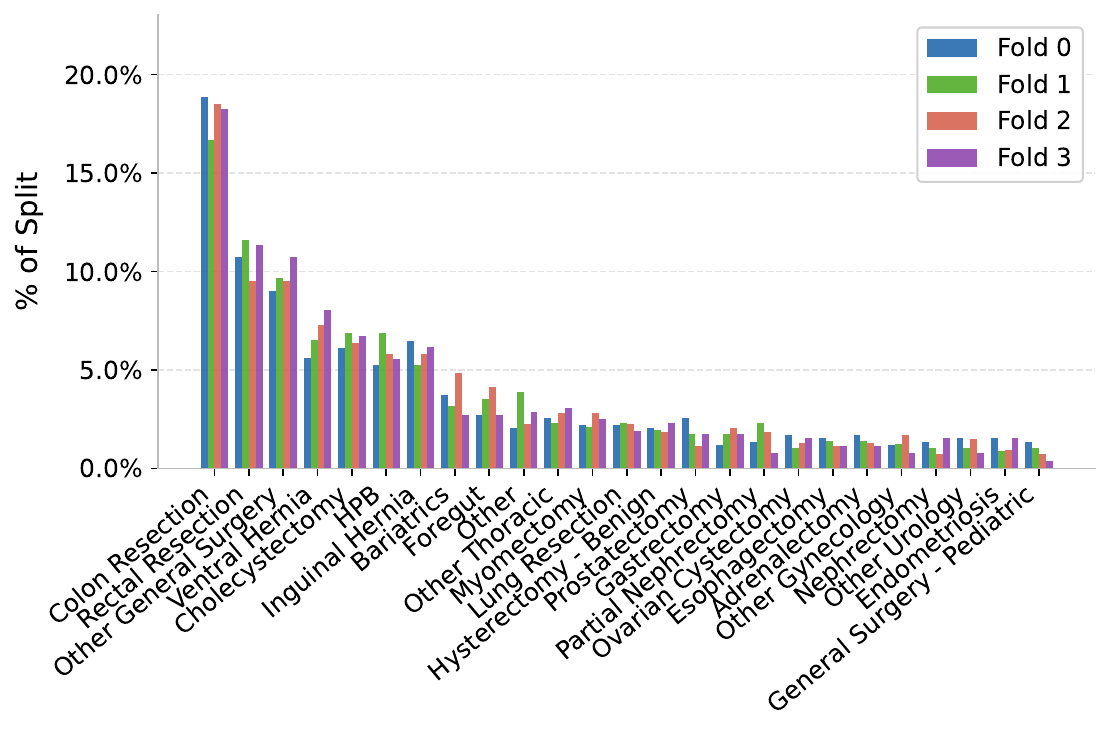}
    \caption{Video subject distribution across folds within the training set (\textit{Included subjects with at least 1\% of representation}).}
    \label{fig:fold_subject_distr}
\end{figure}

The 2,211 training videos that produced at least one validated QA sample
are partitioned into four folds via stratified splitting on procedure
subject, ensuring that the subject distribution is balanced across folds.
Figure~\ref{fig:fold_subject_distr} shows the resulting subject
distribution per fold, Table~\ref{tab:fold-coverage} summarizes the
video and QA counts per fold, and Table~\ref{tab:fold-distribution}
reports the per-category QA breakdown.

\begin{table}[h]
\centering
\caption{Video and QA counts per cross-validation fold (training set).}
\label{tab:fold-coverage}
\setlength{\tabcolsep}{10pt}
\begin{tabular}{lrr}
\toprule
\textbf{Fold} & \textbf{Videos} & \textbf{Total QAs} \\
\midrule
Fold 0 &   587 &  47,153 \\
Fold 1 &   569 &  45,496 \\
Fold 2 &   535 &  43,317 \\
Fold 3 &   520 &  40,627 \\
\midrule
\textbf{Total} & \textbf{2,211} & \textbf{176,593} \\
\bottomrule
\end{tabular}
\end{table}

\begin{table}[t]
\centering
\caption{Per-category QA counts across the four cross-validation folds.
         Category proportions are consistent across folds, confirming
         that stratification did not introduce category-level imbalance.}
\label{tab:fold-distribution}
\setlength{\tabcolsep}{5pt}
\resizebox{\linewidth}{!}{%
\begin{tabular}{lrr rr rr rr}
\toprule
 & \multicolumn{2}{c}{\textbf{Fold 0}}
 & \multicolumn{2}{c}{\textbf{Fold 1}}
 & \multicolumn{2}{c}{\textbf{Fold 2}}
 & \multicolumn{2}{c}{\textbf{Fold 3}} \\
\cmidrule(lr){2-3}\cmidrule(lr){4-5}
\cmidrule(lr){6-7}\cmidrule(lr){8-9}
\textbf{Category}
    & \textbf{Count} & \textbf{\%}
    & \textbf{Count} & \textbf{\%}
    & \textbf{Count} & \textbf{\%}
    & \textbf{Count} & \textbf{\%} \\
\midrule
AD  & 7,415 & 15.7 & 7,131 & 15.7 & 6,911 & 16.0 & 6,319 & 15.6 \\
EA  & 7,191 & 15.3 & 6,862 & 15.1 & 6,408 & 14.8 & 6,224 & 15.3 \\
EE  & 6,222 & 13.2 & 5,884 & 12.9 & 5,716 & 13.2 & 5,309 & 13.1 \\
LAR & 5,111 & 10.8 & 4,867 & 10.7 & 4,685 & 10.8 & 4,288 & 10.6 \\
TO  & 4,514 &  9.6 & 4,488 &  9.9 & 4,468 & 10.3 & 4,130 & 10.2 \\
PAD & 4,330 &  9.2 & 4,289 &  9.4 & 3,919 &  9.0 & 3,653 &  9.0 \\
SM  & 3,979 &  8.4 & 3,836 &  8.4 & 3,632 &  8.4 & 3,479 &  8.6 \\
IAI & 3,543 &  7.5 & 3,478 &  7.6 & 3,147 &  7.3 & 2,965 &  7.3 \\
SPI & 2,612 &  5.5 & 2,457 &  5.4 & 2,367 &  5.5 & 2,173 &  5.3 \\
F   & 2,236 &  4.7 & 2,204 &  4.8 & 2,064 &  4.8 & 2,087 &  5.1 \\
\midrule
\textbf{Total}
    & \textbf{47,153} & \textbf{100.0}
    & \textbf{45,496} & \textbf{100.0}
    & \textbf{43,317} & \textbf{100.0}
    & \textbf{40,627} & \textbf{100.0} \\
\bottomrule
\end{tabular}}
\end{table}

\subsubsection{Guesser Ensemble and Curation Protocol}
\label{sec:appendix-stage3-guessers}

\noindent \textbf{Guesser composition.}
The ensemble $\mathcal{G}$ comprises four guesser types, all evaluated
under blind testing: \textit{(i)} models blind-fine-tuned on the refined
distractors $\mathcal{D}_{\mathrm{ref}}$; \textit{(ii)} models
blind-fine-tuned on the baseline (Stage~1) distractors; \textit{(iii)} an
off-the-shelf base model with no fine-tuning; and \textit{(iv)} models
fine-tuned with full video access but evaluated under blind testing, which
expose shortcuts that video-trained models may absorb from language
co-occurrences during training. Types \textit{(i)}--\textit{(iii)} are used for training-set curation; the
full ensemble \textit{(i)}--\textit{(iv)} is used for validation-set curation.

\noindent \textbf{Model and inference configuration.}
All guessers instantiate \texttt{Qwen/Qwen3-VL-8B-Instruct}~\cite{bai2025qwen3},
the same architecture used in Stage~2 (Sec.~\ref{sec:appendix-stage2}). Ensemble diversity therefore arises from the training setup: the same model is used off-the-shelf and in fine-tuned variants to find language shortcuts (on refined distractors, on baseline
distractors, and with full video access), and all guessers are queried in
text-only mode.

Blind fine-tuned guessers (types \textit{(i)}--\textit{(ii)}, and
\textit{(iv)} for validation-set curation) are trained for one epoch with a
learning rate of $1\times10^{-5}$ on 4 NVIDIA H200 GPUs.
Temporal Ordering; Sequence Summarization is excluded, as summarizing would require the video and no distractors are used. \\

\noindent \textbf{Consistency score and thresholds.}
Each guesser $g \in \mathcal{G}$ is queried $R{=}5$ times at temperature
$T{=}0.7$ with no video input, receiving only the question $Q_{ij}^k$ and
the refined options $\mathcal{O}_{ij}^{k,\mathrm{ref}}$. Repeated queries
are necessary because a single blind query carries a non-negligible
probability of a correct random guess, which would cause spurious removal
of visually grounded samples. The consistency score is the fraction of
runs in which $g$ recovers the correct answer:
\begin{equation}
    S(g,x_{ij}^{k}) = \frac{1}{R}\sum_{r=1}^{R} \mathbf{1}\!\left[
    g_r\!\left(Q_{ij}^k,\;\mathcal{O}_{ij}^{k,\mathrm{ref}}\right)
    = A_{ij}^k\right],
\end{equation}
so that filtering targets only samples reliably solvable from text alone
rather than those guessed correctly by chance. A sample is retained only
if no guesser reaches the consistency threshold $\tau{=}3/5$:
\begin{equation}
    x_{ij}^{k}\ \text{retained} \quad\Longleftrightarrow\quad
    \max_{g \in \mathcal{G}}\,S(g,x_{ij}^{k}) < \tau,
\end{equation}
i.e. a sample is discarded whenever any guesser achieves $S \geq 3/5$.

\noindent \textbf{Training set curation.}
For each fold $f$, guessers of types \textit{(i)}--\textit{(iii)} are
trained on the three complementary folds and scored on the held-out fold.

\noindent \textbf{Validation set curation.}
All guesser types \textit{(i)}--\textit{(iv)} are trained on the full \textsc{Sureon} training set and evaluated on the validation set.

\textcolor{black}{\noindent \textsc{Sureon}-DB quality proxy. We use the out-of-fold blind-test pass rate as a proxy for text-answerable contamination in the automatically generated training data. Of the 161,667 training samples eligible for blind curation (all types except Sequence Summarization), 61,135 (37.8\%) pass the criterion above, while 100,532 (62.2\%) are identified as consistently answerable from text alone and removed. By construction, every sample retained in \textsc{Sureon}-DB passes the held-out blind-guesser criterion, providing a sample-level quality control against language shortcuts without requiring expert review of the full training corpus.}

\subsubsection{Distractor Hardening via Iterative Blind Testing}
\label{sec:appendix-stage3-hardening}


As a final curation step applied to the validation set, we fine-tune
\texttt{Qwen/Qwen3-VL-8B-Instruct}~\cite{bai2025qwen3} on the full
curated training set \textsc{Sureon}-DB
using text only (question and answer options, no video) for
$10$ epochs with a learning rate of $1\times10^{-5}$ on 4 NVIDIA H200
GPUs. Training for longer than in
Sec.~\ref{sec:appendix-stage3-guessers} produces a stronger adversary
that has more thoroughly absorbed any residual shortcut patterns in the
training distribution.

For each retained validation sample, we search for the distractor
configuration from the pool $\mathcal{D}_{ij}^{k,\mathrm{pool}}$
generated in Stage~2 that the hardened guesser fails to answer
correctly under blind testing. Up to $M{=}32$ candidate distractor configurations are drawn at random (seed=42) from $\mathcal{D}_{ij}^{k,\mathrm{pool}}$;
the first configuration for which the guesser fails
($S < \tau{=}3/5$) is retained as the final distractor set.
If no such configuration is found within the $M$ attempts,
the sample is discarded. Algorithm~\ref{alg:hardening} formalizes
this procedure.

\begin{algorithm}[t]
\caption{Distractor Hardening via Iterative Blind Testing}
\label{alg:hardening}
\begin{algorithmic}[1]
\Require Curated validation sample $x_{ij}^k$,
         distractor pool $\mathcal{D}_{ij}^{k,\mathrm{pool}}$,
         hardened guesser $\mathcal{G}$,
         threshold $\tau{=}3/5$,
         max attempts $M{=}32$
\Ensure  Hardened sample with final distractors, or $\varnothing$
\For{$m \leftarrow 1$ \textbf{to} $M$}
\State Sample $\mathcal{O}_{\mathrm{cand}}$ from
       $\mathcal{D}_{ij}^{k,\mathrm{pool}}$
    \State Compute $S(\mathcal{G},\, x_{ij}^k)$ using
           $\mathcal{O}_{\mathrm{cand}}$
           \hfill $\triangleright$ $R{=}5$ blind queries at $T{=}0.7$
    \If{$S(\mathcal{G},\, x_{ij}^k) < \tau$}
        \State \Return $\left(x_{ij}^k,\; \mathcal{O}_{\mathrm{cand}}\right)$
               \hfill $\triangleright$ guesser fails: configuration hardened
    \EndIf
\EndFor
\State \Return $\varnothing$
       \hfill $\triangleright$ no valid configuration found: discard
\end{algorithmic}
\end{algorithm}

\textcolor{black}{\textbf{Held-out guesser generalization.} To test whether hardening overfits to the model family used during curation, we additionally evaluate \texttt{Gemma-4-31B-IT}, which is not used at any stage of blind filtering or distractor hardening, under the same text-only condition. For the Qwen3 model trained on SUREON, blind accuracy decreases from 36.5\% on the pre-hardening validation set to 21.4\% after hardening. Gemma-4-31B-IT remains comparably low across the same sets (23.2\% to 22.9\%). Thus, while hardening primarily removes shortcuts learned by the adversary used during curation, the resulting samples remain difficult for a held-out model family, providing evidence that the procedure does not simply specialize the benchmark against Qwen-specific failure modes.}

Table~\ref{tab:curation-stats} reports per-category sample counts at
each stage of curation: training and validation after blind-testing
filtering (Sec.~\ref{sec:appendix-stage3-guessers}), and the validation
set after distractor hardening. Sequence Summarization is excluded from
curation and hardening; all SM samples are carried through unchanged
from Stage~1.

\begin{table}[t]
\centering
\caption{Per-category sample counts after Stage~3 blind-testing
         curation and validation-set distractor hardening.
         Survival rates are computed relative to the Stage~1 output.
         \textdagger~SM is excluded from curation and hardening;
         all samples are carried through unchanged.}
\label{tab:curation-stats}
\setlength{\tabcolsep}{5pt}
\resizebox{\linewidth}{!}{%
\begin{tabular}{l rr rr rr}
\toprule
 & \multicolumn{2}{c}{\textbf{Train (after curation)}}
 & \multicolumn{2}{c}{\textbf{Val (after curation)}}
 & \multicolumn{2}{c}{\textbf{Val (after hardening)}} \\
\cmidrule(lr){2-3}\cmidrule(lr){4-5}\cmidrule(lr){6-7}
\textbf{Category}
    & \textbf{Count} & \textbf{Rate}
    & \textbf{Count} & \textbf{Rate}
    & \textbf{Count} & \textbf{Rate} \\
\midrule
AD  & 14,184 & 51.1\% & 1,534 & 44.7\% & 1,102 & 32.1\% \\
EA  &  5,978 & 22.4\% &   582 & 18.1\% &   439 & 13.7\% \\
EE  &  7,816 & 33.8\% &   772 & 26.4\% &   607 & 20.8\% \\
LAR &  5,859 & 30.9\% &   570 & 25.5\% &   377 & 16.8\% \\
TO  &  7,889 & 44.8\% &   891 & 43.4\% &   641 & 31.2\% \\
PAD &  6,905 & 42.6\% &   747 & 36.8\% &   653 & 32.2\% \\
SM\textdagger
    & 14,926 & 100.0\% & 1,784 & 100.0\% & 1,784 & 100.0\% \\
IAI &  5,351 & 40.7\% &   580 & 34.6\% &   536 & 32.0\% \\
SPI &  3,870 & 40.3\% &   408 & 34.2\% &   299 & 25.1\% \\
F   &  3,283 & 38.2\% &   335 & 30.7\% &   239 & 21.9\% \\
\midrule
\textbf{Total}
    & \textbf{76,061} & \textbf{43.1\%}
    & \textbf{8,203}  & \textbf{37.9\%}
    & \textbf{6,677}  & \textbf{30.8\%} \\
\bottomrule
\end{tabular}}
\end{table}

\begin{figure}[t]
    \centering
    \includegraphics[width=\linewidth]{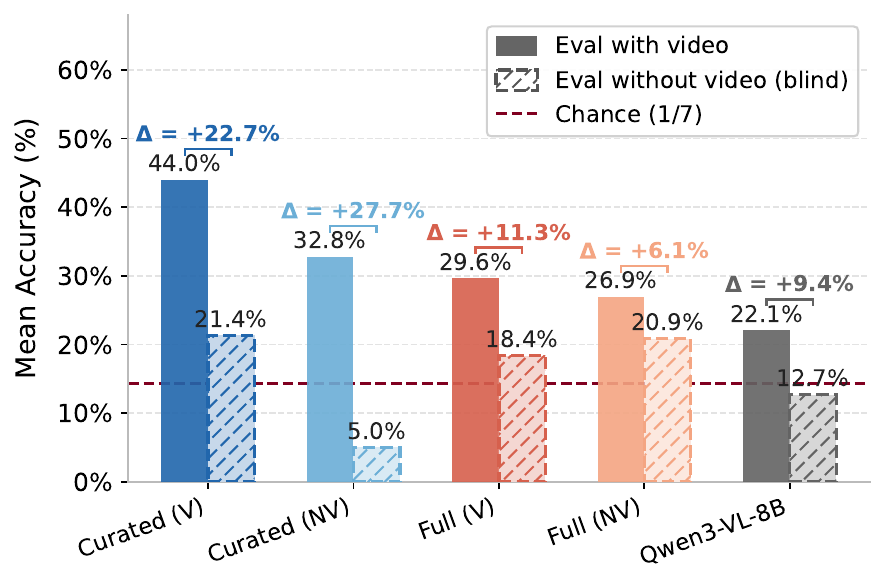}
    \caption{Training-set curation ablation. Mean accuracy on $\mathcal{D}_{\mathrm{val}}^{\star}$ validation set under evaluation \textit{with} video
             (solid bars) and \textit{blind}, without video (hatched
             bars), for models trained on the full uncurated set
             $\mathcal{D}_{\mathrm{ref}}$ (\textit{Full}) vs. the
             curated set \textsc{Sureon}-DB
             (\textit{Curated}), together with the off-the-shelf
             \texttt{Qwen3-VL-8B} baseline. (V) and (NV) denote video-trained vs. text-only–trained}
    \label{fig:stage3-ablation}
\end{figure}

\noindent \textbf{Training set curation ablation.}
To validate that curating the training set encourages models to
ground predictions in visual evidence, we compare two training
configurations: a model trained on the full uncurated set
$\mathcal{D}_{\mathrm{ref}}$ and a model trained on the curated set
\textsc{Sureon}-DB, both evaluated on
the hardened validation set. Training on the curated set is expected
to yield higher vision-enabled accuracy, as the model is no longer
rewarded for exploiting language shortcuts during training.
 Figure~\ref{fig:stage3-ablation} reports vision and blind accuracy under both training conditions.

\subsection{Stage 4: Expert Verification Details}
\label{sec:appendix-stage4}

The hardened validation set $\mathcal{D}_{\mathrm{val}}^{\star}$ produced
by Stage~3 is submitted to manual verification by expert surgeons; \textcolor{black}{no additional LLM-based filtering or adjudication is applied at this stage.} Only samples that pass expert review are retained in the final
\textsc{Sureon} benchmark $\mathcal{B}_{\mathrm{\textsc{Sureon}}}$.

\subsubsection{Annotation Platform}
\label{sec:appendix-platform}

\begin{figure}[t]
    \centering
    \includegraphics[width=\linewidth]{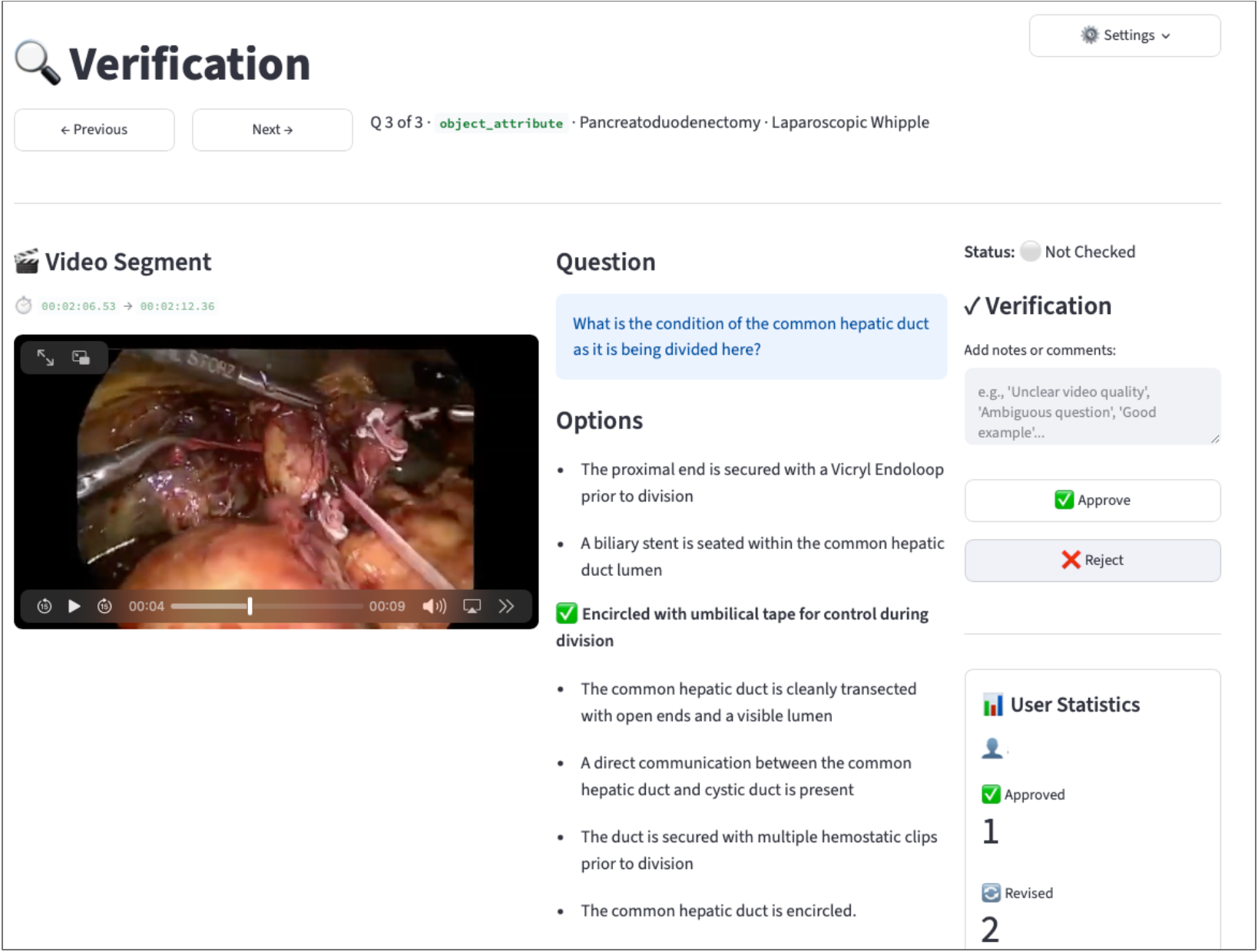}
    \caption{Main verification screen of the annotation platform.
             Reviewers are presented with the video clip, the question,
             all answer options (correct answer marked in green), and
             the approve/reject interface.}
    \label{fig:platform-verification}
\end{figure}

\begin{figure}[t]
    \centering
    \includegraphics[width=\linewidth]{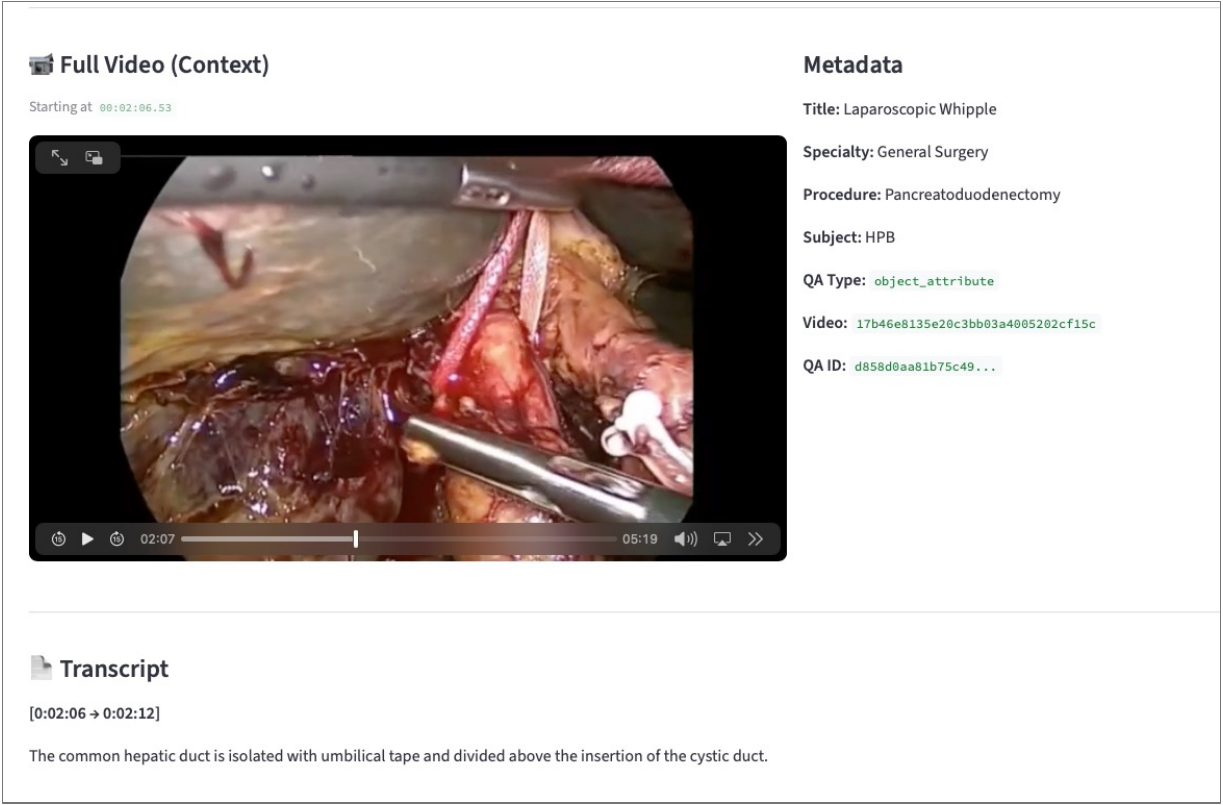}
    \caption{Extended context view, accessible by scrolling below the
             verification screen. Provides the full surgery video,
             aligned transcript, and video-level metadata.}
    \label{fig:platform-context}
\end{figure}

Verification is conducted on a dedicated web-based annotation platform.
Figure~\ref{fig:platform-verification} shows the main verification
screen, which presents: \textit{(i)} the video clip and its timestamp,
\textit{(ii)} the question with all answer options (the correct answer
marked in green), and \textit{(iii)} an approve/reject interface
requiring a written note for every rejection.
Figure~\ref{fig:platform-context} shows the extended context view,
available by scrolling below the verification screen, which provides
the full surgery video (starting at the clip's timestamp), the aligned
transcript segment, and video-level metadata including title, specialty,
procedure, and QA type. Reviewers are encouraged to consult the full
video whenever the short clip is insufficient to make a judgement.
The platform also exposes specialty, procedure, and QA type filters,
allowing reviewers to focus on samples within their clinical domain.

\subsubsection{Reviewer Panel and Protocol}
\label{sec:appendix-reviewers}
Verification is conducted by 6 surgeons affiliated to different institutions. 

\noindent \textbf{Sample assignment.}
Each sample is assigned to a single reviewer, drawn uniformly at random from the pool of available samples across the different question types, and
reviewers may skip any sample they consider outside their domain. A single reviewer per sample is sufficient because reviewers do not annotate from
scratch: each candidate QA is already transcript-grounded, with its correct answer derived from a surgeon's narration (Sec.~\ref{sec:appendix-stage1}).
Verification is therefore a confirmation step rather than an open-ended labeling task, in which the reviewer checks video--transcript alignment, the
clinical correctness of the narration-derived answer, and distractor validity for the specific clip. Because the underlying supervision already originates
from a domain expert, a single qualified reviewer suffices to confirm these conditions, and we apply a
conservative retention policy under which any rejected or flagged sample is discarded, so that residual subjectivity removes samples rather than admitting
incorrect ones.

\textcolor{black}{\noindent\textbf{Inter-rater reliability.} Although each benchmark candidate was reviewed by one surgeon in the primary verification protocol, we independently dual-reviewed a subset of 74 QA samples distributed among QA types to estimate inter-rater agreement. Each sample was independently reviewed by two raters drawn from a pool of five surgeons, with reviewer pairs assigned to ensure balanced coverage across all C(5,2)=10 possible pairings. The two reviewers independently rendered the same binary approve/reject decision and agreed on 52 of 74 samples (70.3\% agreement). The duplicate ratings were used only to estimate agreement and did not alter the primary single-review retention decision.}

\noindent \textbf{Review criteria.}
Each reviewer watches the clip and assesses the following four conditions, all of which must hold simultaneously for a sample to be approved:
\begin{itemize}[leftmargin=1.5em]
    \item \textit{Question soundness.} The question is clear,
    clinically meaningful, and answerable from the specific clip or segment provided.
    \item \textit{Answer correctness.} The marked answer is
    clinically correct and is visible or directly supported within the clip's time window.
    \item \textit{Video--transcript alignment.} The QA reflects what is actually shown in the clip, and the clip depicts a real operative scene rather than, e.g., a slide or diagram.
    \item \textit{Distractor validity.} All distractors are
    clinically plausible for the context but unambiguously
    incorrect for this specific clip. Any sample for which a
    second option could also be correct is rejected.
\end{itemize}

\noindent Reviewers are additionally asked to confirm that the question \textit{requires} the video to answer and cannot be resolved from generic procedural knowledge alone, serving as a human analog of the blind-testing filter applied in Stage~3. We encourage rejected samples to  include a brief written note identifying the reason for rejection.

For Forecasting questions, the correct answer describes the step that occurs \textit{after} the clip ends and therefore cannot be verified from the clip alone. Reviewers are instructed to note the clip's end timestamp, open the full video, and confirm that the marked next step actually occurs immediately after the clip before approving.

\noindent \textbf{Quantitative summary.}
Table~\ref{tab:stage4-approval} reports the number of reviewed, approved, and rejected samples per category, along with the approval rate. \textcolor{black}{The experts provided rejection reasons for 178 and the breakdown is distributed as follows: 22.5\% flagged as ambiguous/multiple-correct, 14.6\% inaccurate, 9.6\% too-short or insufficient context, and 9.6\% not visible in the video.} The final benchmark $\mathcal{B}_{\mathrm{\textsc{Sureon}}}$ comprises all approved samples across categories. \textcolor{black}{Expert verification required approximately 32 hours of cumulative surgeon time across all reviewers.}

\begin{table}[t]
\centering
\caption{Expert verification results per category.
         \textit{Reviewed}: samples assessed.
         \textit{Approved}: retained in $\mathcal{B}_{\mathrm{\textsc{Sureon}}}$.
         \textit{Rejected}: samples that failed review.
         \textit{Rate}: approved / reviewed.}
\label{tab:stage4-approval}
\setlength{\tabcolsep}{5pt}
\resizebox{\columnwidth}{!}{%
\begin{tabular}{lrrrr}
\toprule
\textbf{Category}
    & \textbf{Reviewed}
    & \textbf{Approved}
    & \textbf{Rejected}
    & \textbf{Rate} \\
\midrule
AD  & 62 & 35 &  27 & 56.5\% \\
EA  & 63 & 34 & 29 & 54.0\% \\
EE  & 55 & 37 &  18 & 67.3\% \\
LAR & 59 & 37 &  22 & 62.7\% \\
TO  & 69 & 34 &  35 & 49.3\% \\
PAD & 65 & 29 &  36 & 44.6\% \\
SM  & 51 & 41 & 10 & 80.4\% \\
IAI & 62 & 31 & 31 & 50.0\% \\
SPI & 53 & 31 & 22 & 58.5\% \\
F   & 63 & 31 &  32 & 49.2\% \\
\midrule
\textbf{Total}
    & \textbf{602} & \textbf{340} & \textbf{262} & \textbf{56.5\%} \\
\bottomrule
\end{tabular}}
\end{table}

\subsection{Final Dataset Configuration}
\label{sec:appendix-final-config}
\begin{figure}[t]
    \centering
    \includegraphics[width=0.7\linewidth]{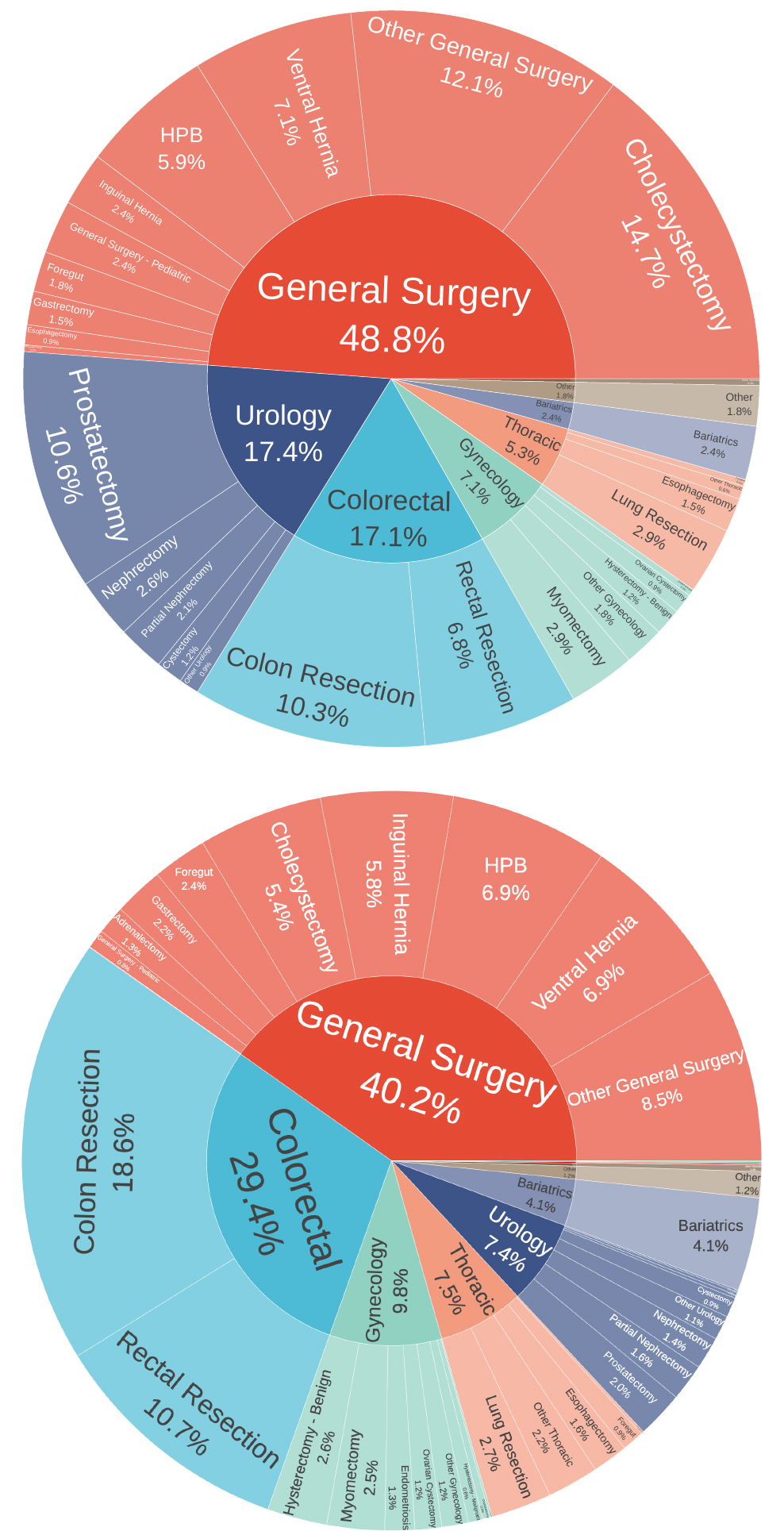}
    \caption{\textcolor{black}{Sunburst visualizations of the \textsc{Sureon} benchmark
(top) and \textsc{Sureon}-DB (bottom), showing the hierarchical distribution of videos across surgical specialties (inner ring) and subjects (outer ring)}}
    \label{fig:specialty-dist}
\end{figure}

We consolidate the configuration retained after all four stages; per-stage counts are in Tab.~\ref{tab:curation-stats} and key attributes in Tab.~\ref{tab:sureon_stats}. The pipeline yields two resources from a common process. \textsc{Sureon}-DB is the curated training database: of the 2,211 training videos with validated Stage-1 QA, 2,191 retain at least one sample after blind-testing curation (Sec.~\ref{sec:appendix-stage3}), giving $76{,}061$ QA pairs. The \textsc{Sureon} benchmark is the expert-verified set: the hardened validation set $\mathcal{D}^{\star}_{\mathrm{val}}$ is reviewed by surgeons (Sec.~\ref{sec:appendix-stage4}), and the $340$ approved samples, from $165$ distinct source videos, form $\mathcal{B}_{\mathrm{\textsc{Sureon}}}$. The two splits share no source videos, as splitting is video-level prior to generation (Sec.~\ref{sec:appendix-dataset-construction}). \textcolor{black}{Figure~\ref{fig:specialty-dist} shows the hierarchical distribution of both resources across surgical specialties and subjects.}

\noindent\textbf{Distractor configuration.} Each multiple-choice question is paired with six distractors via LLM-Mix: the three generators ($M_1$, $M_2$, $M_3$) each emit six candidates, and the final six are drawn from the $18$-candidate pool. We adopt this six-distractor LLM-Mix setting because it most consistently drives blind-tested accuracy toward chance across both off-the-shelf and fine-tuned conditions (Tab.~\ref{tab:distractor}). In \textsc{Sureon}-DB these are the curated Stage-2 sets; in the benchmark they are the configurations selected by iterative hardening (Sec.~\ref{sec:appendix-stage3-hardening}). Sequence Summarization is open-ended and carries no distractors.

\section{Training Data Construction from public dataset sources}
\label{sec:appendix-model-data}

In the main paper (Sec.~\ref{sec:model}) we consolidate \textsc{Sureon} with supervision derived from existing public surgical datasets to study two questions: whether the standard perception-centric training recipe transfers to reasoning (Q1), and whether combining both sources yields a
single model that is strong at perception \emph{and} reasoning (Q2). This appendix details how we transform that public data --- denoted $\mathcal{D}_{\mathrm{pub}}$ in the main text --- into instruction-tuning QA pairs that share \textsc{Sureon}'s format. We deliberately follow a similar recipe used by prior surgical assistants~\cite{bellos2026surgatlas,zeng2025surgvlm}: (1)~unification of heterogeneous annotations into a common schema, (2)~rendering of those annotations into natural-language QA pairs via task-specific templates,
and (3)~LLM-based linguistic augmentation. After stage~1 the pipeline diverges into an \emph{image} branch (frame-level perception tasks) and a
\emph{clip} branch (workflow/temporal-segmentation tasks). Combined with \textsc{Sureon}-DB, the result is the
consolidated corpus
$\mathcal{D}_{\mathrm{all}} = \mathcal{D}_{\text{\textsc{Sureon}-DB}} \cup \mathcal{D}_{\mathrm{pub}}$..

\subsection{Stage 1: Common Annotation Schema}
\label{sec:appendix-schema}

Each public dataset ships with its own annotation format, label
vocabulary, and temporal representation. We normalize all datasets into a
single relational schema that captures: dataset-level capability flags
(\textit{has\_phase}, \textit{has\_tool\_bbox}, \textit{has\_triplets},
etc.); video metadata (split, frame rate, total frames); per-frame
classification labels (phase, step, action indices, tool/anatomy presence
lists); per-frame object instances (class, bounding box in, optional segmentation mask stored as a COCO RLE); temporal segments (start/end frame, segment
type); and inter-object relations (instrument--verb--target triplets).

\noindent\textbf{Canonical vocabulary.}
Dataset-specific label names are normalized to a shared canonical vocabulary. For the image branch this mapping is stored in a flat reverse-lookup dictionary covering
tools, anatomical structures, surgical phases, procedural steps, and
action verbs (e.g., CalotTriangleDissection to 
calot triangle dissection; hook to
electrocautery hook). For the clip branch an equivalent
per-dataset mapping handles the same normalization, and each dataset is further annotated with its procedure type.

\noindent\textbf{Source datasets.}
The image branch covers 17 source datasets
(CholecT50~\cite{nwoye2021rendezvous},
CholecInstanceSeg~\cite{alabi2025cholecinstanceseg},
CholecSeg8k~\cite{hong2020cholecseg8k}, GraSP~\cite{ayobi2025pixel},
EndoVis2017~\cite{allan20192017}, EndoVis2018~\cite{allan20202018},
Endoscapes~\cite{murali2023latent}, AutoLaparo~\cite{wang2022autolaparo},
LapGyn~\cite{leibetseder2018lapgyn4}, Phakir~\cite{rueckert2026comparative},
HeiChole~\cite{wagner2023comparative},
HeiCo~\cite{maier2021heidelberg},
GynSurg~\cite{nasirihaghighi2025gynsurg}, Dresden~\cite{carstens2023dresden},
HemoSet~\cite{miao2024hemoset}),
providing tool presence and localization, anatomy localization, instrument--verb--target triplets, Critical View of Safety (CVS), and
frame-level phase/action labels. The clip branch covers 13 source datasets
(Cholec80~\cite{twinanda2016endonet}, AutoLaparo, GraSP, GynSurg, HeiChole,
HeiCo-Proctocolectomy, HeiCo-Rectal Resection, HeiCo-Sigmoid Resection,
MultiBypass~\cite{Lavanchy2024},
Phakir, SARRARP50~\cite{psychogyios2023sar},
SurgicalActions160~\cite{surgicalactions160}),
providing continuous temporal annotations for phases, steps, actions, and tool presence. Eight datasets contribute to both branches, for 22 distinct
sources in total; Table~\ref{tab:public-datasets} lists each dataset, the
branches it feeds, and the resulting sample counts.


\begin{table}[t]
\centering
\caption{Public surgical datasets transformed into
         $\mathcal{D}_{\mathrm{pub}}$. \textit{Branch} indicates whether a
         dataset contributes frame-level (\textit{Img}), clip-level
         (\textit{Clip}), or both (\textit{Both}) supervision; counts are
         the number of source items (frames/clips) and the QA pairs
         produced after templating and augmentation.}
\label{tab:public-datasets}
\setlength{\tabcolsep}{6pt}
\resizebox{\linewidth}{!}{%
\begin{tabular}{lcrr}
\toprule
\textbf{Dataset} & \textbf{Branch} & \textbf{\#Items} & \textbf{\#QA} \\
\midrule
CholecT50~\cite{nwoye2021rendezvous}                & \multirow{9}{*}{Img}  &  72{,}815 & 1{,}919{,}383 \\
CholecInstanceSeg~\cite{alabi2025cholecinstanceseg} &                       &  26{,}830 &   834{,}398 \\
CholecSeg8k~\cite{hong2020cholecseg8k}              &                       &   5{,}600 &   321{,}268 \\
EndoVis2017~\cite{allan20192017}                    &                       &   1{,}800 &   108{,}282 \\
EndoVis2018~\cite{allan20202018}                    &                       &   1{,}639 &   126{,}343 \\
Endoscapes~\cite{murali2023latent}                  &                       &   6{,}960 &   124{,}656 \\
LapGyn~\cite{leibetseder2018lapgyn4}                &                       &  53{,}107 &   225{,}350 \\
Dresden~\cite{carstens2023dresden}                  &                       &   8{,}050 &    55{,}568 \\
HemoSet~\cite{miao2024hemoset}                      &                       &     828 &     8{,}280 \\
\midrule
GraSP~\cite{ayobi2025pixel}                         & \multirow{8}{*}{Both} &   4{,}340 &   293{,}736 \\
AutoLaparo~\cite{wang2022autolaparo}                &                       &   1{,}336 &    82{,}524 \\
GynSurg~\cite{nasirihaghighi2025gynsurg}            &                       &   4{,}313 &    62{,}703 \\
HeiChole~\cite{wagner2023comparative}               &                       &   2{,}064 &    57{,}405 \\
HeiCo - Proctocolectomy~\cite{maier2021heidelberg}              &                       &   1{,}681 &    40{,}281 \\
HeiCo - Rectal Resection ~\cite{maier2021heidelberg}              &                       &   2{,}293 &    52{,}089 \\
HeiCo Sigmoid Resection ~\cite{maier2021heidelberg}             &                       &   1{,}316 &    31{,}656 \\
Phakir~\cite{rueckert2026comparative}               &                       &  14{,}447 &   499{,}099 \\
\midrule
Cholec80~\cite{twinanda2016endonet}                 & \multirow{5}{*}{Clip} &   3{,}834 &    85{,}799 \\
MultiBypass-Bern~\cite{Lavanchy2024}                &                       &   3{,}598 &   174{,}270 \\
MultiBypass-Strasbourg~\cite{Lavanchy2024}          &                       &   3{,}907 &   226{,}995 \\
SARRARP50~\cite{psychogyios2023sar}                 &                       &   2{,}676 &   171{,}886 \\
SurgicalActions160~\cite{surgicalactions160}        &                       &     157 &     1{,}260 \\
\midrule
\textbf{Total (22 datasets)}                        &                       & \textbf{223{,}591} & \textbf{5{,}503{,}231} \\
\bottomrule
\end{tabular}}
\end{table}

\subsection{Stage 2a: Image-Level QA Template Generation}
\label{sec:appendix-image-templates}

\noindent\textbf{Segmentation-to-instance conversion.}
Several datasets provide only semantic segmentation masks. For each class we apply a 
conversion policy: \emph{single-instance} classes (anatomical structures
known to appear once, such as the gallbladder or uterus) are treated as a
single pseudo-instance by taking the union of all connected components, then fitting a tight bounding box;
\emph{connected-components} classes (blood pools, clips) are split into individual pseudo-instances via 8-connected-component labeling, sorted
top-to-bottom then left-to-right by centroid; and \emph{strict-semantic}
classes (large background structures such as the abdominal wall or liver
bed) are retained as presence-only annotations without bounding-box
assignment. All bounding boxes are normalized to a $[0,1000]$ coordinate
space and stored in $[x_1, y_1, x_2, y_2]$ format.

\noindent\textbf{QA types.}
Starting from the normalized frame annotations, we instantiate task-specific QA templates across two families. \emph{Open-ended
templates} produce free-text answers and cover: tool/anatomy presence (``Which surgical instruments are visible?''); counting (``How many surgical instruments are visible?''); view-region localization (``Where is the \{tool\} with respect to
the surgical field?''); action presence (``What action is happening in this frame?''; triplet recognition (``What is the \{tool\}
doing?''); Critical
View of Safety (three binary criteria for safe cholecystectomy); mask classification (identify structures from rendered contour
overlays); and procedure-type recognition. Binary and negative variants are generated by sampling negative class labels from the dataset
vocabulary.

\noindent\textbf{Grounded templates.}
Grounded templates instead require structured JSON answers, which lets us
pose localization and recognition tasks with machine-checkable outputs.
They span localization (returning a JSON array of labeled, indexed
bounding boxes for the visible tools or anatomy), presence (a dictionary
flagging each class as present or absent), counting (the number of tools
in view), phase and step recognition (the active phase or step), triplet
recognition (the active instrument--verb--target relations), and action
recognition (the action performed by each tool in view). In every grounded
template the question explicitly enumerates the full class vocabulary as
the option set, anchoring the model to a closed label space. Each QA pair
is serialized as a JSONL record with the question, answer, an optional
reasoning trace (omitted for non-CoT types), and provenance fields
identifying the QA type, source dataset, file, video, and frame.

\subsection{Stage 2b: Video Clip QA Template Generation}
\label{sec:appendix-clip-templates}

\noindent\textbf{Clip sampling.}
For temporal datasets we sample variable-length clips with two strategies.
By default, \emph{strategic sampling} biases selection toward phase and
step transition boundaries and stable mid-phase regions, so that clips
capture both the moments of change and the steady states between them.
Alternatively, \emph{grounded clip sampling} slides a fixed 8-frame window
with a stride of four and keeps a clip only if it lies entirely within a
single constant-label segment across all requested tasks (phase, step,
action), guaranteeing an unambiguous answer.

\noindent\textbf{Language templates.}
For each sampled clip, a clip-description generator draws at random from several templates and fills them with the clip's annotations. Phase and step templates describe which phase or step the clip shows, and
optionally enrich this with the procedure type, the temporal position within the phase (beginning, middle, or end), or both; clips that span a
boundary instead use transition templates that name the two phases and the approximate time of the switch, the procedure transitions from \{phase$_1$\} to \{phase$_2$\}''). Step templates extend
this to single-, two-, and multi-step sequences, action templates to the
actions present, and tool templates to which instruments appear and when they are used (e.g., ``\{tool\} is used from \{start\}~s to \{end\}~s'').

\noindent\textbf{Question types.}
The resulting questions are descriptive (``What is happening in this
clip?''), binary (``Is the \{phase\} phase shown in this clip?''),
temporal (``When does \{event\} occur in this clip?''), or counting (``How
many distinct phases are shown?''). As in the image branch, grounded clip
questions take a structured JSON answer (e.g., the active phase). For each
dataset we keep three versions of every QA pair --- the direct template
output, an instruction-style rephrasing, and an LLM-augmented variant
(Sec.~\ref{sec:appendix-llm-aug}) --- which are pooled at training time.

\subsection{Stage 3: LLM-Based Linguistic Augmentation}
\label{sec:appendix-llm-aug}
To increase linguistic diversity without altering factual content, we
paraphrase each templated QA pair with an Qwen3.5-9B~\cite{team2026qwen3}, adding one
rephrased variant per pair to the training pool.

\noindent\textbf{Image branch.}
We use Qwen3.5-9B~\cite{team2026qwen3} 
($T{=}0.7$, $p{=}0.8$, $k{=}20$, up to 256 new tokens) to rephrase the
question and answer under three constraints: surgical terminology is
preserved exactly (\emph{phase}, \emph{step}, \emph{procedure}, and
\emph{action} are never interchanged), the question retains explicit
visual-grounding cues (e.g., ``in this image'', ``visible in the frame''),
and any ordered or list structure in the original is kept intact. For
tasks with structured answers (localization, presence, triplet, and mask
classification), only the question is paraphrased and the answer is copied
verbatim, so it remains valid and label-consistent. For chain-of-thought
samples, the accompanying reasoning is paraphrased alongside the question
and answer.

\noindent\textbf{Clip branch.}
We use Qwen3-VL-8B-Instruct~\cite{bai2025qwen3} in text-only mode
($T{=}0.7$, $p{=}0.9$, up to 256 new tokens) with an analogous prompt that
additionally requires each question to refer to a video clip or segment
and, as above, keeps \emph{phase}, \emph{step}, and \emph{action} distinct.

\section{Evaluation}
\label{sec:appendix-evaluation}

\begin{figure}[t]
    \centering
    \includegraphics[width=\linewidth]{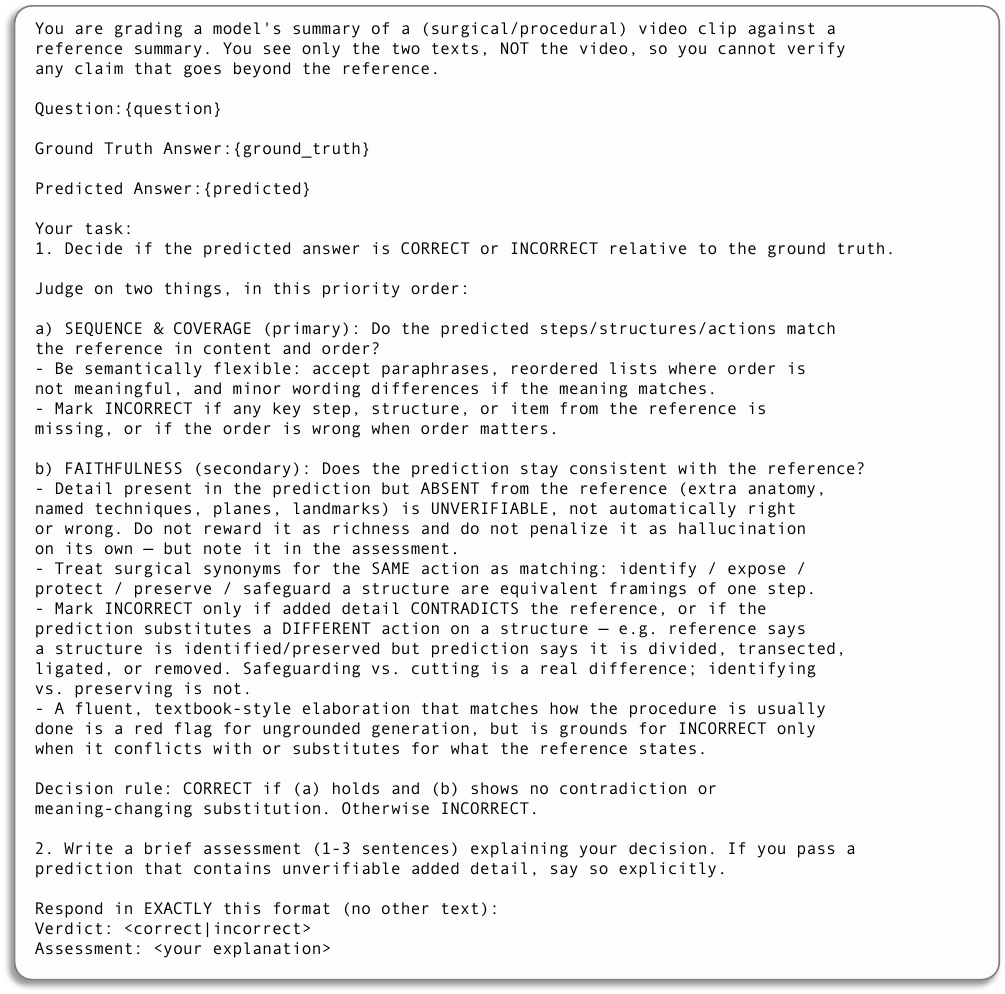}
    \caption{LLM judge prompt to evaluate answers to the open-ended sequence summarization question type.}
    \label{fig:llmjudge-prompt}
\end{figure}

This section specifies the inference and scoring protocols used in our experiments (Sec.~\ref{sec:experiments}), covering the \textsc{Sureon} benchmark (Tab.~\ref{tab:generalization}) and the public perception benchmarks for recognition and spatial grounding (Tab.~\ref{tab:public-benchmarks}, Tab.~\ref{tab:training-ablation}). All close-ended questions and Temporal Ordering are evaluated under accuracy. The open-ended Sequence Summarization (SM) type is scored by an LLM judge (Claude Opus 4.8) using the prompt in Fig.~\ref{fig:llmjudge-prompt}. We organize the protocols into evaluation on \textsc{Sureon} (Sec.~\ref{sec:appendix-eval-sureon}) and on public benchmarks (Sec.~\ref{sec:appendix-eval-public}).

\subsection{SUREON Benchmark Evaluation}
\label{sec:appendix-eval-sureon}

\begin{figure}[t]
    \centering
    \includegraphics[width=\linewidth]{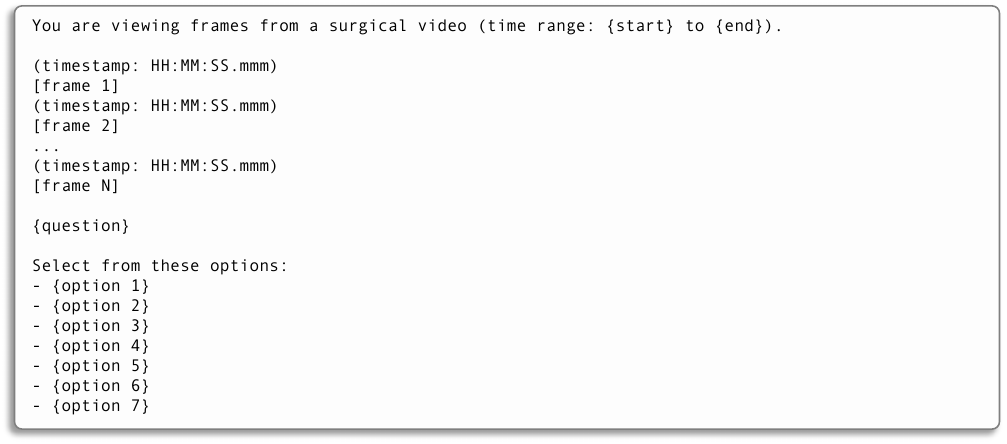}
    \caption{Prompt template used to query all models on the \textsc{Sureon} benchmark. Frames are interleaved with their timestamps, followed by the question and seven answer options.}
    \label{fig:sureon-prompt}
\end{figure}

\noindent \textcolor{black}{ \textbf{Frame sampling and inference.} Every model receives the same visual input for a given question. For each clip we uniformly sample up to 8 frames across its duration at the video's frame rate, clamped to a minimum of 4 frames. Before inference, we apply a two-stage blurring pass to the sampled frames: a YOLO-based non-surgical object detector identifies non-operative regions (faces, equipment logos, and other non-surgical content) trained for YouTube videos~\cite{che2025lemon}, and optical character recognition detects on-screen textual overlays such as procedure labels, instrument names in the user interface, and equipment text; both sets of detected regions are blurred. Surgical academic videos can contain these overlays, which modern VLMs may exploit through OCR to infer answers without interpreting the underlying surgical scene; blurring them prevents this source of leakage and better isolates visual surgical understanding. Each frame in the prompt is tagged with its absolute timestamp, preceded by the clip's time range, followed by the question and the option list (prompt template in Fig.~\ref{fig:sureon-prompt}). The question text, option set, and option order are held fixed across all evaluated models.}

\textbf{Option format.} Multiple-choice questions are presented with seven options giving a random baseline of $1/7 = 14.3\%$. Temporal Ordering (TO) requires ordering three events, for which the random baseline over the $3! = 6$ permutations is $1/6 = 16.7\%$. Distractor selection and option shuffling are seeded per sample for reproducibility.

\textbf{Blind condition and multimodal gain.} The blind condition removes the visual input entirely: the model receives a text-only payload containing the identical question and option list, with no frames. We report multimodal gain $\mathrm{MG} = S_v - S_{wv}$ as the accuracy difference between the visual ($S_v$) and blind ($S_{wv}$) passes.

\textbf{Scoring.} The eight multiple-choice types are scored by exact-match accuracy against the correct option. For TO, we extract the predicted sequence of event labels from the free-form output and require exact, order-sensitive equality with the reference sequence. SM is scored by the LLM judge described below. Reported $\mu$ is the pooled accuracy over the 299 accuracy-scored items; SM is reported separately and excluded from $\mu$.

\textcolor{black}{\textbf{Human baseline protocol.}
The human baseline is collected from 5 surgeons, a subset of the 6-surgeon panel used for Stage 4 expert verification; Sequence
Summarization is excluded from the baseline due to the time cost of open-ended grading, leaving 299 accuracy-scored items. Baseline collection uses a separate interface from the verification platform: each surgeon is shown only the video clip, the question, and the answer options (no blind testing mode for expert baseline). Items are distributed uniformly at random across QA types for each surgeon, without matching raters to their own specialty, as our reviewer pool is not large enough to guarantee specialty coverage per item. Each item is answered by a single surgeon, and the reported score aggregates across raters. To avoid trivial recall, no surgeon is shown a question they had already adjudicated during expert verification. After each answer, surgeons self-report confidence (Confident / Somewhat / Not confident); across recorded responses the breakdown is 31.4\% / 37.5\% / 17.7\%, with 13.4\% not recorded. Total collection required approximately 8.5 hours of cumulative surgeon time. Several surgeons reported difficulty answering questions outside their own specialty, consistent with the per-category gap between near-ceiling performance on Procedure-Agnostic Action Description and comparatively weaker performance
on  categories that require specialty-specific understanding, which also motivates a surgical VLM capable of breadth across specialties, which no single human specialist reviewer possesses.}

\textcolor{black}{
\textbf{Uncertainty and statistical comparisons.} We quantify
uncertainty in the \textsc{Sureon} results using 95\% confidence intervals. For
the overall accuracy, we verify our non-parametric bootstrap
intervals (10{,}000 resamples over benchmark items) against the
analytic Wilson score interval. The two estimates closely agree, with a mean absolute discrepancy of 0.25 percentage points and a maximum of 0.54 points across all model estimates. We additionally bootstrap accuracy separately for each question category and for the perception and reasoning aggregates, and we report a surgeon baseline collected
under the same video-based protocol as the model evaluations. Because model (and surgeon) predictions are paired on the same benchmark samples, pairwise differences are tested using McNemar's test with Holm--Bonferroni correction across all $\binom{12}{2}=66$ model pairs (11 evaluated models and the human baseline). Tables~\ref{tab:overall_bootstrap_ci}, \ref{tab:type_bootstrap_ci}, \ref{tab:category_ci_perception}, and \ref{tab:category_ci_reasoning} report the corresponding confidence intervals; Table~\ref{tab:pairwise_sig} reports the full pairwise significance matrix underlying the claims above.
}
 
\begin{table}[t]
\centering
\caption{Overall accuracy and bootstrap 95\% confidence intervals on
the 299 accuracy-scored \textsc{Sureon} benchmark samples.}
\label{tab:overall_bootstrap_ci}
\small
\resizebox{0.45\textwidth}{!}{
\begin{tabular}{lccc}
\toprule
Model & $n$ & Accuracy (\%) & Bootstrap 95\% CI (\%) \\
\midrule
Human                                          & 299 & 64.9 & [59.5, 70.2] \\
Gemini 3.1 Pro                                 & 299 & 52.2 & [46.8, 57.9] \\
\textsc{Sureon}-VLM                            & 299 & 48.2 & [42.5, 53.8] \\
GPT-5.1 (high)                                 & 299 & 46.8 & [41.1, 52.5] \\
\textsc{Sureon}-VLM (\textsc{Sureon}-only)     & 299 & 44.1 & [38.5, 49.8] \\
Gemma 4                                        & 299 & 35.8 & [30.4, 41.1] \\
Qwen3VL-32B-T                                  & 299 & 34.1 & [28.8, 39.5] \\
Qwen3VL-8B-I                                   & 299 & 23.1 & [18.4, 27.8] \\
\textsc{Sureon}-VLM (Public-only)              & 299 & 20.7 & [16.4, 25.4] \\
MedGRPO                                        & 299 & 16.7 & [12.7, 21.1] \\
Molmo2                                         & 299 & 15.0 & [11.0, 19.1] \\
MedGemma 1.5                                   & 299 & 10.7 & [7.4, 14.4] \\
\bottomrule
\end{tabular}}
\end{table}
 
\begin{table}[t]
\centering
\caption{Accuracy and bootstrap 95\% confidence intervals for the perception and reasoning subsets of \textsc{Sureon}. Perception contains $n=166$ samples and reasoning contains $n=133$.}
\label{tab:type_bootstrap_ci}
\small
\resizebox{0.45\textwidth}{!}{
\begin{tabular}{lcc}
\toprule
Model & Perception (\%) & Reasoning (\%) \\
\midrule
Human
    & 72.9 [66.3, 79.5]
    & 54.9 [46.6, 63.2] \\
Gemini 3.1 Pro
    & 53.0 [45.2, 60.8]
    & 51.1 [42.9, 59.4] \\
\textsc{Sureon}-VLM
    & 48.8 [41.6, 56.0]
    & 47.4 [39.1, 55.6] \\
GPT-5.1 (high)
    & 50.0 [42.2, 57.8]
    & 42.9 [34.6, 51.1] \\
\textsc{Sureon}-VLM (\textsc{Sureon}-only)
    & 42.8 [35.5, 50.6]
    & 45.9 [37.6, 54.1] \\
Gemma 4
    & 39.8 [32.5, 47.0]
    & 30.8 [23.3, 39.1] \\
Qwen3VL-32B-T
    & 42.2 [34.3, 49.4]
    & 24.1 [17.3, 31.6] \\
Qwen3VL-8B-I
    & 26.5 [19.9, 33.1]
    & 18.8 [12.0, 25.6] \\
\textsc{Sureon}-VLM (Public-only)
    & 27.1 [20.5, 33.7]
    & 12.8 [7.5, 18.8] \\
MedGRPO
    & 19.3 [13.2, 25.3]
    & 13.5 [8.3, 19.6] \\
Molmo2
    & 20.5 [14.5, 27.1]
    & 8.3 [3.8, 13.5] \\
MedGemma 1.5
    & 12.0 [7.2, 17.5]
    & 9.0 [4.5, 14.3] \\
\bottomrule
\end{tabular}}
\end{table}

\begin{table*}[t]
\centering
\caption{Per-category accuracy and bootstrap 95\% confidence intervals for perception-oriented \textsc{Sureon} question types. Entries report accuracy [95\% CI], in percent. AD: Action Description; IAI: Instrument--Action Interaction; EA: Entity (Object) Attribute; EE: Entity (Object) Existence; PAD: Procedure-Agnostic Action Description.}
\label{tab:category_ci_perception}
\scriptsize
\resizebox{\textwidth}{!}{
\begin{tabular}{lccccc}
\toprule
Model & AD ($n=35$) & IAI ($n=31$) & EA ($n=34$) & EE ($n=37$) & PAD ($n=29$) \\
\midrule
Human
& 74.3 [60.0, 88.6]
& 74.2 [58.1, 87.1]
& 73.5 [58.8, 88.2]
& 51.3 [35.1, 67.6]
& 96.5 [89.7, 100.0] \\
 
Gemini 3.1 Pro
& 54.3 [37.1, 71.4]
& 54.8 [38.7, 71.0]
& 41.2 [23.5, 58.8]
& 48.6 [32.4, 64.9]
& 69.0 [51.7, 86.2] \\
 
\textsc{Sureon}-VLM
& 65.7 [48.6, 80.0]
& 38.7 [22.6, 54.8]
& 47.1 [29.4, 64.7]
& 48.6 [32.4, 64.9]
& 41.4 [24.1, 58.6] \\
 
GPT-5.1 (high)
& 57.1 [40.0, 74.3]
& 61.3 [45.2, 77.4]
& 38.2 [23.5, 55.9]
& 54.0 [37.8, 70.3]
& 37.9 [20.7, 55.2] \\
 
\textsc{Sureon}-VLM (\textsc{Sureon}-only)
& 51.4 [34.3, 68.6]
& 25.8 [9.7, 41.9]
& 44.1 [26.5, 61.8]
& 48.6 [32.4, 64.9]
& 41.4 [24.1, 58.6] \\
 
Gemma 4
& 48.6 [31.4, 65.7]
& 41.9 [25.8, 58.1]
& 32.4 [17.6, 50.0]
& 27.0 [13.5, 40.5]
& 51.7 [34.5, 69.0] \\
 
Qwen3VL-32B-T
& 40.0 [22.9, 57.1]
& 45.2 [29.0, 61.3]
& 41.2 [23.5, 58.8]
& 27.0 [13.5, 43.2]
& 62.1 [44.8, 79.3] \\
 
Qwen3VL-8B-I
& 28.6 [14.3, 42.9]
& 38.7 [22.6, 54.8]
& 17.6 [5.9, 32.4]
& 16.2 [5.4, 29.7]
& 34.5 [17.2, 51.7] \\
 
\textsc{Sureon}-VLM (Public-only)
& 42.9 [25.7, 60.0]
& 29.0 [12.9, 45.2]
& 14.7 [2.9, 26.5]
& 16.2 [5.4, 29.7]
& 34.5 [17.2, 51.7] \\
 
MedGRPO
& 25.7 [11.4, 40.0]
& 25.8 [12.9, 41.9]
& 17.6 [5.9, 32.4]
& 16.2 [5.4, 29.7]
& 10.3 [0.0, 24.1] \\
 
Molmo2
& 31.4 [17.1, 48.6]
& 16.1 [3.2, 29.0]
& 20.6 [8.8, 35.3]
& 10.8 [2.7, 21.6]
& 24.1 [10.3, 41.4] \\
 
MedGemma 1.5
& 11.4 [2.9, 22.9]
& 6.5 [0.0, 16.1]
& 8.8 [0.0, 20.6]
& 10.8 [2.7, 21.6]
& 24.1 [10.3, 41.4] \\
 
\bottomrule
\end{tabular}}
\end{table*}

\begin{table*}[t]
\centering
\caption{Per-category accuracy and bootstrap 95\% confidence intervals for the accuracy-scored reasoning \textsc{Sureon} question types. Entries report accuracy [95\% CI], in percent. LAR: Local Action Reasoning; SPI: Safety Practice Identification; F: Forecasting; TO: Temporal Ordering. Sequence Summarization is open-ended and LLM-judge scored, and is excluded here.}
\label{tab:category_ci_reasoning}
\scriptsize
\resizebox{\textwidth}{!}{
\begin{tabular}{lcccc}
\toprule
Model & LAR ($n=37$) & SPI ($n=31$) & F ($n=31$) & TO ($n=34$) \\
\midrule
Human
& 35.1 [18.9, 51.3]
& 71.0 [54.8, 87.1]
& 54.8 [38.7, 71.0]
& 61.8 [44.1, 76.5] \\
 
Gemini 3.1 Pro
& 56.8 [40.5, 73.0]
& 61.3 [45.2, 77.4]
& 19.4 [6.5, 32.3]
& 64.7 [50.0, 79.4] \\
 
\textsc{Sureon}-VLM
& 43.2 [27.0, 59.5]
& 58.1 [41.9, 74.2]
& 35.5 [19.4, 51.6]
& 52.9 [35.3, 70.6] \\
 
GPT-5.1 (high)
& 46.0 [29.7, 62.2]
& 45.2 [29.0, 64.5]
& 25.8 [9.7, 41.9]
& 52.9 [35.3, 70.6] \\
 
\textsc{Sureon}-VLM (\textsc{Sureon}-only)
& 40.5 [24.3, 56.8]
& 61.3 [45.2, 77.4]
& 38.7 [22.6, 54.8]
& 44.1 [26.5, 61.8] \\
 
Gemma 4
& 29.7 [16.2, 46.0]
& 35.5 [19.4, 51.6]
& 16.1 [3.2, 29.0]
& 41.2 [23.5, 58.8] \\
 
Qwen3VL-32B-T
& 27.0 [13.5, 40.5]
& 19.4 [6.5, 35.5]
& 12.9 [3.2, 25.8]
& 35.3 [20.6, 52.9] \\
 
Qwen3VL-8B-I
& 18.9 [8.1, 32.4]
& 16.1 [3.2, 29.0]
& 12.9 [3.2, 25.8]
& 26.5 [11.8, 41.2] \\
 
\textsc{Sureon}-VLM (Public-only)
& 10.8 [2.7, 21.6]
& 19.4 [6.5, 35.5]
& 6.5 [0.0, 16.1]
& 14.7 [2.9, 26.5] \\
 
MedGRPO
& 13.5 [2.7, 24.3]
& 12.9 [3.2, 25.8]
& 6.5 [0.0, 16.1]
& 20.6 [8.8, 35.3] \\
 
Molmo2
& 8.1 [0.0, 18.9]
& 12.9 [3.2, 25.8]
& 3.2 [0.0, 9.7]
& 8.8 [0.0, 17.6] \\
 
MedGemma 1.5
& 10.8 [2.7, 21.6]
& 6.5 [0.0, 16.1]
& 6.5 [0.0, 16.1]
& 11.8 [2.9, 23.5] \\
 
\bottomrule
\end{tabular}}
\end{table*}
 
\textcolor{black}{Category-level estimates are necessarily less precise
because each accuracy-based category contains only 29--37 samples,
compared with 299 samples for overall $\mu$. Across the 12 models, the average bootstrap 95\% CI half-width is
14.3 percentage points, compared with approximately 5.0 percentage
points overall. We therefore use category-level results primarily to
characterize task-specific strengths and weaknesses rather than to
establish fine-grained model rankings. Perception ($n=166$) and
reasoning ($n=133$) aggregates provide intermediate precision. Notably,
the surgeon baseline attains near-ceiling accuracy on PAD (96.5\%) but
is markedly lower on EE (51.3\%, statistically indistinguishable from
several models), suggesting category difficulty is not uniform even
for expert humans.}

\begin{table}[t]
\centering
\tiny
\caption{Full pairwise significance matrix. McNemar's
test on paired predictions, Holm--Bonferroni corrected across all
$\binom{12}{2}=66$ model pairs. Rows are sorted by corrected $p$-value
(most significant first); $^{*}$ denotes significance at the
Holm-corrected 0.05 level. }
\label{tab:pairwise_sig}
\begin{tabular}{llrrr}
\toprule
Model A & Model B & $\Delta$ Acc.\ (pts) & $p$ (raw) & $p_{\text{holm}}$ \\
\midrule
Human & MedGemma 1.5 & 54.18 & $<$0.001 & $<$0.001$^{*}$ \\
Human & Molmo2 & 49.83 & $<$0.001 & $<$0.001$^{*}$ \\
Human & MedGRPO & 48.16 & $<$0.001 & $<$0.001$^{*}$ \\
Human & \textsc{Sureon}-VLM (Public-only) & 44.15 & $<$0.001 & $<$0.001$^{*}$ \\
Human & Qwen3VL-8B-I & 41.81 & $<$0.001 & $<$0.001$^{*}$ \\
Gemini 3.1 Pro & MedGemma 1.5 & 41.47 & $<$0.001 & $<$0.001$^{*}$ \\
MedGemma 1.5 & \textsc{Sureon}-VLM & -37.46 & $<$0.001 & $<$0.001$^{*}$ \\
Gemini 3.1 Pro & Molmo2 & 37.12 & $<$0.001 & $<$0.001$^{*}$ \\
GPT-5.1 (high) & MedGemma 1.5 & 36.12 & $<$0.001 & $<$0.001$^{*}$ \\
Gemini 3.1 Pro & MedGRPO & 35.45 & $<$0.001 & $<$0.001$^{*}$ \\
MedGemma 1.5 & \textsc{Sureon}-VLM (\textsc{Sureon}-only) & -33.44 & $<$0.001 & $<$0.001$^{*}$ \\
Molmo2 & \textsc{Sureon}-VLM & -33.11 & $<$0.001 & $<$0.001$^{*}$ \\
GPT-5.1 (high) & Molmo2 & 31.77 & $<$0.001 & $<$0.001$^{*}$ \\
Gemini 3.1 Pro & \textsc{Sureon}-VLM (Public-only) & 31.44 & $<$0.001 & $<$0.001$^{*}$ \\
MedGRPO & \textsc{Sureon}-VLM & -31.44 & $<$0.001 & $<$0.001$^{*}$ \\
Human & Qwen3VL-32B-T & 30.77 & $<$0.001 & $<$0.001$^{*}$ \\
GPT-5.1 (high) & MedGRPO & 30.10 & $<$0.001 & $<$0.001$^{*}$ \\
Human & Gemma 4 & 29.10 & $<$0.001 & $<$0.001$^{*}$ \\
Gemini 3.1 Pro & Qwen3VL-8B-I & 29.10 & $<$0.001 & $<$0.001$^{*}$ \\
Molmo2 & \textsc{Sureon}-VLM (\textsc{Sureon}-only) & -29.10 & $<$0.001 & $<$0.001$^{*}$ \\
MedGRPO & \textsc{Sureon}-VLM (\textsc{Sureon}-only) & -27.42 & $<$0.001 & $<$0.001$^{*}$ \\
\textsc{Sureon}-VLM (Public-only) & \textsc{Sureon}-VLM & -27.42 & $<$0.001 & $<$0.001$^{*}$ \\
GPT-5.1 (high) & \textsc{Sureon}-VLM (Public-only) & 26.09 & $<$0.001 & $<$0.001$^{*}$ \\
Qwen3VL-8B-I & \textsc{Sureon}-VLM & -25.08 & $<$0.001 & $<$0.001$^{*}$ \\
Gemma 4 & MedGemma 1.5 & 25.08 & $<$0.001 & $<$0.001$^{*}$ \\
GPT-5.1 (high) & Qwen3VL-8B-I & 23.75 & $<$0.001 & $<$0.001$^{*}$ \\
Qwen3VL-32B-T & MedGemma 1.5 & 23.41 & $<$0.001 & $<$0.001$^{*}$ \\
\textsc{Sureon}-VLM (\textsc{Sureon}-only) & \textsc{Sureon}-VLM (Public-only) & 23.41 & $<$0.001 & $<$0.001$^{*}$ \\
Qwen3VL-8B-I & \textsc{Sureon}-VLM (\textsc{Sureon}-only) & -21.07 & $<$0.001 & $<$0.001$^{*}$ \\
Human & \textsc{Sureon}-VLM (\textsc{Sureon}-only) & 20.74 & $<$0.001 & $<$0.001$^{*}$ \\
Gemma 4 & Molmo2 & 20.74 & $<$0.001 & $<$0.001$^{*}$ \\
Gemma 4 & MedGRPO & 19.06 & $<$0.001 & $<$0.001$^{*}$ \\
Qwen3VL-32B-T & Molmo2 & 19.06 & $<$0.001 & $<$0.001$^{*}$ \\
Human & GPT-5.1 (high) & 18.06 & $<$0.001 & $<$0.001$^{*}$ \\
Gemini 3.1 Pro & Qwen3VL-32B-T & 18.06 & $<$0.001 & $<$0.001$^{*}$ \\
Qwen3VL-32B-T & MedGRPO & 17.39 & $<$0.001 & $<$0.001$^{*}$ \\
Gemini 3.1 Pro & Gemma 4 & 16.39 & $<$0.001 & $<$0.001$^{*}$ \\
Gemma 4 & \textsc{Sureon}-VLM (Public-only) & 15.05 & $<$0.001 & $<$0.001$^{*}$ \\
Qwen3VL-8B-I & MedGemma 1.5 & 12.37 & $<$0.001 & $<$0.001$^{*}$ \\
Human & \textsc{Sureon}-VLM & 16.72 & $<$0.001 & 0.0027$^{*}$ \\
Qwen3VL-32B-T & \textsc{Sureon}-VLM (Public-only) & 13.38 & $<$0.001 & 0.0027$^{*}$ \\
Qwen3VL-8B-I & Gemma 4 & -12.71 & $<$0.001 & 0.0027$^{*}$ \\
Qwen3VL-32B-T & \textsc{Sureon}-VLM & -14.05 & $<$0.001 & 0.0072$^{*}$ \\
Qwen3VL-32B-T & Qwen3VL-8B-I & 11.04 & $<$0.001 & 0.0092$^{*}$ \\
GPT-5.1 (high) & Qwen3VL-32B-T & 12.71 & $<$0.001 & 0.0110$^{*}$ \\
MedGemma 1.5 & \textsc{Sureon}-VLM (Public-only) & -10.03 & $<$0.001 & 0.0126$^{*}$ \\
Gemma 4 & \textsc{Sureon}-VLM & -12.37 & $<$0.001 & 0.0140$^{*}$ \\
Human & Gemini 3.1 Pro & 12.71 & 0.0013 & 0.0247$^{*}$ \\
GPT-5.1 (high) & Gemma 4 & 11.04 & 0.0013 & 0.0247$^{*}$ \\
Qwen3VL-8B-I & Molmo2 & 8.03 & 0.0022 & 0.0374$^{*}$ \\
Qwen3VL-32B-T & \textsc{Sureon}-VLM (\textsc{Sureon}-only) & -10.03 & 0.0064 & 0.1024 \\
Gemma 4 & \textsc{Sureon}-VLM (\textsc{Sureon}-only) & -8.36 & 0.0261 & 0.3915 \\
Gemini 3.1 Pro & \textsc{Sureon}-VLM (\textsc{Sureon}-only) & 8.03 & 0.0323 & 0.4522 \\
MedGemma 1.5 & MedGRPO & -6.02 & 0.0414 & 0.5382 \\
Qwen3VL-8B-I & MedGRPO & 6.35 & 0.0448 & 0.5382 \\
Molmo2 & \textsc{Sureon}-VLM (Public-only) & -5.69 & 0.0533 & 0.5863 \\
Molmo2 & MedGemma 1.5 & 4.35 & 0.1175 & 1.0000 \\
GPT-5.1 (high) & Gemini 3.1 Pro & -5.35 & 0.1448 & 1.0000 \\
MedGRPO & \textsc{Sureon}-VLM (Public-only) & -4.01 & 0.2007 & 1.0000 \\
\textsc{Sureon}-VLM (\textsc{Sureon}-only) & \textsc{Sureon}-VLM & -4.01 & 0.2185 & 1.0000 \\
Gemini 3.1 Pro & \textsc{Sureon}-VLM & 4.01 & 0.3232 & 1.0000 \\
Qwen3VL-8B-I & \textsc{Sureon}-VLM (Public-only) & 2.34 & 0.4944 & 1.0000 \\
GPT-5.1 (high) & \textsc{Sureon}-VLM (\textsc{Sureon}-only) & 2.68 & 0.5363 & 1.0000 \\
Molmo2 & MedGRPO & -1.67 & 0.6305 & 1.0000 \\
Qwen3VL-32B-T & Gemma 4 & -1.67 & 0.6718 & 1.0000 \\
GPT-5.1 (high) & \textsc{Sureon}-VLM & -1.34 & 0.7926 & 1.0000 \\
\bottomrule
\end{tabular}
\end{table}

\textbf{LLM-judge for Sequence Summarization.} The SM judge scores each prediction against a fixed transcript-derived reference, not the judge's own surgical knowledge. Seeing only the two texts, it is prompted to be strict (Fig.~\ref{fig:llmjudge-prompt}): a prediction is incorrect if any reference step is missing, if meaningful ordering is violated, or if an action is substituted (e.g.\ a structure the reference preserves is reported as divided), while paraphrases and surgical synonyms count as matching. \\

\noindent \textcolor{black}{\textit{LLM Judge Validation.}  To validate the judge, three annotators independently rated 44 stratified predictions sampled across all 11 models (4 per model, balanced by verdict). Raters were shown only the question, reference, and extracted prediction and applied the same two-criterion rubric. Inter-rater agreement was moderate (mean pairwise raw agreement 64.5\%), consistent with the subjectivity of surgical sequence evaluation. Majority-vote human verdicts agreed with the LLM judge on 79.1\% of samples. Disagreements were asymmetric: in 6 of the 10 discordant cases the judge marked a prediction incorrect that humans accepted, versus 3 in the opposite direction, indicating that the judge applies the rubric more strictly than human raters.}

\subsection{Qualitative Error Analysis}
\begin{figure*}
    \centering
    \includegraphics[width=0.7\linewidth]{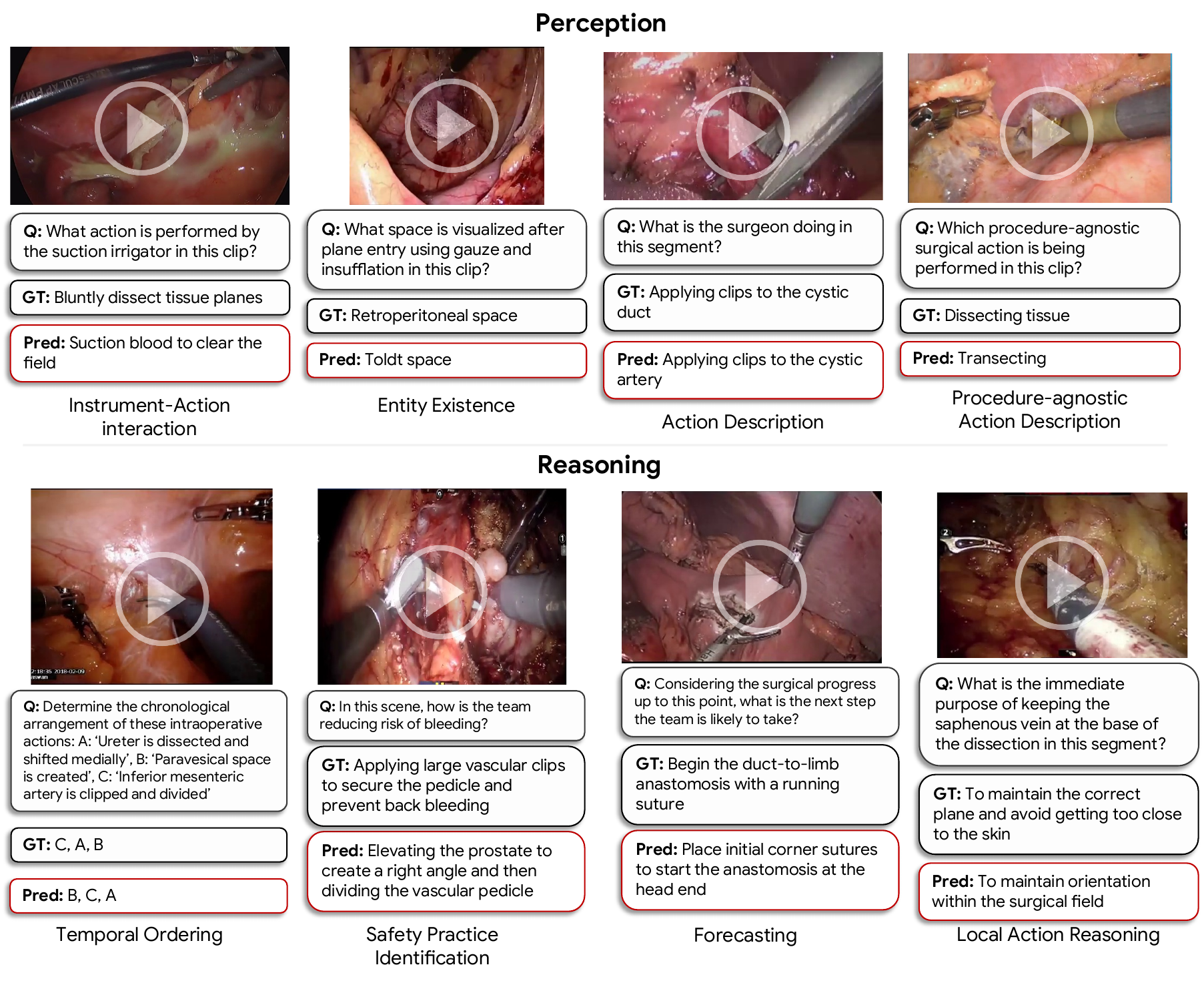}
    \caption{\textcolor{black}{\textbf{Representative \textsc{Sureon}-VLM failure cases} across four perception categories (Instrument-Action Interaction, Entity Existence, Action Description, Procedure-Agnostic Action Description) and four reasoning categories (Temporal Ordering, Safety Practice Identification, Forecasting, Local Action Reasoning). For each example we show the question (Q), \textsc{Sureon}-VLM's prediction (Pred), and the ground-truth answer (GT).}}
    \label{fig:failures}
\end{figure*}

Figure~\ref{fig:failures} presents representative \textsc{Sureon}-VLM failures across four perception and four reasoning categories, revealing two distinct error modes.

\textcolor{black}{\textbf{Perception errors are near-misses, not random guesses.} Predictions consistently land in the correct local neighborhood of the ground truth rather than on implausible alternatives. In Action Description, the model correctly locates clip application in Calot's triangle but confuses the cystic \textit{artery} for the cystic \textit{duct}. In Entity Existence, ``Toldt space'' is predicted for ``retroperitoneal space,'' an adjacent landmark. These suggest coarse scene understanding without the fine-grained discrimination the benchmark targets. The Instrument-Action example is a sharper failure: the model predicts the suction irrigator's \textit{canonical} function (clearing blood) over its atypical use in this clip (blunt dissection), which is a language prior-driven error rather than a visual one, and a reminder that dataset-level shortcut removal (RADG, blind-testing) still has limitations in preventing a trained model from defaulting to real-world instrument priors when visual attention is weak.}

\textcolor{black}{\textbf{Reasoning errors tend toward generic, plausible-sounding answers.} In Safety Practice Identification, the model describes a real but different aspect of the scene (a dissection step) rather than the specific hemostatic action asked about. Local Action Reasoning shows this most clearly: ``maintain orientation within the surgical field'' is a generic justification that could apply to nearly any dissection, where the ground truth requires reasoning about a specific plane and proximity to skin. These are fluent, surgically literate answers that fail on specificity. Forecasting errors are comparatively softer, correctly anticipating the coarse next phase while erring on technical detail.}

\subsection{Public Perception Benchmark Evaluation}
\label{sec:appendix-eval-public}

\textbf{Inputs and clip construction.} Public benchmarks are evaluated with three input modes depending on the task. Frame-level tasks (tool localization and frame-level phase/action recognition) use a single frame. Clip-level recognition uses a fixed window of $8$ frames at a stride of $4$ (spanning up to $\sim$$28$~s at $1$~fps extraction) anchored on the keyframe. We evaluate all clip tasks in the online setting, where the keyframe is the final frame of the window with preceding frames providing causal context. \textsc{Sureon}-VLM is evaluated in both frame and clip variants where applicable. Commercial and open-source baselines are evaluated in frame-only mode on public benchmarks—commercial models due to inference costs and surgical models due to unavailable weights—using numbers reported in their original papers.

\begin{figure}[t]
    \centering
    \includegraphics[width=\linewidth]{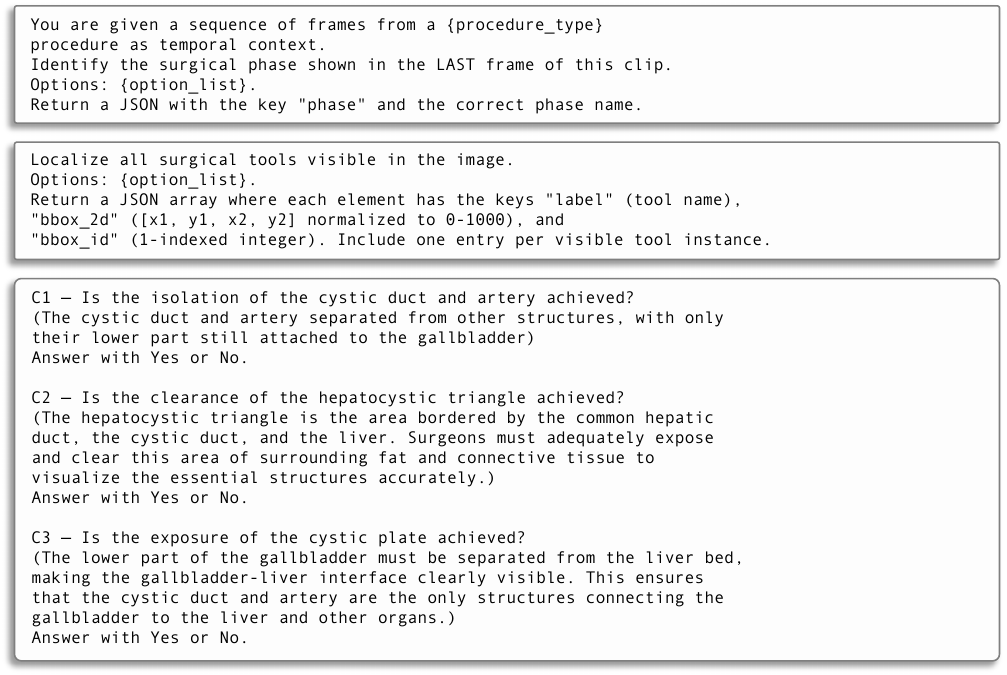}
    \caption{Prompts used for the public perception benchmarks: phase/step recognition, tool localization, and the three Critical View of Safety criteria (C1–C3).}
    \label{fig:downstream-prompts}
\end{figure}

\textbf{Recognition and classification metrics.} Phase and action recognition are scored by video-averaged macro-F1 (per-video macro-F1 averaged over videos), which avoids long videos dominating the aggregate.  For Cholec80 we additionally report accuracy, recall, and precision, all video-averaged. For Endoscapes Critical View of Safety (CVS), each of the three criteria (C1–C3) is posed as an independent binary (Yes/No) query (prompts in Fig.~\ref{fig:downstream-prompts}). We report per-criterion accuracy and the macro-average over C1–C3.

\textbf{Localization metrics.} Tool localization is scored by mAP$_{50}$, the COCO bounding-box average precision at an IoU threshold of $0.50$. Because generative models emit boxes without confidence scores, all predicted boxes are assigned a constant score, so every prediction participates in the precision–recall computation. The category universe is defined by the ground-truth class names; predicted labels are canonicalized (lowercased, normalized for spacing and trivial pluralization) before matching. 

\textbf{Test splits and exclusions.} Each benchmark is evaluated on its official test split. For Cholec80 clip evaluation we exclude $17$ videos that overlap with training sources from Endoscapes and CholecT50 to prevent train/test leakage \footnote{As suggested in \url{https://github.com/CAMMA-public/camma_dataset_overlaps}}.

\textbf{Downstream task prompts.} The prompts used to query models for phase recognition, tool localization, and CVS assessment are shown in Fig.~\ref{fig:downstream-prompts}. Each prompt enumerates the closed label set as the option list and specifies a structured (JSON) answer format to enable checkable scoring. 

\section{\textsc{Sureon}-VLM training details}
\label{sec:appendix-training}

A common approach to domain adaptation is vanilla supervised fine-tuning (SFT),
which we take as a representative baseline to study how data composition and
staging order affect both reasoning and grounding on the  \texttt{Qwen3-VL-8B-Instruct} backbone. Our adaptation combines two complementary resources: public datasets
supplying annotated images and clips for spatial grounding and workflow
recognition, and \textsc{Sureon}-DB supplying narration-derived VQA for reasoning
over actions, states, and procedural progression. We evaluate on \textsc{Sureon}
for visual reasoning alongside four public grounding benchmarks: CholecInstanceSeg
and GraSP for tool localization, Cholec80 for phase recognition, and SARRARP-50
for action recognition.

\noindent\textbf{Setup.} We vary the data---\textbf{P} (public images),
\textbf{S} (\textsc{Sureon}-DB), \textbf{P\,+\,S} (full corpus)---and exposure
order. \emph{Single} trains in one stage; \emph{P$\rightarrow$mix} and
\emph{S$\rightarrow$mix} warm-start on public images or \textsc{Sureon} video
before a second stage on the full mixture; \emph{mix$_1\rightarrow$mix$_2$}
splits the mixture into two stages, image-heavy first, then balanced.

\begin{table}[t]
\centering
\caption{Training data and staging ablation, evaluated on \textsc{Sureon} and public
benchmarks covering spatial grounding (\colorbox{toolcol}{tool recognition})
and temporal tasks (\colorbox{phasecol}{phase} and \colorbox{actioncol}{action}
recognition). Data sources: \textbf{P} (public images), \textbf{S} (\textsc{Sureon}-DB),
\textbf{P\,+\,S} (full corpus of public images, public clips, and \textsc{Sureon}).
Staging schedules: \emph{single} trains in one stage;
\emph{P$\rightarrow$mix} / \emph{S$\rightarrow$mix} warm-start on public images /
\textsc{Sureon} video, then train on the full mixture; \emph{mix$_1\rightarrow$mix$_2$}
splits the full mixture into an image-heavy stage (mix$_1$) followed by a balanced
stage (mix$_2$).
\textbf{C80:} Cholec80, \textbf{CISeg:} CholecInstanceSeg,
\textbf{SP50:} SAR-RARP50. The backbone (Qwen3-VL-8B) is a zero-shot reference. Best in \textbf{bold}, second best \underline{underlined}.}
\label{tab:training-ablation}
\setlength{\tabcolsep}{5pt}
\renewcommand{\arraystretch}{1.1}
\resizebox{\columnwidth}{!}{%
\begin{tabular}{@{}cc ccccc@{}}
\toprule
Data & Stage
  & \makecell{\textsc{Sureon}\\(Acc.)}
  & \cellcolor{phasecol}\makecell{C80~\cite{twinanda2016endonet}\\(F1)}
  & \cellcolor{actioncol}\makecell{SP50~\cite{psychogyios2023sar}\\(F1)}
  & \cellcolor{toolcol}\makecell{CISeg~\cite{alabi2025cholecinstanceseg}\\(mAP$_{50}$)}
  & \cellcolor{toolcol}\makecell{GraSP~\cite{ayobi2025pixel}\\(mAP$_{50}$)} \\
\midrule
\multicolumn{2}{c}{Qwen3-VL-8B (Base model)} & 24.71 & 17.18 & 14.46 & 9.92 & 0.86 \\
\midrule
P       & single                      & 14.72 & 53.72 & 8.35  & 75.92 & \textbf{60.66} \\
S       & single                      & 43.10 & 6.67 & 7.38 & 3.34 & 3.54 \\
\midrule
P\,+\,S & P\,$\rightarrow$\,mix        & 43.97 & 74.87 & 23.19 & 67.09 & \underline{59.93} \\
P\,+\,S & S\,$\rightarrow$\,mix        & \underline{46.02} & 70.87 & 20.51 & 61.53 & 42.86 \\
\midrule
P\,+\,S & single                      & \textbf{46.67} & \underline{76.90} & \underline{24.50} & \textbf{77.70} & 59.38 \\
P\,+\,S & mix$_1$\,$\rightarrow$\,mix$_2$ & 45.10 & \textbf{78.86} & \textbf{25.21} & \underline{77.45} & 59.56 \\
\bottomrule
\end{tabular}}
\end{table}
\noindent\textbf{Single-source fine-tuning erodes existing capabilities.} We
read the zero-shot backbone as a reference for \emph{forgetting}: when training
on source $A$ drops task $B$ below the untuned base, training has overwritten
capability the base already had. Both single-source runs do this. Public-only
training raises grounding but drops \textsc{Sureon} (24.7$\rightarrow$14.7);
\textsc{Sureon}-only raises \textsc{Sureon} but collapses tool localization on
CISeg (9.9$\rightarrow$3.3). Action recognition (SP50) falls below the base under
\emph{both} (14.5$\rightarrow$8.4 public, $\rightarrow$7.4 \textsc{Sureon}), and
\textsc{Sureon} also degrades phase recognition on C80 (17.2$\rightarrow$6.7).
Single-source supervision thus sacrifices abilities the model already possessed
rather than specializing it.

\noindent\textbf{Staging order determines what is retained.} Exposure order
matters in the mixed corpus. Public-then-video (P$\rightarrow$mix) recovers
\textsc{Sureon} but partly forgets the grounding learned first
(CISeg 75.9$\rightarrow$67.1; GraSP 60.7$\rightarrow$59.9). The reverse is more
revealing: adding public data after \textsc{Sureon} video lifts \textsc{Sureon}
by over three points (43.1$\rightarrow$46.0), yet grounding never recovers
(GraSP only 42.9, $\sim$17 points below schedules that see images early). Grounding
installed late is harder to add than reasoning layered onto a perception-first
model---an asymmetry favoring early image exposure.

\noindent\textbf{Strong downstream scores do not mean the model understands the
scene.} The combined schedules look like clear wins on the public benchmarks, but
these reward fitting a fixed label set, not seeing the scene; adaptation moves
capability around more than it adds understanding. Even the balanced schedules,
which avoid the worst forgetting, fall short on \textsc{Sureon} and stay sensitive
to training order. The open problem is how to adapt a model so it reasons over the
surgical scene \emph{without} forgetting how to see it---which \textsc{Sureon},
requiring perception to answer, is built to track. 

\noindent\textbf{Final \textsc{Sureon}-VLM Configuration.} Based on this ablation, \textsc{Sureon}-VLM uses the P+S (\emph{mix$_1\rightarrow$mix$_2$}) staging schedule, as it achieves the best average performance across \textsc{Sureon} and downstream perception tasks. This model fine-tunes \texttt{Qwen3-VL-8B-Instruct} on the consolidated corpus $\mathcal{D}_{\mathrm{all}}$ in two stages: Stage 1 with image-heavy composition, Stage 2 with balanced data mixture, as described above.

\noindent\textbf{Training implementation.}
All runs fully fine-tune \texttt{Qwen3-VL-8B-Instruct} (vision encoder, MLP
projector, and LLM) for 10 epochs in bfloat16, using AdamW with a learning rate
of $2\times10^{-6}$, a cosine schedule with 10\% linear warmup, and no weight
decay. We use a per-device batch size of 8 with gradient accumulation of 4 over
6 H200 GPUs (effective batch size 192) under DeepSpeed ZeRO-3, a context length
of 8{,}192 tokens, and 8 frames per clip. For each \textsc{Sureon}-DB
multiple-choice item, the distractor set is resampled at random from its Stage-2
candidate pool (Sec.~\ref{sec:appendix-stage2}) each time the item is loaded, so
option content and ordering vary across epochs as augmentation. Both stages of the
best two-stage schedule (\emph{mix$_1\rightarrow$mix$_2$}) share these
hyperparameters: Stage~1 initializes from the pretrained base with an image-heavy
mixture (image 72.7\% / clip 14.9\% / \textsc{Sureon}-DB 12.4\%), and Stage~2
warm-starts from the Stage-1 checkpoint with an equal-thirds mixture to re-balance
the sources. Given the relative corpus sizes, the \textsc{Sureon}-DB video data is
effectively seen $\sim$20$\times$ over the schedule.

\section{Ethics}   
\label{sec:ethics}
\noindent\textbf{Ethical use of data.}
\textsc{Sureon} is constructed from publicly available surgical lecture videos and is intended for research use only. \textsc{Sureon}-DB is automatically pseudo-labeled and is not expert-verified; models trained on it are research artifacts and must not be used for clinical decision-making or surgical practice in their current form. The expert-verified benchmark is intended for evaluation, not as clinical ground truth.

\noindent\textbf{Data licensing and usage.}
\textsc{Sureon} is derived from SurgLaVi-$\beta$~\cite{perez2026surglavi}, which is released under the Creative Commons Attribution-NonCommercial-ShareAlike 4.0 International License (CC BY-NC-SA 4.0). In accordance with the ShareAlike provision, we release both \textsc{Sureon}-DB and the \textsc{Sureon} benchmark under the same CC BY-NC-SA 4.0 license, restricting use to non-commercial research and requiring attribution and identical redistribution terms. We release the derived QA annotations, temporal windows, and metadata; the underlying lecture videos remain governed by their original sources and are not redistributed.

\textcolor{black}{\noindent\textbf{De-identification.}
The source material consists of publicly available educational lecture videos that are edited and narrated for pedagogical purposes; they do not contain patient names, medical record numbers, or other protected health identifiers. Laparoscopic and robot-assisted footage depicts internal anatomy rather than external patient features, inherently limiting re-identification risk. }

%% file: main.bbl
\begin{thebibliography}{78}
\providecommand{\natexlab}[1]{#1}
\providecommand{\url}[1]{\texttt{#1}}
\expandafter\ifx\csname urlstyle\endcsname\relax
  \providecommand{\doi}[1]{doi: #1}\else
  \providecommand{\doi}{doi: \begingroup \urlstyle{rm}\Url}\fi

\bibitem[Alabi et~al.(2025)]{alabi2025cholecinstanceseg}
Oluwatosin Alabi et~al.
\newblock Cholecinstanceseg: A tool instance segmentation dataset for laparoscopic surgery.
\newblock \emph{Scientific Data}, 12\penalty0 (1):\penalty0 825, 2025.

\bibitem[Allan et~al.(2019)]{allan20192017}
Max Allan et~al.
\newblock 2017 robotic instrument segmentation challenge.
\newblock \emph{arXiv preprint arXiv:1902.06426}, 2019.

\bibitem[Allan et~al.(2020)]{allan20202018}
Max Allan et~al.
\newblock 2018 robotic scene segmentation challenge.
\newblock \emph{arXiv preprint arXiv:2001.11190}, 2020.

\bibitem[Asadi et~al.(2026)Asadi, O'Sullivan, Cao, Nedaee, Rajabalifardi, Li, Adeli, and Ashley]{asadi2026mirage}
Mohammad Asadi, Jack~W O'Sullivan, Fang Cao, Tahoura Nedaee, Kamyar Rajabalifardi, Fei-Fei Li, Ehsan Adeli, and Euan Ashley.
\newblock Mirage: The illusion of visual understanding.
\newblock \emph{arXiv preprint arXiv:2603.21687}, 2026.

\bibitem[Ayobi et~al.(2025)]{ayobi2025pixel}
Nicol{\'a}s Ayobi et~al.
\newblock Pixel-wise recognition for holistic surgical scene understanding.
\newblock \emph{MedIA}, page 103726, 2025.

\bibitem[Badaloni et~al.(2026)Badaloni, Kuiper, de~Jong, van Jaarsveld, Li, van~der Neut, Asag-Gau, Zeyara, Breeuwer, van Hillegersberg, et~al.]{badaloni2026we}
Franco Badaloni, Gino Kuiper, Ronald de Jong, Romy van Jaarsveld, Yiping Li, Mart van~der Neut, Alae Asag-Gau, Adam Zeyara, Marcel Breeuwer, Richard van Hillegersberg, et~al.
\newblock Are we developing the right intraoperative ai assistance? surgeons’ perspectives and desired functions.
\newblock \emph{Surgical Endoscopy}, 40\penalty0 (6):\penalty0 5259--5266, 2026.

\bibitem[Bai et~al.(2023)Bai, Islam, Seenivasan, and Ren]{bai2023surgical}
Long Bai, Mobarakol Islam, Lalithkumar Seenivasan, and Hongliang Ren.
\newblock Surgical-vqla: Transformer with gated vision-language embedding for visual question localized-answering in robotic surgery.
\newblock \emph{arXiv preprint arXiv:2305.11692}, 2023.

\bibitem[Bai et~al.(2025)]{bai2025qwen3}
Shuai Bai et~al.
\newblock Qwen3-vl technical report.
\newblock \emph{arXiv preprint arXiv:2511.21631}, 2025.

\bibitem[Bellos et~al.(2026)Bellos, Gala-Garza, Wang, Hardin, Hider, Li, Bi, Liang, Xu, Likosky, and Corso]{bellos2026surgatlas}
Filippos Bellos, Andre~S. Gala-Garza, Miaowei Wang, Alyssa~M. Hardin, Ahmad~M. Hider, Yayuan Li, Jing Bi, Susan Liang, Chenliang Xu, Donald~S. Likosky, and Jason~J. Corso.
\newblock {SurgAtlas}: A large-scale surgical video--language dataset with 2,391 hours of open and minimally invasive surgery.
\newblock \emph{arXiv preprint arXiv:2606.25905}, 2026.

\bibitem[Burgess et~al.(2025)Burgess, Nirschl, Bravo-S{\'a}nchez, Lozano, Gupte, Galaz-Montoya, Zhang, Su, Bhowmik, Coman, et~al.]{burgess2025microvqa}
James Burgess, Jeffrey~J Nirschl, Laura Bravo-S{\'a}nchez, Alejandro Lozano, Sanket~Rajan Gupte, Jesus~G Galaz-Montoya, Yuhui Zhang, Yuchang Su, Disha Bhowmik, Zachary Coman, et~al.
\newblock Microvqa: A multimodal reasoning benchmark for microscopy-based scientific research.
\newblock In \emph{Proceedings of the IEEE/CVF Conference on Computer Vision and Pattern Recognition}, pages 19552--19564, 2025.

\bibitem[Carstens et~al.(2023)]{carstens2023dresden}
Matthias Carstens et~al.
\newblock The dresden surgical anatomy dataset for abdominal organ segmentation in surgical data science.
\newblock \emph{Scientific Data}, 10\penalty0 (1):\penalty0 3, 2023.

\bibitem[Che et~al.(2025)Che, Wang, Vercauteren, Tsoka, and Garcia-Peraza-Herrera]{che2025lemon}
Chengan Che, Chao Wang, Tom Vercauteren, Sophia Tsoka, and Luis~C Garcia-Peraza-Herrera.
\newblock Lemon: A large endoscopic monocular dataset and foundation model for perception in surgical settings.
\newblock \emph{arXiv preprint arXiv:2503.19740}, 2025.

\bibitem[Chen et~al.(2024{\natexlab{a}})Chen, Xiao, Zhang, Luo, Lian, and Liu]{chen2024m3}
Jianlyu Chen, Shitao Xiao, Peitian Zhang, Kun Luo, Defu Lian, and Zheng Liu.
\newblock M3-embedding: Multi-linguality, multi-functionality, multi-granularity text embeddings through self-knowledge distillation.
\newblock In \emph{Findings of the association for computational linguistics: ACL 2024}, pages 2318--2335, 2024{\natexlab{a}}.

\bibitem[Chen et~al.(2025)Chen, Xu, Fei, Feng, and Elhoseiny]{chen2025document}
Jun Chen, Dannong Xu, Junjie Fei, Chun-Mei Feng, and Mohamed Elhoseiny.
\newblock Document haystacks: Vision-language reasoning over piles of 1000+ documents.
\newblock In \emph{Proceedings of the Computer Vision and Pattern Recognition Conference}, pages 24817--24826, 2025.

\bibitem[Chen et~al.(2024{\natexlab{b}})Chen, Li, Dong, Zhang, Zang, Chen, Duan, Wang, Qiao, Lin, et~al.]{chen2024we}
Lin Chen, Jinsong Li, Xiaoyi Dong, Pan Zhang, Yuhang Zang, Zehui Chen, Haodong Duan, Jiaqi Wang, Yu Qiao, Dahua Lin, et~al.
\newblock Are we on the right way for evaluating large vision-language models?
\newblock \emph{Advances in Neural Information Processing Systems}, 37:\penalty0 27056--27087, 2024{\natexlab{b}}.

\bibitem[Chen et~al.(2024{\natexlab{c}})Chen, Zhu, Zheng, Li, and Yang]{chen2024wsi}
Pingyi Chen, Chenglu Zhu, Sunyi Zheng, Honglin Li, and Lin Yang.
\newblock Wsi-vqa: Interpreting whole slide images by generative visual question answering.
\newblock In \emph{European Conference on Computer Vision}, pages 401--417. Springer, 2024{\natexlab{c}}.

\bibitem[Cheng et~al.(2026)Cheng, Hu, Liu, Si, Li, and Gong]{cheng2026v}
Zixu Cheng, Jian Hu, Ziquan Liu, Chenyang Si, Wei Li, and Shaogang Gong.
\newblock V-star: Benchmarking video-llms on video spatio-temporal reasoning.
\newblock In \emph{Proceedings of the IEEE/CVF Conference on Computer Vision and Pattern Recognition}, pages 9155--9164, 2026.

\bibitem[Choi et~al.(2025)Choi, Jeong, Kim, Lee, Koh, Choi, Chung, Park, and Park]{choi2025surgmllmbench}
Tae-Min Choi, Tae~Kyeong Jeong, Garam Kim, Jaemin Lee, Yeongyoon Koh, In~Cheul Choi, Jae-Ho Chung, Jong~Woong Park, and Juyoun Park.
\newblock Surgmllmbench: A multimodal large language model benchmark dataset for surgical scene understanding.
\newblock \emph{arXiv preprint arXiv:2511.21339}, 2025.

\bibitem[Clark et~al.(2026)Clark, Zhang, Ma, Park, Tripathi, Lee, Salehi, Ren, Kim, Yang, et~al.]{clark2026molmo2}
Christopher Clark, Jieyu Zhang, Zixian Ma, Jae~Sung Park, Rohun Tripathi, Sangho Lee, Mohammadreza Salehi, Jason Ren, Chris~Dongjoo Kim, Yinuo Yang, et~al.
\newblock Molmo2: Open weights and data for vision-language models with video understanding and grounding.
\newblock In \emph{Proceedings of the IEEE/CVF Conference on Computer Vision and Pattern Recognition}, pages 28652--28668, 2026.

\bibitem[Dekoninck et~al.(2026)Dekoninck, Jovanović, Gehrunger, Rögnvaldsson, Petrov, Sun, and Vechev]{dekoninck2026matharena}
Jasper Dekoninck, Nikola Jovanović, Tim Gehrunger, Kári Rögnvaldsson, Ivo Petrov, Chenhao Sun, and Martin Vechev.
\newblock Beyond benchmarks: Matharena as an evaluation platform for mathematics with llms.
\newblock 2026.

\bibitem[Deng et~al.(2025)Deng, Yang, Yu, Spencer, Bansal, Chen, Yeung-Levy, and Wang]{deng2025scivideobench}
Andong Deng, Taojiannan Yang, Shoubin Yu, Lincoln Spencer, Mohit Bansal, Chen Chen, Serena Yeung-Levy, and Xiaohan Wang.
\newblock Scivideobench: Benchmarking scientific video reasoning in large multimodal models.
\newblock \emph{arXiv preprint arXiv:2510.08559}, 2025.

\bibitem[Drago et~al.(2026)Drago, Carlini, Balyemez, Pierantozzi, Lena, Hassan, Stoyanov, Momi, Bano, and Hoque]{drago2026surgvivqa}
Mauro~Orazio Drago, Luca Carlini, Pelinsu~Celebi Balyemez, Dennis Pierantozzi, Chiara Lena, Cesare Hassan, Danail Stoyanov, Elena~De Momi, Sophia Bano, and Mobarak~I Hoque.
\newblock Surgvivqa: temporally grounded video question answering for surgical scene understanding.
\newblock \emph{International Journal of Computer Assisted Radiology and Surgery}, pages 1--9, 2026.

\bibitem[{Google DeepMind}(2025)]{gemini3pro2025}
{Google DeepMind}.
\newblock Gemini 3 pro frontier safety framework report.
\newblock Technical report, Google, 2025.

\bibitem[{Google DeepMind}(2026)]{gemma42026}
{Google DeepMind}.
\newblock Gemma 4: Open multimodal and reasoning models.
\newblock Model documentation, 2026.

\bibitem[Grattafiori et~al.(2024)Grattafiori, Dubey, Jauhri, Pandey, Kadian, Al-Dahle, Letman, Mathur, Schelten, Vaughan, et~al.]{grattafiori2024llama}
Aaron Grattafiori, Abhimanyu Dubey, Abhinav Jauhri, Abhinav Pandey, Abhishek Kadian, Ahmad Al-Dahle, Aiesha Letman, Akhil Mathur, Alan Schelten, Alex Vaughan, et~al.
\newblock The llama 3 herd of models.
\newblock \emph{arXiv preprint arXiv:2407.21783}, 2024.

\bibitem[Hong et~al.(2020)Hong, Kao, Kuo, Wang, Chang, and Shih]{hong2020cholecseg8k}
W-Y Hong, C-L Kao, Y-H Kuo, J-R Wang, W-L Chang, and C-S Shih.
\newblock Cholecseg8k: a semantic segmentation dataset for laparoscopic cholecystectomy based on cholec80.
\newblock \emph{arXiv preprint arXiv:2012.12453}, 2020.

\bibitem[Kwon et~al.(2023)Kwon, Li, Zhuang, Sheng, Zheng, Yu, Gonzalez, Zhang, and Stoica]{kwon2023vllm}
Woosuk Kwon, Zhuohan Li, Siyuan Zhuang, Ying Sheng, Lianmin Zheng, Cody~Hao Yu, Joseph~E. Gonzalez, Hao Zhang, and Ion Stoica.
\newblock Efficient memory management for large language model serving with pagedattention.
\newblock In \emph{Proceedings of the ACM SIGOPS 29th Symposium on Operating Systems Principles}, 2023.

\bibitem[Lau et~al.(2018)Lau, Gayen, Ben~Abacha, and Demner-Fushman]{lau2018dataset}
Jason~J Lau, Soumya Gayen, Asma Ben~Abacha, and Dina Demner-Fushman.
\newblock A dataset of clinically generated visual questions and answers about radiology images.
\newblock \emph{Scientific data}, 5\penalty0 (1):\penalty0 180251, 2018.

\bibitem[Lavanchy et~al.(2024)]{Lavanchy2024}
Joël~L. Lavanchy et~al.
\newblock Challenges in multi-centric generalization: phase and step recognition in roux-en-y gastric bypass surgery.
\newblock \emph{IJCARS}, 2024.

\bibitem[Leibetseder et~al.(2018)]{leibetseder2018lapgyn4}
Andreas Leibetseder et~al.
\newblock Lapgyn4: a dataset for 4 automatic content analysis problems in the domain of laparoscopic gynecology.
\newblock In \emph{Proceedings of the 9th ACM multimedia systems conference}, pages 357--362, 2018.

\bibitem[Li et~al.(2024)Li, Skinner, Yang, Quaranto, Schwaitzberg, Kim, and Xiong]{li2024llava}
Jiajie Li, Garrett Skinner, Gene Yang, Brian~R Quaranto, Steven~D Schwaitzberg, Peter~CW Kim, and Jinjun Xiong.
\newblock Llava-surg: towards multimodal surgical assistant via structured surgical video learning.
\newblock \emph{arXiv preprint arXiv:2408.07981}, 2024.

\bibitem[Li et~al.(2026)Li, Yang, Xu, Yu, Zhao, Hu, Li, and Heng]{li2026surgpub}
Yaoqian Li, Xikai Yang, Dunyuan Xu, Yang Yu, Litao Zhao, Xiaowei Hu, Jinpeng Li, and Pheng-Ann Heng.
\newblock Surgpub-video: A comprehensive surgical video framework for enhanced surgical intelligence in vision-language model.
\newblock In \emph{Proceedings of the AAAI Conference on Artificial Intelligence}, pages 6628--6635, 2026.

\bibitem[Liu et~al.(2023)Liu, Li, Wu, and Lee]{liu2023visual}
Haotian Liu, Chunyuan Li, Qingyang Wu, and Yong~Jae Lee.
\newblock Visual instruction tuning.
\newblock \emph{Advances in neural information processing systems}, 36:\penalty0 34892--34916, 2023.

\bibitem[Liu et~al.(2026)Liu, Zheng, Chen, Peng, Yin, Shao, Hu, and Yuan]{liu2026endobench}
Shengyuan Liu, Boyun Zheng, Wenting Chen, Zhihao Peng, Zhenfei Yin, Jing Shao, Jiancong Hu, and Yixuan Yuan.
\newblock Endobench: A comprehensive evaluation of multi-modal large language models for endoscopy analysis.
\newblock \emph{Advances in Neural Information Processing Systems}, 38, 2026.

\bibitem[Maaz et~al.(2024)Maaz, Rasheed, Khan, and Khan]{maaz2024video}
Muhammad Maaz, Hanoona Rasheed, Salman Khan, and Fahad Khan.
\newblock Video-chatgpt: Towards detailed video understanding via large vision and language models.
\newblock In \emph{Proceedings of the 62nd Annual Meeting of the Association for Computational Linguistics (Volume 1: Long Papers)}, pages 12585--12602, 2024.

\bibitem[Maier-Hein et~al.(2021)]{maier2021heidelberg}
Lena Maier-Hein et~al.
\newblock Heidelberg colorectal data set for surgical data science in the sensor operating room.
\newblock \emph{Scientific data}, 8\penalty0 (1):\penalty0 101, 2021.

\bibitem[Miao et~al.(2024)]{miao2024hemoset}
Albert~J Miao et~al.
\newblock Hemoset: The first blood segmentation dataset for automation of hemostasis management.
\newblock In \emph{2024 International Symposium on Medical Robotics (ISMR)}, pages 1--7. IEEE, 2024.

\bibitem[Mitra et~al.(2024)Mitra, Del~Corro, Zheng, Mahajan, Rouhana, Codas, Lu, Chen, Vrousgos, Rosset, et~al.]{mitra2024agentinstruct}
Arindam Mitra, Luciano Del~Corro, Guoqing Zheng, Shweti Mahajan, Dany Rouhana, Andres Codas, Yadong Lu, Wei-ge Chen, Olga Vrousgos, Corby Rosset, et~al.
\newblock Agentinstruct: Toward generative teaching with agentic flows.
\newblock \emph{arXiv preprint arXiv:2407.03502}, 2024.

\bibitem[Murali et~al.(2023)]{murali2023latent}
Aditya Murali et~al.
\newblock Latent graph representations for critical view of safety assessment.
\newblock \emph{IEEE TMI}, pages 1--1, 2023.

\bibitem[Nasirihaghighi et~al.(2025)]{nasirihaghighi2025gynsurg}
Sahar Nasirihaghighi et~al.
\newblock Gynsurg: A comprehensive gynecology laparoscopic surgery dataset.
\newblock In \emph{Proceedings of the 33rd ACM International Conference on Multimedia}, pages 13141--13147, 2025.

\bibitem[Nwoye et~al.(2022)]{nwoye2021rendezvous}
Chinedu~Innocent Nwoye et~al.
\newblock Rendezvous: Attention mechanisms for the recognition of surgical action triplets in endoscopic videos.
\newblock \emph{MedIA}, 78:\penalty0 102433, 2022.

\bibitem[Perez et~al.(2026)Perez, Nwoye, Kermani, Mohareri, and Jamal]{perez2026surglavi}
Alejandra Perez, Chinedu Nwoye, Ramtin~Raji Kermani, Omid Mohareri, and Muhammad~Abdullah Jamal.
\newblock Surglavi: Large-scale hierarchical dataset for surgical vision--language representation learning.
\newblock \emph{MedIA}, page 103982, 2026.

\bibitem[Psychogyios et~al.(2023)]{psychogyios2023sar}
Dimitrios Psychogyios et~al.
\newblock Sar-rarp50: Segmentation of surgical instrumentation and action recognition on robot-assisted radical prostatectomy challenge.
\newblock \emph{arXiv preprint arXiv:2401.00496}, 2023.

\bibitem[Rahmanzadehgervi et~al.(2024)Rahmanzadehgervi, Bolton, Taesiri, and Nguyen]{rahmanzadehgervi2024vision}
Pooyan Rahmanzadehgervi, Logan Bolton, Mohammad~Reza Taesiri, and Anh~Totti Nguyen.
\newblock Vision language models are blind.
\newblock In \emph{Proceedings of the Asian Conference on Computer Vision}, pages 18--34, 2024.

\bibitem[Rueckert et~al.(2026)Rueckert, Rauber, Maerkl, Klausmann, Yildiran, Gutbrod, Nunes, Moreno, Luengo, Stoyanov, et~al.]{rueckert2026comparative}
Tobias Rueckert, David Rauber, Raphaela Maerkl, Leonard Klausmann, Suemeyye~R Yildiran, Max Gutbrod, Danilo~Weber Nunes, Alvaro~Fernandez Moreno, Imanol Luengo, Danail Stoyanov, et~al.
\newblock Comparative validation of surgical phase recognition, instrument keypoint estimation, and instrument instance segmentation in endoscopy: Results of the phakir 2024 challenge.
\newblock \emph{Medical Image Analysis}, page 103945, 2026.

\bibitem[Schmidgall et~al.(2024)Schmidgall, Cho, Zakka, and Hiesinger]{schmidgall2024gp}
Samuel Schmidgall, Joseph Cho, Cyril Zakka, and William Hiesinger.
\newblock Gp-vls: A general-purpose vision language model for surgery.
\newblock \emph{arXiv preprint arXiv:2407.19305}, 2024.

\bibitem[Schmidgall et~al.(2025)Schmidgall, Opfermann, Kim, and Krieger]{schmidgall2025will}
Samuel Schmidgall, Justin~D Opfermann, Ji~Woong Kim, and Axel Krieger.
\newblock Will your next surgeon be a robot? autonomy and ai in robotic surgery.
\newblock \emph{Science robotics}, 10\penalty0 (104):\penalty0 eadt0187, 2025.

\bibitem[Schoeffmann et~al.(2018)Schoeffmann, Husslein, Kletz, Petscharnig, M{\"{u}}nzer, and Beecks]{surgicalactions160}
Klaus Schoeffmann, Heinrich Husslein, Sabrina Kletz, Stefan Petscharnig, Bernd M{\"{u}}nzer, and Christian Beecks.
\newblock Video retrieval in laparoscopic video recordings with dynamic content descriptors.
\newblock \emph{Multim. Tools Appl.}, 77\penalty0 (13):\penalty0 16813--16832, 2018.

\bibitem[Seenivasan et~al.(2022)Seenivasan, Islam, Krishna, and Ren]{seenivasan2022surgical}
Lalithkumar Seenivasan, Mobarakol Islam, Adithya~K Krishna, and Hongliang Ren.
\newblock Surgical-vqa: Visual question answering in surgical scenes using transformer.
\newblock In \emph{MICCAI}, pages 33--43. Springer, 2022.

\bibitem[Seenivasan et~al.(2023)Seenivasan, Islam, Kannan, and Ren]{seenivasan2023surgicalgpt}
Lalithkumar Seenivasan, Mobarakol Islam, Gokul Kannan, and Hongliang Ren.
\newblock Surgicalgpt: end-to-end language-vision gpt for visual question answering in surgery.
\newblock In \emph{International conference on medical image computing and computer-assisted intervention}, pages 281--290. Springer, 2023.

\bibitem[Sellergren et~al.(2025)Sellergren, Kazemzadeh, Jaroensri, Kiraly, Traverse, Kohlberger, Xu, Jamil, Hughes, Lau, et~al.]{sellergren2025medgemma}
Andrew Sellergren, Sahar Kazemzadeh, Tiam Jaroensri, Atilla Kiraly, Madeleine Traverse, Timo Kohlberger, Shawn Xu, Fayaz Jamil, C{\'\i}an Hughes, Charles Lau, et~al.
\newblock Medgemma technical report.
\newblock \emph{arXiv preprint arXiv:2507.05201}, 2025.

\bibitem[Shabtay et~al.(2025)Shabtay, Maia~Polo, Doveh, Lin, Mirza, Choshen, Yurochkin, Sun, Arbelle, Karlinsky, et~al.]{shabtay2025livexiv}
Nimrod Shabtay, Felipe Maia~Polo, Sivan Doveh, Wei Lin, Muhammad~Jehanzeb Mirza, Leshem Choshen, Mikhail Yurochkin, Yuekai Sun, Assaf Arbelle, Leonid Karlinsky, et~al.
\newblock Livexiv-a multi-modal live benchmark based on arxiv papers content.
\newblock In \emph{International Conference on Learning Representations}, pages 11470--11502, 2025.

\bibitem[Shein et~al.(2023)Shein, Brodie, and Mintz]{shein2023human}
Gabriel~Szydlo Shein, Ronit Brodie, and Yoav Mintz.
\newblock Human-machine collaboration in ai-assisted surgery: Balancing autonomy and expertise.
\newblock \emph{Artificial Intelligence in Medicine and Surgery-An Exploration of Current Trends, Potential Opportunities, and Evolving Threats-Volume 1}, 2023.

\bibitem[Shin et~al.(2026)Shin, Kim, Cho, Kim, and Oh]{shin2026surgcheck}
Jongmin Shin, Ka~Young Kim, Eunki Cho, Seong~Tae Kim, and Namkee Oh.
\newblock Surgcheck: Do vision--language models really look at images in surgical vqa?
\newblock \emph{International Journal of Computer Assisted Radiology and Surgery}, pages 1--9, 2026.

\bibitem[Singh et~al.(2025)Singh, Fry, Perelman, Tart, Ganesh, El-Kishky, McLaughlin, Low, Ostrow, Ananthram, et~al.]{singh2025openai}
Aaditya Singh, Adam Fry, Adam Perelman, Adam Tart, Adi Ganesh, Ahmed El-Kishky, Aidan McLaughlin, Aiden Low, AJ Ostrow, Akhila Ananthram, et~al.
\newblock Openai gpt-5 system card.
\newblock \emph{arXiv preprint arXiv:2601.03267}, 2025.

\bibitem[Su et~al.(2026)Su, Choudhuri, Gao, Planche, Nguyen, Zheng, Shen, Innanje, Chen, Elhamifar, et~al.]{su2026medgrpo}
Yuhao Su, Anwesa Choudhuri, Zhongpai Gao, Benjamin Planche, Van~Nguyen Nguyen, Meng Zheng, Yuhan Shen, Arun Innanje, Terrence Chen, Ehsan Elhamifar, et~al.
\newblock Medgrpo: Multi-task reinforcement learning for heterogeneous medical video understanding.
\newblock In \emph{Proceedings of the IEEE/CVF Conference on Computer Vision and Pattern Recognition}, pages 2788--2798, 2026.

\bibitem[Team(2026)]{team2026qwen3}
Qwen Team.
\newblock Qwen3. 5-omni technical report.
\newblock \emph{arXiv preprint arXiv:2604.15804}, 2026.

\bibitem[Twinanda et~al.(2016)]{twinanda2016endonet}
Andru~P Twinanda et~al.
\newblock Endonet: a deep architecture for recognition tasks on laparoscopic videos.
\newblock \emph{IEEE TMI}, 36\penalty0 (1):\penalty0 86--97, 2016.

\bibitem[Wagner et~al.(2023)]{wagner2023comparative}
Martin Wagner et~al.
\newblock Comparative validation of machine learning algorithms for surgical workflow and skill analysis with the heichole benchmark.
\newblock \emph{MedIA}, 86:\penalty0 102770, 2023.

\bibitem[Wang et~al.(2026)Wang, Zhou, Deng, Cheah, Qu, Ren, and Shen]{wang2026surgcot}
Gui Wang, YongSong Zhou, Kaijun Deng, Wooi~Ping Cheah, Rong Qu, Jianfeng Ren, and Linlin Shen.
\newblock Surgcot: Advancing spatiotemporal reasoning in surgical videos through a chain-of-thought benchmark.
\newblock In \emph{Proceedings of the IEEE/CVF Conference on Computer Vision and Pattern Recognition}, pages 17011--17021, 2026.

\bibitem[Wang et~al.(2025{\natexlab{a}})]{wang2025copesd}
Guankun Wang et~al.
\newblock Copesd: A multi-level surgical motion dataset for training large vision-language models to co-pilot endoscopic submucosal dissection.
\newblock In \emph{Proceedings of the 33rd ACM International Conference on Multimedia}, pages 12636--12643, 2025{\natexlab{a}}.

\bibitem[Wang et~al.(2025{\natexlab{b}})]{wang2025endochat}
Guankun Wang et~al.
\newblock Endochat: Grounded multimodal large language model for endoscopic surgery.
\newblock \emph{MedIA}, page 103789, 2025{\natexlab{b}}.

\bibitem[Wang et~al.(2023{\natexlab{a}})Wang, Kordi, Mishra, Liu, Smith, Khashabi, and Hajishirzi]{wang2023self}
Yizhong Wang, Yeganeh Kordi, Swaroop Mishra, Alisa Liu, Noah~A Smith, Daniel Khashabi, and Hannaneh Hajishirzi.
\newblock Self-instruct: Aligning language models with self-generated instructions.
\newblock In \emph{Proceedings of the 61st annual meeting of the association for computational linguistics (volume 1: long papers)}, pages 13484--13508, 2023{\natexlab{a}}.

\bibitem[Wang et~al.(2022)Wang, Lu, Long, Zhong, Cheung, Dou, and Liu]{wang2022autolaparo}
Ziyi Wang, Bo Lu, Yonghao Long, Fangxun Zhong, Tak-Hong Cheung, Qi Dou, and Yunhui Liu.
\newblock Autolaparo: A new dataset of integrated multi-tasks for image-guided surgical automation in laparoscopic hysterectomy.
\newblock In \emph{International Conference on Medical Image Computing and Computer-Assisted Intervention}, pages 486--496. Springer, 2022.

\bibitem[Wang et~al.(2023{\natexlab{b}})Wang, Chen, You, Xu, He, Li, Codella, Chang, and Chang]{wang2023dataset}
Zhecan Wang, Long Chen, Haoxuan You, Keyang Xu, Yicheng He, Wenhao Li, Noel Codella, Kai-Wei Chang, and Shih-Fu Chang.
\newblock Dataset bias mitigation in multiple-choice visual question answering and beyond.
\newblock In \emph{Findings of the Association for Computational Linguistics: EMNLP 2023}, pages 8598--8617, 2023{\natexlab{b}}.

\bibitem[Yang et~al.(2026)Yang, Liu, Zhai, Sun, Liu, Barsoum, Li, and Xu]{yang2026captionqa}
Shijia Yang, Yunong Liu, Bohan Zhai, Ximeng Sun, Zicheng Liu, Emad Barsoum, Manling Li, and Chenfeng Xu.
\newblock Captionqa: Is your caption as useful as the image itself?
\newblock In \emph{Proceedings of the IEEE/CVF Conference on Computer Vision and Pattern Recognition}, pages 23741--23750, 2026.

\bibitem[Yi et~al.(2026)Yi, Pan, He, Liu, Zhang, Oguntola, and Bertasius]{yi2026exact}
Han Yi, Yulu Pan, Feihong He, Xinyu Liu, Benjamin Zhang, Oluwatumininu Oguntola, and Gedas Bertasius.
\newblock Exact: A video-language benchmark for expert action analysis.
\newblock \emph{Advances in Neural Information Processing Systems}, 38, 2026.

\bibitem[Yu et~al.(2025)Yu, Wu, Chu, Ren, Huang, Chu, Zhang, He, Li, Li, et~al.]{yu2025vrbench}
Jiashuo Yu, Yue Wu, Meng Chu, Zhifei Ren, Zizheng Huang, Pei Chu, Ruijie Zhang, Yinan He, Qirui Li, Songze Li, et~al.
\newblock Vrbench: A benchmark for multi-step reasoning in long narrative videos.
\newblock In \emph{Proceedings of the IEEE/CVF International Conference on Computer Vision}, pages 21655--21666, 2025.

\bibitem[Yu et~al.(2026)Yu, Ma, Huang, Lee, Yang, Choi, Brown, Michel, Zheng, Shin, et~al.]{yu2026benchmarking}
Sihyun Yu, Nanye Ma, Pinzhi Huang, Hyunseok Lee, Shusheng Yang, June~Suk Choi, Ellis Brown, Oscar Michel, Boyang Zheng, Jinwoo Shin, et~al.
\newblock Benchmarking visual state tracking in multimodal video understanding.
\newblock \emph{arXiv preprint arXiv:2606.03920}, 2026.

\bibitem[Yu et~al.(2023)Yu, Yang, Li, Wang, Lin, Liu, Wang, and Wang]{yu2023mm}
Weihao Yu, Zhengyuan Yang, Linjie Li, Jianfeng Wang, Kevin Lin, Zicheng Liu, Xinchao Wang, and Lijuan Wang.
\newblock Mm-vet: Evaluating large multimodal models for integrated capabilities.
\newblock \emph{arXiv preprint arXiv:2308.02490}, 2023.

\bibitem[Yuan et~al.(2024{\natexlab{a}})Yuan, Kattel, Lavanchy, Navab, Srivastav, and Padoy]{yuan2024advancing}
Kun Yuan, Manasi Kattel, Jo{\"e}l~L Lavanchy, Nassir Navab, Vinkle Srivastav, and Nicolas Padoy.
\newblock Advancing surgical vqa with scene graph knowledge.
\newblock \emph{IJCARS}, pages 1--9, 2024{\natexlab{a}}.

\bibitem[Yuan et~al.(2024{\natexlab{b}})Yuan, Srivastav, Navab, and Padoy]{yuan2024procedure}
Kun Yuan, Vinkle Srivastav, Nassir Navab, and Nicolas Padoy.
\newblock Procedure-aware surgical video-language pretraining with hierarchical knowledge augmentation.
\newblock \emph{Advances in Neural Information Processing Systems}, 37:\penalty0 122952--122983, 2024{\natexlab{b}}.

\bibitem[Yuan et~al.(2025)]{yuan2025learning}
Kun Yuan et~al.
\newblock Learning multi-modal representations by watching hundreds of surgical video lectures.
\newblock \emph{MedIA}, 105:\penalty0 103644, 2025.

\bibitem[Yue et~al.(2025)Yue, Zheng, Ni, Wang, Zhang, Tong, Sun, Yu, Zhang, Sun, et~al.]{yue2025mmmu}
Xiang Yue, Tianyu Zheng, Yuansheng Ni, Yubo Wang, Kai Zhang, Shengbang Tong, Yuxuan Sun, Botao Yu, Ge Zhang, Huan Sun, et~al.
\newblock Mmmu-pro: A more robust multi-discipline multimodal understanding benchmark.
\newblock In \emph{Proceedings of the 63rd Annual Meeting of the Association for Computational Linguistics (Volume 1: Long Papers)}, pages 15134--15186, 2025.

\bibitem[Zan et~al.(2026)Zan, Huang, Liu, Chen, Xin, Zhang, Liu, Aoyan, Chen, Zhong, et~al.]{zan2026multi}
Daoguang Zan, Zhirong Huang, Wei Liu, Hanwu Chen, Shulin Xin, Linhao Zhang, Qi Liu, Li Aoyan, Lu Chen, Xiaojian Zhong, et~al.
\newblock Multi-swe-bench: A multilingual benchmark for issue resolving.
\newblock \emph{Advances in Neural Information Processing Systems}, 38, 2026.

\bibitem[Zeng et~al.(2025)]{zeng2025surgvlm}
Zhitao Zeng et~al.
\newblock Surgvlm: A large vision-language model and systematic evaluation benchmark for surgical intelligence.
\newblock \emph{arXiv preprint arXiv:2506.02555}, 2025.

\bibitem[Zhang et~al.(2025)Zhang, Su, Liu, Wang, Burgess, Sui, Wang, Aklilu, Lozano, Wei, et~al.]{zhang2025automated}
Yuhui Zhang, Yuchang Su, Yiming Liu, Xiaohan Wang, James Burgess, Elaine Sui, Chenyu Wang, Josiah Aklilu, Alejandro Lozano, Anjiang Wei, et~al.
\newblock Automated generation of challenging multiple-choice questions for vision language model evaluation.
\newblock In \emph{Proceedings of the Computer Vision and Pattern Recognition Conference}, pages 29580--29590, 2025.

\bibitem[Zhou et~al.(2024)Zhou, Faith, Xu, Leng, Xu, Liu, and Goh]{zhou2024benchx}
Yang Zhou, Tan~L Faith, Yanyu Xu, Sicong Leng, Xinxing Xu, Yong Liu, and Rick~S Goh.
\newblock Benchx: A unified benchmark framework for medical vision-language pretraining on chest x-rays.
\newblock \emph{Advances in Neural Information Processing Systems}, 37:\penalty0 6625--6647, 2024.

\end{thebibliography}
